\documentclass[10pt,twocolumn,letterpaper]{article}

\usepackage[accsupp]{axessibility}

\usepackage{cvpr}

\usepackage{graphicx}
\usepackage{amsmath}
\usepackage{amssymb}
\usepackage{booktabs}

\usepackage[pagebackref,breaklinks,colorlinks]{hyperref}

\usepackage{tabularx}
\usepackage{amssymb}
\usepackage{bbding}
\usepackage{soul,color}
\usepackage{multirow}
\usepackage{tipa}
\usepackage{amsthm}
\theoremstyle{definition}

\usepackage{bbold}
\usepackage{mathrsfs}
\usepackage{colortbl}
\usepackage[table,xcdraw]{xcolor}
\usepackage{enumitem}
\usepackage{tabularx}

\usepackage[capitalize]{cleveref}
\crefname{section}{Sec.}{Secs.}
\Crefname{section}{Section}{Sections}
\Crefname{table}{Table}{Tables}
\crefname{table}{Tab.}{Tabs.}

\def\cvprPaperID{8211}
\def\confName{CVPR}
\def\confYear{2023}

\begin{document}

\newcommand\red[1]{\textcolor{red}{#1}}
\newcommand\blue[1]{\textcolor{blue}{#1}}
\newcommand\gray[1]{\textcolor{gray}{#1}}
\newcommand\green[1]{\textcolor{green}{#1}}
\newcommand\magenta[1]{\textcolor{magenta}{#1}}

\definecolor{darkgreen}{rgb}{0.0, 0.4, 0.0}
\definecolor{orange}{rgb}{1.0, 0.49, 0.0}
\definecolor{purple}{rgb}{0.54, 0.17, 0.89}
\definecolor{ForestGreen}{RGB}{34,139,34}
\definecolor{ModernBlue}{RGB}{0,153,153}
\definecolor{LightModernBlue}{RGB}{0,204,204}
\definecolor{DarkPink}{RGB}{204,0,102}

\newcommand{\darkgreen}{\textcolor{darkgreen}}
\newcommand{\orange}{\textcolor{orange}}
\newcommand{\purple}{\textcolor{purple}}
\newcommand{\forestgreen}{\textcolor{ForestGreen}} 
\newcommand{\modernblue}{\textcolor{ModernBlue}} 
\newcommand{\darkpink}{\textcolor{DarkPink}} 

\newcommand{\hlc}[2][yellow]{{%
    \colorlet{foo}{#1}%
    \sethlcolor{foo}\hl{#2}}%
}

\newcommand{\norm}[1]{\left\lVert#1\right\rVert}

\newcommand{\ours}{\texttt{Paprika}\ }
\newcommand{\ourseos}{\texttt{Paprika}}
\newcommand{\g}{\texttt{PKG}\ }
\newcommand{\geos}{\texttt{PKG}}

\newcommand{\ct}{\magenta{CrossTask}\ }
\newcommand{\cteos}{\magenta{CrossTask}}

\newcommand{\coin}{\magenta{COIN}\ }
\newcommand{\coineos}{\magenta{COIN}}

\newcommand{\APP}{Supplementary Material\ }
\newcommand{\APPeos}{Supplementary Material}

\newcommand{\RNum}[1]{\uppercase\expandafter{\romannumeral #1\relax}}

\title{Procedure-Aware Pretraining for Instructional Video Understanding}

\author{Honglu Zhou$^{1,2}$, Roberto Martín-Martín$^{1,3}$, Mubbasir Kapadia$^2$, Silvio Savarese$^1$ and Juan Carlos Niebles$^1$\\
$^1$Salesforce Research, $^2$Rutgers University, $^3$UT Austin\\
{\tt\small $\{$hz289,mk1353$\}$@cs.rutgers.edu, robertomm@cs.utexas.edu, $\{$ssavarese,jniebles$\}$@salesforce.com}
}

\maketitle

\begin{abstract}
Our goal is to learn a video representation
that is useful for downstream
procedure understanding
tasks in instructional videos. 
Due to the small amount of available annotations, a key challenge in procedure understanding is to be able to extract from unlabeled videos the procedural knowledge such as the identity of the task (e.g., ‘make latte’), its steps (e.g., ‘pour milk’), or the potential next steps given partial progress in its execution.
Our main insight is that instructional videos depict sequences of steps that repeat between instances of the same or different tasks, and that this structure can be well represented by a Procedural Knowledge Graph (\geos{}), where nodes are discrete steps and edges connect steps that occur sequentially in the instructional activities.
This graph can then be used to generate pseudo labels to train a video representation that encodes the procedural knowledge in a more accessible form to generalize to multiple procedure understanding tasks.
We build a \geos{} by combining information from a text-based procedural knowledge database and an unlabeled instructional video corpus and then use it to generate training pseudo labels with four novel pre-training objectives. We call this \geos{}-based pre-training procedure and the resulting model \ourseos{}, \textbf{P}rocedure-\textbf{A}ware \textbf{PR}e-training for \textbf{I}nstructional \textbf{K}nowledge \textbf{A}cquisition. 
We evaluate \ourseos{} on COIN and CrossTask for procedure understanding tasks such as task recognition, step recognition, and step forecasting.
\ourseos{} yields a video representation that improves over the state of the art: up to 
$\mathbf{11.23\%}$ gains in accuracy 
in $12$ evaluation settings. 
Implementation is available at 
\url{https://github.com/salesforce/paprika}.

\end{abstract}

\section{Introduction}
\label{sec:intro}

\begin{figure}[t]
	\centering
	\includegraphics[scale=0.155]{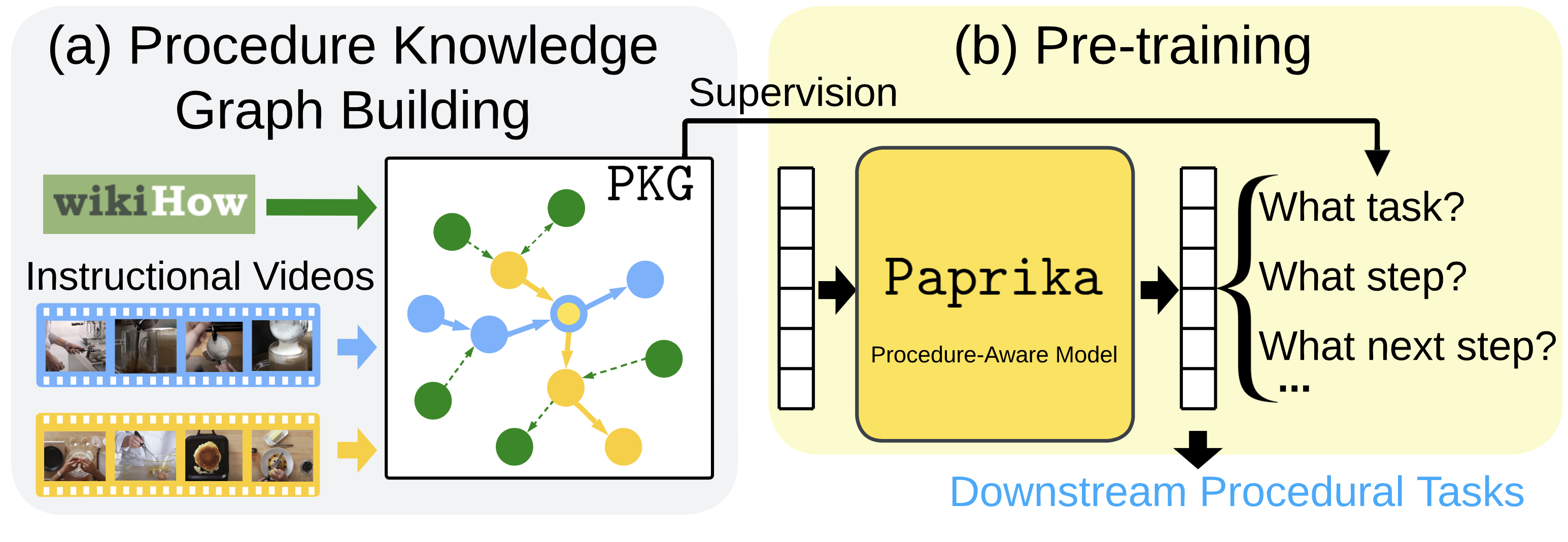}
	\vspace{-20pt}
	\caption{\textbf{Training a video representation for procedure understanding with supervision from a procedural knowledge graph}: the structure observed in instructions for procedures (from text, from videos) corresponds to sequences of steps that repeat between instances of the same or different tasks; this structure is well represented by a Procedural Knowledge Graph (\geos{}). (a) We build a \g combining text instructions with unlabeled video data, and (b) obtain a video representation by encoding the human procedural knowledge from the \g into a more general procedure-aware model (\ourseos) generating pseudo labels with the \g for several procedure understanding objectives. \ours can then be easily applied to multiple downstream procedural tasks.
    }
\label{fig:intro}
\end{figure}

Instructional videos depict humans demonstrating how to perform multi-step tasks such as cooking, making up and embroidering, repairing, or creating new objects.
For a holistic instructional video understanding, an agent has to acquire \textit{procedural knowledge}:
structural information about tasks
such as the identification of the task, its steps, or forecasting the next steps.
An agent that has acquired procedural knowledge is said to have gained \textit{procedure understanding} of instructional videos,
which can be then exploited in multiple real-world applications such as instructional video labeling, video chapterization, process mining and, when connected to a robot, robot task planning.

Our goal is to learn a novel video representation that can be applicable to a variety of procedure understanding tasks in instructional videos. 
Unfortunately, 
prior methods for video representation learning
are inadequate for this goal, as they lack the ability to capture procedural knowledge. This is because 
most of them are trained to learn the (weak) correspondence between visual and text modalities, where the text comes either from automatic-speech recognition (ASR) on the audio~\cite{s3d,videoclip}, which is noisy and error-prone, or from a caption-like
descriptive sentence (e.g., ``a video of a dog'')~\cite{alpro},
which does not contain sufficient information for
fine-grained
procedure understanding tasks such as step recognition or anticipation.
Others
are pre-trained on masked frame modeling~\cite{hero},
frame order modeling~\cite{hero} or video-audio matching~\cite{vatt}, which gives them basic video spatial, temporal or multimodal understanding but is too generic for procedure understanding tasks.

Closer to our goal,
Lin et al.~\cite{distant} propose a video foundation model for procedure understanding of instructional videos by matching 
the videos' ASR transcription (i.e., subtitle/narration)
to 
procedural steps from a text procedural knowledge database (wikiHow \cite{wikihow}) and training the 
video-representation-learning model 
to match each part of an instructional video to the corresponding step. 
Their method
only acquires 
isolated step knowledge in pre-training and
is not as suitable to gain
sophisticated
procedural knowledge.

We propose \ourseos{}, from \textbf{P}rocedure-\textbf{A}ware \textbf{PR}e-training for \textbf{I}nstructional \textbf{K}nowledge \textbf{A}cquisition, a method 
to learn a 
novel 
video representation
that encodes
procedural knowledge (\red{Fig.}~\ref{fig:intro}).
Our main insight is that the structure observed in instructional videos corresponds to sequences of steps that repeat between instances of the same or different tasks. This structure can be captured by a Procedural Knowledge Graph (\geos) where nodes are discretized steps annotated with features, and edges connect steps that occur sequentially 
in the instructional activities.
We build such a graph by combining the text and step information from 
wikiHow
and the visual and step information from 
unlabeled instructional video datasets such as HowTo100M~\cite{howto100m} automatically.
The resulting graph encodes procedural knowledge about tasks and steps, and about the temporal order and relation information of steps.

We then train our \ourseos{} model 
on multiple pre-training objectives using the \g to obtain the training labels. 
The proposed four  
pre-training objectives (\red{Sec.}~\ref{subsec:pretext_tasks}) respectively
focuses on 
procedural knowledge
about the step of a video, tasks that a step may belong to, steps that a task would require, and the general
order of steps.
These pre-training objectives are designed to allow a model 
to answer 
questions about the subgraph of the \g that a video segment may belong to.
The \g produces pseudo labels for these questions
as supervisory signals to
\textit{adapt} video representations 
produced by a video foundation model~\cite{bommasani2021opportunities} for robust and generalizable procedure understanding.

Our contributions are summarized as follows:\\
\textbf{(i)}
We propose 
a Procedural Knowledge Graph (\geos) that encodes  
human procedural knowledge from collectively leveraging a text procedural knowledge database (wikiHow) and an unlabeled instructional video corpus (HowTo100M).\\   
 \textbf{(ii)} We propose to elicit the knowledge in the \g into \ourseos, a procedure-aware 
 model, using four pre-training objectives.
    To that end, we produce pseudo lables with the \g that serve 
    as supervisory signals to train \ourseos~to learn to answer multiple questions about the subgraph of the \g that a video segment may belong to.\\  
\textbf{(iii)} We evaluate our method on the challenging COIN and CrossTask datasets on downstream procedure understanding tasks: task recognition, step recognition, and step forecasting. Regardless of the capacity of the downstream model (from simple MLP to the powerful Transformer), our method yields a representation that  
outperforms the state of the art -- up to  
$\mathbf{11.23\%}$ gains in accuracy out of $12$ evaluation settings.

\begin{figure*}[t]
	\centering
	\vspace{-10pt}
    \includegraphics[scale=0.315]{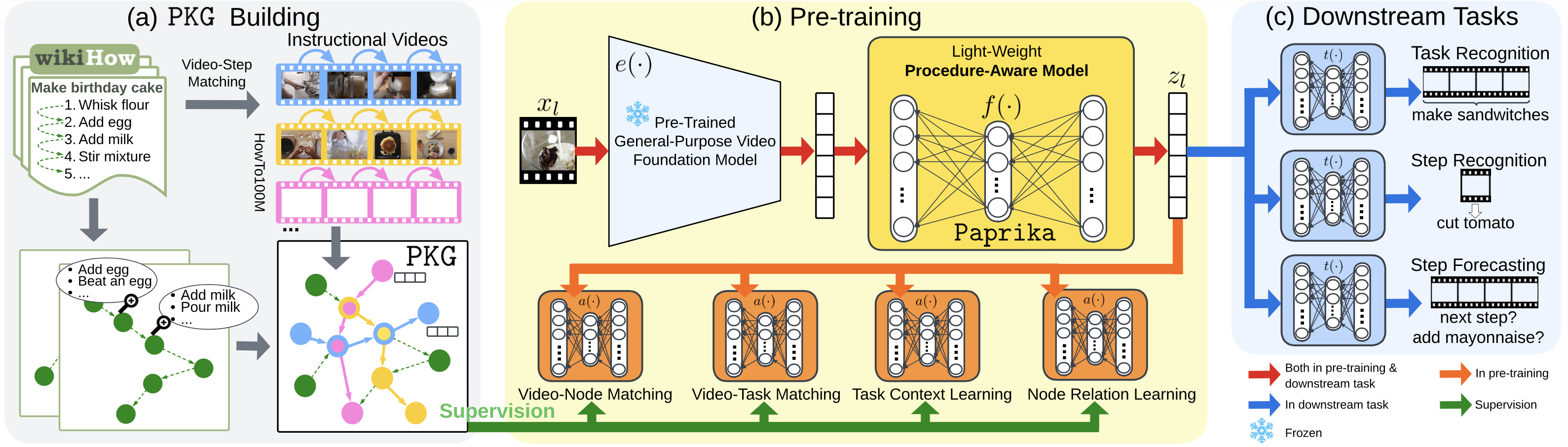}
    \vspace{-15pt}
	\caption{
	\textbf{Overview.} 
We encode procedural knowledge in a Procedural Knowledge Graph (\geos):
nodes are (clustered) steps from wikiHow that are annotated with features, and
edges connect steps that occur sequentially in the instructional activities from wikiHow or an unlabeled instructional video corpus.
Four pre-training objectives 
elicit the knowledge in the \g to \ourseos, a procedure-aware model.
We achieve this by querying the \g to produce pseudo labels for pre-training as supervisory signals.
\ours learns a  video representation that encodes procedural knowledge and thus lead to improved performance on multiple downstream procedure understanding tasks.
}
\label{fig:method}
\end{figure*}

\section{Related Work}
\label{sec:related}
\vspace{-5pt}

We focus on learning a novel 
video representation 
that can be easily adapted to downstream instructional video 
procedure understanding tasks~\cite{distant,bridgeprompt} such as procedural task recognition~\cite{ghoddoosian2022hierarchical}, step recognition~\cite{crosstask,piergiovanni2021unsupervised,kazakos2021little},  anticipation~\cite{yang2022cross,li2022order,sener2021learning,girdhar2021anticipative,miech2019leveraging}, localization~\cite{dvornik2022graph2vid,coin,elhamifar2019unsupervised,zhou2017procnets} or segmentation~\cite{ji2022learning,lu2022set,shen2022semi,youcook2,wang2021temporal,ghoddoosian2022weakly,lu2021weakly,shen2021learning}, procedure planing~\cite{chang2020procedure,bi2021procedure,sun2022plate},
and so on~\cite{xu2020benchmark,souvcek2022look,yang2021induce,instructionalvqa,jia2022egotaskqa,wang2022multimedia,huang2018finding,alayrac2016unsupervised,elhamifar2020self,doughty2020action}.

Our goal
is related to 
self-supervised learning of video representations~\cite{ruan2022survey,schiappa2022self}. \textit{Self-supervised}
pre-training objectives 
include predicting the video pace~\cite{wang2020self,benaim2020speednet}, future frame~\cite{srivastava2015unsupervised,vondrick2016anticipating,lotter2016deep}, future ASR~\cite{seo2022end} or the context~\cite{recasens2021broaden}, motion and appearance statistics~\cite{wang2019self}, solving space or/and time jigsaw puzzles~\cite{kim2019self,lee2017unsupervised,xu2019self,hero,merlot}, and identifying the odd video segment~\cite{fernando2017self} -- all exploiting the temporal signals. 
Masking has also gained popularity  
where signals of one or multiple modalities (frame/text/audio) are masked and required to be predicted~\cite{videobert,hero,clipbert,alpro,merlot,merlotreserve,perceiverio,xu2021vlm,luo2020univl}. 
Other pre-training objectives are based on spatiotemporal data augmentation~\cite{qian2021spatiotemporal,merlotreserve},
cross-modality clustering~\cite{mcn} matching~\cite{alwassel2020self,morgado2021audio,s3d,videoclip,vatt,alpro,merlot,merlotreserve,mmv,luo2020univl,kevin2022egovlp}, as well as
fine-grained noun/object-level or verb-level objectives~\cite{alpro,ge2022bridging}.

DS~\cite{distant}, MIL-NCE~\cite{s3d}, VATT~\cite{vatt},  VideoCLIP~\cite{videoclip}, VLM~\cite{xu2021vlm}, MCN~\cite{mcn}, MMV~\cite{mmv}, Hero~\cite{hero} and CBT~\cite{cbt} 
have utilized HowTo100M -- a large-scale instructional video dataset~\cite{howto100m} for pre-training. Except DS, they have utilized strictly or weakly temporally overlapped ASR with video frames as the source for contrastive learning or masked based modeling. 
DS~\cite{distant} 
argued that ASR is a suboptimal source to describe
procedural videos. 
They utilized a pre-trained language foundation model to match step headlines in wikiHow~\cite{wikihow,zhang2020reasoning,zhang2020intent,yang2021visual,zhou2022show}
to ASR sentences of video segments. The matched step headlines were then used to replace ASR sentences to learn a video 
representation learning
model.
On procedure understanding downstream tasks, DS outperforms 
models including S3D pre-trained with MIL-NCE~\cite{s3d} (MIL-NCE for short in the rest of the paper), ClipBERT~\cite{clipbert} and VideoCLIP~\cite{videoclip}.

ActionCLIP~\cite{actionclip} and Bridge-Prompt~\cite{bridgeprompt} are prompt-based models inspired by CLIP~\cite{clip}; ActionCLIP focuses on atomic action recognition~\cite{ghadiyaram2019large} (i.e., recognizing an atomic action such as ``falling'' from a short
clip
),
whereas Bridge-Prompt is for ordinal action understanding related downstream applications. 
They are related to our work but both require action annotations for training.
Instead, we focus on more effective pre-training methods for a procedure understanding 
model that encodes the procedural knowledge and avoids
laborious annotations on step class and time boundary of instructional videos.
This enables training on rich but unlabeled web data.

DS \cite{distant} leveraged wikiHow 
for instructional video understanding. This is similar to our goal of building the \g from wikiHow and instructional videos. 
However, DS 
does not focus on encoding
procedural knowledge during pre-training beyond video segment and text matching. For example,
DS does not encode
relationships between steps in the pre-trained video representation.
This is in part due to multiple challenges that need to be addressed: 
\ul{\textit{\textbf{(1)}}} the order of steps to execute a task follows certain temporal or causal constraints, \ul{\textit{\textbf{(2)}}} the execution order of steps of the task in another video instance can be different from the order that is demonstrated in the current video instance, and \ul{\textit{\textbf{(3)}}} some steps may belong to tasks that are not demonstrated in the current video instance (i.e., the cross-task characteristics of steps).
We propose the
\g 
to address these challenges. 
A model trained using our method can acquire the higher-level prior human 
procedural knowledge
instead of just the \textit{isolated} step knowledge 
that DS provides.

\section{Methodology}
\label{sec:method}

\vspace{-5pt}
\subsection{Problem Formulation}
\label{subsec:problem_formulation}
\vspace{-3pt}

Technically, 
video representation learning methods 
learn to 
represent a long video as a sequence of segment embeddings~\cite{s3d,vatt,distant}.
A video is viewed as a sequence of $L$ segments $\left[x_1, \ldots, x_l, . ., x_L\right]$, 
where
$x_l \in \mathbb{R}^{H \times W \times 3 \times F}$, $H$ and $W$ denote the
spatial 
resolution height and width, and $F$ is $\#$RGB frames of the video segment (``$\#$'' denotes ``the number of''). 
A model is pre-trained to
learn the mapping
$x_l \rightarrow z_l \in \mathbb{R}^{d}$.
Downstream models are applied on the (whole or partial)
sequence of segment embeddings $\left[z_1, \ldots, z_l, . ., z_L\right]$ to perform various tasks. 

Our goal is to learn 
$z_l$ that encodes 
procedural knowledge for downstream
procedure understanding 
tasks 
for instructional videos. 
However, pre-training a new (or fine-tuning a pre-trained) 
video model
becomes impractical for real-world settings as the model size 
grows rapidly~\cite{ruan2022survey,bommasani2021opportunities}.
We propose instead a practical framework that trains a light-weight procedure-aware  
model $f(\cdot)$ 
that refines the 
video 
segment 
feature extracted from a \textit{frozen} general-purpose 
video foundation model 
$e(\cdot)$, i.e., $z_l := f(e(x_l))$ (\red{Fig.~\ref{fig:method} (b)}). 
Our framework 
exploits the success of  
existing large 
foundation models~\cite{bommasani2021opportunities} 
and
enables parameter-efficient transfer learning (similar practices used in~\cite{flamingo,alpro}).  
$f(\cdot)$ serves as a feature adapter~\cite{houlsby2019parameter,sung2022vl} to allow the refined 
video
feature to encode the previously missing 
procedural knowledge 
for a stronger downstream 
procedure understanding
capability.
We coin our trained $f(\cdot)$ as  \ourseos.

Our key insight 
is that a text 
procedural knowledge 
database combined with unlabeled instructional videos can be utilized to build a 
Procedural Knowledge Graph (\geos)
(\red{Fig.~\ref{fig:method} (a)}) to encode procedural knowledge. The \g can provide supervisory signals for training a \textit{procedure-aware}
model. We now describe how to build the \g
from wikiHow and unlabeled procedural videos (\red{Sec.}~\ref{subsec:pkg}), and then introduce four pre-training objectives (\red{Sec.~\ref{subsec:pretext_tasks}}) that allow \ours 
to learn $z_l$  
infused with   
procedural knowledge
by mining the \geos.

\subsection{Procedural Knowledge Graph}
\label{subsec:pkg}
\vspace{-5pt}

The \g is a homogeneous graph $\mathcal{G}=(\mathcal{V}, \mathcal{E})$ with vertex set $\mathcal{V}$ and edge set $\mathcal{E}$. Nodes represent steps (e.g., `add milk') from a wide variety of tasks (e.g., `how to make latte'), and edges represent directed step transitions. That is, edge $(i, j)$ indicates
that a transition between steps in nodes $i$ and $j$ that was observed in real-life procedural data.

\noindent \textbf{\texttt{Step 1}: Obtain nodes of the \geos.} 
$\mathcal{V}$ contains
steps of tasks that may appear in instructional videos. Pre-training 
uses \textit{unlabeled}  videos, thus, there are no step annotations provided by the pre-training video corpus that can be directly used to form the
discrete
node entities. 
Inspired by DS~\cite{distant},
we resort to the step headlines in wikiHow~\cite{wikihow}. 

wikiHow is a text-based procedural knowledge database $\mathbb{B}$
that contains
articles describing the sequence of 
steps needed for the completion of a wide range of
tasks. $\mathbb{B} = $
{\scriptsize$\left\{[s_1^{(1)}, \ldots, s_{b_1}^{(1)}], \ldots, [s_1^{(t)}, \ldots, s_{b_t}^{(t)}], \ldots, [s_1^{(T)}, \ldots, s_{b_T}^{(T)}]\right\}$} where $T$ is $\#$tasks,
the subscript $b_t$ is $\#$steps of task $t$, and $s_{i}^{(t)}$ represents the natural language based summary (i.e., step headline) of $i$-th step for task $t$. Examples of wikiHow task articles 
are available in 
\red{Fig.}~\ref{fig:qualitative_one_video} and~\ref{fig:qualitative_one_segment}.

Since two step headlines in $\mathbb{B}$ can represent the same step 
but are described slightly differently, e.g., ``jack up the car'' and ``jack the car up'', 
we perform step deduplication by clustering similar step headlines. The resulting clusters are \textit{step nodes} that constitute $\mathcal{V}$. 
We list the largest step nodes in \APPeos. We find cross-task characteristics of steps, i.e., one step may belong to multiple tasks.

\noindent \textbf{\texttt{Step 2}: Add edges to the \geos.} $\mathcal{E}$ is the set of direct transitions observed in data between any two step nodes. However, most tasks in $\mathcal{B}$ have only one article, which provides only one way to complete the task through a sequence of steps.   
How to encode the different ways to complete a task $t$ that involve different execution order of steps or new steps that are absent in the article of task $t$, becomes a challenge.

Our solution is to additionally leverage an unlabeled instructional video corpus to provide more 
step transition observations.
In practice, we use MIL-NCE~\cite{s3d}, a pre-trained video-language model, 
to compute the matching score between a segment $x_l$ and a step headline $s_{i}^{(t)}$. MIL-NCE was trained to learn video and text embeddings with high matching scores 
on co-occurring frames and ASR subtitles.

We then use a thresholding criterion 
and the correspondence between step \textit{headlines} and step \textit{nodes} obtained from \textbf{\texttt{Step 1}} to match step nodes to video segments and obtain step node transitions given the temporal order of segments in videos.
The step node transitions from wikiHow or the video corpus constitute
$\mathcal{E}$.
Please refer to Supplementary Material for implementation details on graph construction.

$\mathcal{E}$ encodes the structure observed in instructional videos as it encompasses multiple sequences of steps that repeat between video instances of the same or different tasks.
In addition, 
$\mathcal{E}$ captures the relations of steps; the type of relation is not strictly defined -- the steps could be temporal or causal related -- because the step transitions that form $\mathcal{E}$ are \textit{observed} from human-provided real-life demonstrations.

\noindent \textbf{\texttt{Step 3}: Populate graph attributes.} 
It is possible to collect various forms of attributes for the \g depending on the desired use cases of the \geos.
For example, 
node attributes can be the step headline texts, task names associated with the step headlines of the node, video segments matched to the node, the distribution of 
timestamps
of the matched segments, the aggregated multimodal features of the matched segments, and so on. Edge attributes can be the source of the step node transition (from wikiHow or the video corpus), task occurrence,  distribution of 
timestamps of the transition, etc.
We describe the graph attributes we used and how we used them in Sec.~\ref{subsec:pretext_tasks}.

\subsection{Training \ours}
\label{subsec:pretext_tasks}

The \g is a rich source of supervision
for training models for procedure understanding. 
We propose four pre-training objectives as exemplars to show the possible ways in which the \g can provide supervisory signals to train $f(\cdot)$ to learn good video representations using unlabeled instructional videos.

\noindent \textbf{Video-Node Matching (VNM)}
aims at answering: what are the step nodes 
of the \g 
that are likely to be matched to the input video segment $x_l$? This pre-training objective leverages the \textit{node identity} information of the \geos, and it resembles the downstream application of independently recognizing steps of video segments.
Formally, 
{\small \begin{equation}
    a(f(e(x_l))) \rightarrow  \mathcal{V_{\text{VNM}}} 
\end{equation}
}where $a(\cdot)$ denotes the answer head model that performs the pre-training objective given the refined video segment feature $z_l$ produced by $f(\cdot)$ as input, and $ \mathcal{V_{\text{VNM}}} \subseteq \mathcal{V}$.

\noindent\textbf{Video-Task Matching (VTM)}
aims at answering: what are the tasks of the matched step nodes of the input video segment $x_l$? 
This pre-training objective leverages the node's task attribute in the \geos. VTM focuses on inferring 
the cross-task knowledge of the step nodes without the video context.
Formally, 
{\small \begin{equation}
    a(f(e(x_l))) \rightarrow  \mathcal{T_{\text{VTM}}} 
\end{equation}
}where $ \mathcal{T_{\text{VTM}}} \subseteq \mathcal{T}$, and $\mathcal{T}$ is the set of tasks ($\norm{\mathcal{T}}:=T$). 
Since HowTo100M provides task names of the long videos, we experiment with using the task names from wikiHow and/or from HowTo100M (\red{Sec.~\ref{subsec:quanti_results}}). When task names from both sources are used, VNM leads to 2 answer heads.

\noindent\textbf{Task Context Learning (TCL)}
aims at answering: for tasks
the input video segment may belong to (produced by VTM), what are the step nodes that the tasks would typically need?  TCL also leverages the node's task attribute in the \geos, but it focuses on inferring step nodes that may co-occur with the matched step node of the video segment in demonstrations. 
TCL learns the task's step context that is commonly observed in data, without the context of the current video segment. 
Formally, 
{\small \begin{equation}
    a(f(e(x_l))) \rightarrow  \mathcal{V_{\text{TCL}}} 
\end{equation}
}where $ \mathcal{V_{\text{TCL}}} \subseteq \mathcal{V}$.  When task names from both wikiHow and HowTo100M are used, TCL leads to 2 answer heads.

\noindent\textbf{Node Relation Learning (NRL)}
aims at
answering: what are the $k$-hop in-neighbors and out-neighbors of the matched step nodes of 
the input video segment $x_l$? 
$k$ ranges from 1 to a pre-defined integer $K$, and thus NRL leads to $2K$ sub-questions ($2K$ answer heads). NRL leverages the edge information of the \geos, and it focuses on learning the local multi-scale graph structure of the matched nodes of $x_l$. Predicting the in-neighbors resembles predicting the historical steps, whereas predicting the out-neighbors resembles
forecasting the next steps of 
$x_l$.
Note that the answer to NRL can be steps that come from other tasks different from the task of the current video.
Formally, 
{\small \begin{equation}
    a(f(e(x_l))) \rightarrow  \mathcal{V_{\text{NRL}}} 
\end{equation}
}where $ \mathcal{V_{\text{NRL}}} \subseteq \mathcal{V}$.

\begin{table*}[ht]
\setlength{\tabcolsep}{1.6pt}
\footnotesize
  \aboverulesep=0ex
  \belowrulesep=0ex 
\begin{center}
\begin{tabular}{ll|lll|lll|lll|lll}
\toprule
\multicolumn{2}{l|}{\multirow{3}{*}{\textbf{Pre-training Method}}}      & \multicolumn{6}{c|}{\textbf{Downstream Transformer}}                                                                                                                             & \multicolumn{6}{c}{\textbf{Downstream MLP}}                                                                                                                     \\ \cline{3-14} 
\multicolumn{2}{l|}{}                            & \multicolumn{3}{c|}{\textbf{COIN}}                                                 & \multicolumn{3}{c|}{\textbf{CrossTask}}                                            & \multicolumn{3}{c|}{\textbf{COIN}}                                                 & \multicolumn{3}{c}{\textbf{CrossTask}}                                            \\
\multicolumn{2}{l|}{}                            & \multicolumn{1}{c}{\textbf{SF}} & \multicolumn{1}{c}{\textbf{SR}} & \multicolumn{1}{c|}{\textbf{TR}} & \multicolumn{1}{c}{\textbf{SF}} & \multicolumn{1}{c}{\textbf{SR}} & \multicolumn{1}{c|}{\textbf{TR}} & \multicolumn{1}{c}{\textbf{SF}} & \multicolumn{1}{c}{\textbf{SR}} & \multicolumn{1}{c|}{\textbf{TR}} & \multicolumn{1}{c}{\textbf{SF}} & \multicolumn{1}{c}{\textbf{SR}} & \multicolumn{1}{c}{\textbf{TR}} \\
\midrule 
\multicolumn{2}{l|}{MIL-NCE$^{\ast}$~\cite{s3d} ($e(\cdot)$)}                          &  $36.55$ & $41.98$ & $76.62$ & $57.96$ & $59.90$ & $61.71$ & $3.16$ & $1.17$ &  $21.06$ &  $27.71$ & $24.98$ & $5.27$ \\
\multicolumn{2}{l|}{DS~\cite{distant}}    &  $38.13$ & $42.54$  & $79.94$ & $56.29$   & $57.11$  &  $59.49$    & $32.54$  & $34.07$  &  $72.65$ & $49.95$  &   $50.23$ &  $57.28$     \\ 
\multicolumn{2}{l|}{DS$^{\ast}$~\cite{distant}}      &  $39.54$ &  $45.97$  & $82.66$  & \cellcolor[HTML]{EBFAEB}$\mathbf{61.23}$  &  \cellcolor[HTML]{EBFAEB}$\mathbf{61.91}$  &   $64.24$    & $30.88$  & $32.74$  &  $77.66$  & $52.97$  & $53.69$   &  $61.08$  \\ \hline
\multicolumn{2}{l|}{VSM}      & $39.29$  &  $44.37$ & $82.23$  &  $57.94$ &  $58.92$  &   $62.24$   & $31.45$ & $32.66$  &   $76.51$ &  $49.59$  &  $50.01$ & $58.76$  \\ \hline
\multicolumn{1}{l|}{\multirow{11}{*}{\ours (ours)}} & \textbf{VNM}  & $41.98$  &  $49.80$ &  $82.88$  & $59.45$  &  $61.00$  &   $64.77$   &  $37.56$ &  $42.32$ &  $82.23$    & $57.08$  &  $58.23$  & $64.14$    \\ \cline{2-14} 
\multicolumn{1}{l|}{}                      & \textbf{VTM} (\textit{wikiHow})  & $42.05$  &  $49.89$ &  $84.45$   & $60.27$  &  $61.26$  &  $66.25$ & $38.13$  & $42.56$  & $82.41$  & $58.48$  &  $59.02$  & $65.82$   \\ 
\multicolumn{1}{l|}{}                      & \textbf{VTM} (\textit{HT100M}) & $41.97$  &  $48.59$ &  $83.44$ & $60.19$  &   $60.64$ &  $65.08$  & $36.87$  & $40.07$  &  $81.52$ & $56.45$  & $57.42$   & $65.61$   \\ 
\multicolumn{1}{l|}{}                      & \textbf{VTM} (\textit{wikiHow + HT100M})  & $42.10$  &  $50.02$ & \cellcolor[HTML]{EBFAEB}$\mathbf{84.73}$  &  $60.63$ &  $61.14$  &  $66.14$  & $38.12$  & $42.68$  & $82.77$  & $58.87$  & $59.30$   &  $66.14$  \\ \cline{2-14} 
\multicolumn{1}{l|}{}                      & \textbf{TCL} (\textit{wikiHow}) & $42.42$  & $50.12$  & $84.48$  & $60.27$  &   $61.40$ &  \cellcolor[HTML]{EBFAEB}$\mathbf{66.67}$   & $39.04$  & $44.16$  & $82.84$  & $58.48$  & $59.59$   &   $65.93$  \\ 
\multicolumn{1}{l|}{}                      & \textbf{TCL} (\textit{HT100M})  & $42.05$  & $48.68$  &  $83.20$ &  $60.49$ &  $61.86$  &  $66.03$   & $38.86$  &  $43.56$ & $82.55$  &  $58.38$ &  $58.63$  &  $64.66$   \\ 
\multicolumn{1}{l|}{}                      & \textbf{TCL} (\textit{wikiHow + HT100M}) &  $42.53$ &  $49.79$ & $83.95$  &  $60.19$ &  $61.67$  & $66.14$  & $38.61$  &  $43.27$ & $82.95$  & $58.40$  &  $59.26$  &  $65.08$ \\ \cline{2-14} 
\multicolumn{1}{l|}{}                      & \textbf{NRL} (\textit{1 hop})   & \cellcolor[HTML]{EBFAEB}$\mathbf{42.60}$  &  \cellcolor[HTML]{EBFAEB}$\mathbf{50.23}$ & $84.66$  &  $60.68$ &  $61.36$  &  \cellcolor[HTML]{EBFAEB}$\mathbf{66.67}$   &  \cellcolor[HTML]{EBFAEB}$\mathbf{39.58}$ & \cellcolor[HTML]{BDFABD}$\mathbf{45.38}$  &  \cellcolor[HTML]{EBFAEB}$\mathbf{83.45}$ &  $59.12$ &  $59.59$  &   $65.95$ \\ 
\multicolumn{1}{l|}{}                      & \textbf{NRL} (\textit{2 hops}) & $42.53$  &  $50.13$ & $84.31$  & $60.68$  &  $61.60$  &  \cellcolor[HTML]{EBFAEB}$\mathbf{66.67}$ &  \cellcolor[HTML]{74E974}$\mathbf{40.55}$ & \cellcolor[HTML]{74E974}$\mathbf{45.82}$  & \cellcolor[HTML]{BDFABD}$\mathbf{83.84}$  &  \cellcolor[HTML]{BDFABD}$\mathbf{60.13}$ &  \cellcolor[HTML]{BDFABD}$\mathbf{60.23}$  &  \cellcolor[HTML]{EBFAEB}$\mathbf{66.98}$   \\ \cline{2-14} 
\multicolumn{1}{l|}{}                      & \textbf{VNM} + \textbf{VTM} + \textbf{TCL} + \textbf{NRL}   & \cellcolor[HTML]{BDFABD}$\mathbf{42.65}$   &  \cellcolor[HTML]{BDFABD}$\mathbf{50.48}$ &  \cellcolor[HTML]{BDFABD}$\mathbf{85.31}$ &  \cellcolor[HTML]{BDFABD}$\mathbf{61.42}$ &  \cellcolor[HTML]{BDFABD}$\mathbf{62.38}$  & \cellcolor[HTML]{BDFABD}$\mathbf{67.09}$  & \cellcolor[HTML]{BDFABD}$\mathbf{39.82}$  &  \cellcolor[HTML]{EBFAEB}$\mathbf{44.78}$ & \cellcolor[HTML]{74E974}$\mathbf{83.88}$  & \cellcolor[HTML]{EBFAEB}$\mathbf{59.53}$   &  \cellcolor[HTML]{EBFAEB}$\mathbf{60.16}$  &  \cellcolor[HTML]{BDFABD}$\mathbf{67.41}$ \\ 
\multicolumn{1}{l|}{}                      & \red{Gains to DS~\cite{distant}}   &   \red{\scriptsize$+4.52$} & \red{\scriptsize$+7.94$}   &  \red{\scriptsize$+5.37$}  &  \red{\scriptsize$+5.13$} & \red{\scriptsize$+5.27$}  &  \red{\scriptsize$+7.60$}     &  \red{\scriptsize$+7.28$}   &  \red{\scriptsize$+10.71$}  & \red{\scriptsize$+11.23$}  &  \red{\scriptsize$+9.58$}   & \red{\scriptsize$+9.93$}   & \red{\scriptsize$+10.13$}  \\ \hline 
\multicolumn{1}{l|}{\multirow{2}{*}{\ours (ours)$^{\ast}$}}   & \textbf{VNM} + \textbf{VTM} + \textbf{TCL} + \textbf{NRL}    & \cellcolor[HTML]{74E974}$\mathbf{43.22}$  & \cellcolor[HTML]{74E974}$\mathbf{50.99}$  & \cellcolor[HTML]{74E974}$\mathbf{85.84}$   &  \cellcolor[HTML]{74E974}$\mathbf{62.63}$ &  \cellcolor[HTML]{74E974}$\mathbf{63.53}$  & \cellcolor[HTML]{74E974}$\mathbf{68.35}$ &  $38.38$  & $42.95$  & $83.41$ & \cellcolor[HTML]{74E974}$\mathbf{60.38}$  &  \cellcolor[HTML]{74E974}$\mathbf{61.21}$  &  \cellcolor[HTML]{74E974}$\mathbf{68.35}$  \\ 
\multicolumn{1}{l|}{}                      & \red{Gains to DS$^{\ast}$~\cite{distant}}   &  \red{\scriptsize$+3.68$}  & \red{\scriptsize$+5.02$}  & \red{\scriptsize$+3.18$}   &  \red{\scriptsize$+1.40$} & \red{\scriptsize$+1.62$}  &   \red{\scriptsize$+4.11$}  & \red{\scriptsize$+7.50$}   &  \red{\scriptsize$+10.21$}   &  \red{\scriptsize$+5.75$}  &  \red{\scriptsize$+7.41$}   & \red{\scriptsize$+7.52$}   &  \red{\scriptsize$+7.27$} \\ 
\bottomrule
\end{tabular}
\end{center}
\vspace{-8pt}
\scriptsize{
\textbf{SF}: Step Forecasting; \textbf{SR}: Step Recognition; \textbf{TR}: Task Recognition.\\
The top 3 performance scores of each downstream evaluation setting are highlighted with green
cells (the darker green, the better). \\
$^{\ast}$ denotes the model was pre-trained on the \textit{full} HowTo100M (HT100M) dataset; otherwise, a subset of HowTo100M containing  $85$K videos was used.\\
``\textbf{VNM} + \textbf{VTM} + \textbf{TCL} + \textbf{NRL}'' represents ``\textbf{VNM} + \textbf{VTM} (\textit{wikiHow + HT100M}) + \textbf{TCL} (\textit{wikiHow}) + \textbf{NRL}  (\textit{1 hop})''. Please see Supplementary Material for results when $K$=$2$.\\
DS$^{\ast}$~\cite{distant} reported results on COIN are SF: $38.2$ ($39.4$ from `Transformer w/ KB Transfer'), SR: $54.1$ , and TR: $88.9$ ($90.0$ from `Transformer w/ KB Transfer'). 
Our downstream experimental configurations are 
different from that in} ~\cite{distant} (e.g., w.r.t. temporal length of segments, downstream Transformer model -- ours has less parameters, etc.). 
\vspace{-8pt}
\caption{\textbf{Accuracies ($\%\uparrow$) of the downstream procedure understanding tasks under the $\mathbf{12}$ evaluation settings}. 
\ours that exploits the \g 
outperforms the SOTA methods. 
Among 
our
pre-training objectives,
NRL is the most effective one, because it exploits the structural
information of the \g and elicits the
procedural knowledge on the \textit{order} and \textit{relation} of \textit{cross-task} steps to \ourseos.
}
\label{tab:results}
\end{table*}

\section{Experiments}
\label{sec:exp_eval}

\subsection{Pre-training Dataset}
\label{subsec:pretrain_data}
\vspace{-3pt}

HowTo100M~\cite{howto100m} is a large-scale video dataset that contains over $1$M long instructional videos (videos can be over $30$ minutes) and is commonly used for video 
model pre-training. 
Videos were collected from YouTube using wikiHow article titles as search keywords~\cite{howto100m}. 
To reduce the computational cost, most of our experiments, including the construction of the \geos, only use the HowTo100M subset of size $85$K videos from \cite{bertasius2021space}.

\subsection{Evaluation Settings}
\label{subsec:pretrain_data}
\vspace{-5pt}

We study the transfer learning ability of our \ours model 
trained using the \geos ~on $\mathbf{12}$ evaluation settings: $3$ downstream tasks $\times 2$ downstream datasets $\times 2$ downstream models.
The output of the trained  
procedure-aware 
model $f(\cdot)$ is the input to the downstream model $t(\cdot)$. 
Note that the \g is only used for pre-training and it is discarded at the downstream evaluation time (test time for 
$f(\cdot)$).

\vspace{-10pt}
\subsubsection{Downstream Procedure Understanding Tasks}
\label{subsubsec:downstream_task}
\vspace{-5pt}
\noindent \textbf{Long-Term Activity/Task Recognition (TR)}
aims to classify the activity/task given all
segments from a  video.

\noindent \textbf{Step Recognition (SR)} recognizes the step class given as input the segments of a step in a video.

\noindent \textbf{Future Step Forecasting (SF)}
predicts the class of the next step given the past
video segments.
Such input contains the historical steps \textit{before} the step to predict happens.
As in \cite{distant}, we set the history to contain at least one step.

\vspace{-10pt}
\subsubsection{Downstream Datasets}
\label{subsubsec:downstream_data}
\vspace{-5pt}
We use COIN~\cite{coincvpr,coin} and CrossTask~\cite{crosstask} as the downstream datasets because the two 
cover a wide range of procedural tasks in human daily activities. 

\noindent \textbf{COIN} contains $11$K instructional videos
covering $778$ individual steps from $180$ tasks in various domains. The average number of steps per video is $3.9$.

\noindent \textbf{CrossTask} has $4.7$K instructional videos annotated with task name for each video spanning $83$ tasks with $105$ unique steps. $2.7$K videos have steps' class and temporal boundary annotations; these videos are used for the SR and SF tasks. 
$8$ steps per video on average.

\vspace{-15pt}
\subsubsection{Downstream Task Models}
\label{subsubsec:downstream_model}
\vspace{-5pt}
Segment features from the trained \textit{frozen} 
$f(\cdot)$  are the input to downstream task model $t(\cdot)$. $t(\cdot)$ is trained and evaluated on the smaller-scale downstream dataset to perform downstream tasks. 
We experiment with two 
options for $t(\cdot)$.

\noindent \textbf{MLP}.
MLP with only $1$ hidden layer is the classifier of the downstream tasks, given the input of mean aggregated sequence features. Since a shallow MLP has a limited capacity, 
performance of a MLP downstream task model heavily relies on the  
quality of the input segment features.

\noindent \textbf{Transformer}.
Since context and temporal reasoning is crucial for the downstream TR and SF tasks, 
we follow~\cite{distant} to use a one-layer Transformer~\cite{vaswani2017attention} to allow the downstream task model the capability to automatically learn to reason about segment and step relations. Transformer is a relatively stronger downstream task model compared to MLP.

\vspace{-3pt}
\subsection{Implementation Details}
\label{subsec:implement}
\vspace{-5pt}

We used 
the version of $\mathbb{B}$ 
that has $10,588$ step headlines from $T$=$1,053$ task articles. 
We used Agglomerative Clustering given the  
features of step headlines, which resulted in $10,038$  
step nodes. Length of 
segments was set to be $9.6$ seconds. The pre-trained MIL-NCE
~\cite{s3d} was used as $e(\cdot)$.
$f(\cdot)$ was a MLP with a bottleneck layer that has a dimension of $128$ as the only hidden layer. 
The refined segment feature shares the same dimension as the input segment feature (i.e., $512$).
Our pre-training objectives were cast to a multi-label classification problem with Binary Cross Entropy as the loss function. 
We used the Adam optimizer~\cite{kingma2014adam},  
a batch size of $256$, and $8$ NVIDIA A100 GPUs. 
Interested readers may refer to \APP for more details.

\subsection{Quantitative Results}
\label{subsec:quanti_results}

\subsubsection{Ablation Studies}
\label{subsec:ablation_study}
\begin{figure}[t]
    \vspace{-10pt}
	\centering
	\includegraphics[scale=0.185]{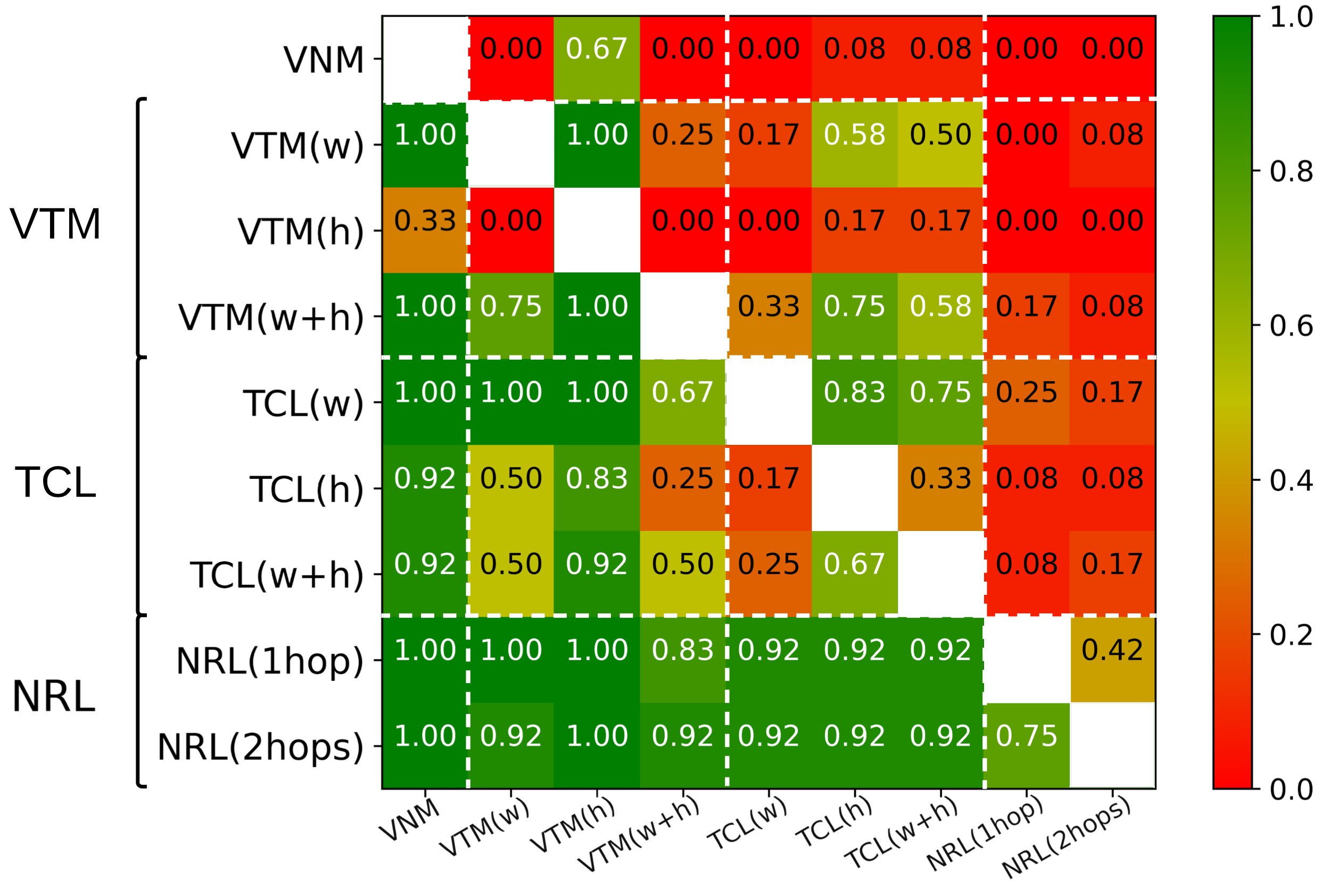}
	\vspace{-8pt}
	\caption{\textbf{Overall Performance Comparison.} This matrix compares the \textit{overall} performance of our proposed pre-training objectives.
	The value in \textbf{entry (i, j)} is the \textbf{ratio of evaluation settings} in which \textbf{the accuracy of method i $\geq$ the accuracy of method j}.
	Here, $1$ indicates method i outperforms method j in all $12$ evaluation settings. 
	The more \darkgreen{\textbf{green}} entries in the row of a method, the better its overall performance.
	NRL is the most effective method.
}
\label{fig:method_compare}
\vspace{-10pt}
\end{figure}

\vspace{-5pt}
We train \ours 
utilizing each of our pre-training objectives from  \red{Sec.~\ref{subsec:pretext_tasks}}; 
the results are in \red{Table~\ref{tab:results}}. 
We also compute a performance matrix with color-based visualization  (\red{Fig.~\ref{fig:method_compare}}) to compare the overall performance  
of the different pre-training objectives more easily.

The performance ranking of the 
pre-training objectives is
NRL $>$ TCL $>$ VTM $>$ VNM.
VNM 
is the least powerful because
it only focus on learning the simpler knowledge of matching single video segments to step nodes.

VTM (w+h) $>$ VTM (w) $>$ VTM (h) where `w' denotes wikiHow 
and `h' for HowTo100M. 
This ranking suggests that if the pre-training video corpus has the annotation of video's task name,
our method can well utilize such annotation to further improve performance. 
Utilizing the wikiHow task names is better than HowTo100M because the mapping between step headlines
and HowTo100M tasks would \textit{not} be as clean as the mapping between step headlines
and wikiHow tasks, because the former depends on the quality of the matching between a
video 
segment to a step headline.

Comparing the three variants of TCL, TCL (w) $>$ TCL (w+h) $>$ TCL (h).  
Overall, TCL (w+h) is worse than 
TCL (w) because TCL depends on the quality of the pseudo labels of VTM. As utilizing the HowTo100M task names already leads to probably problematic matched 
tasks, asking $f(\cdot)$ to further identify the step nodes that these matched 
tasks need would introduce additional noise, which eventually undermines the overall downstream performance.

\begin{figure*}[tb]
	\centering
	\vspace{-20pt} 
    \includegraphics[scale=0.325]{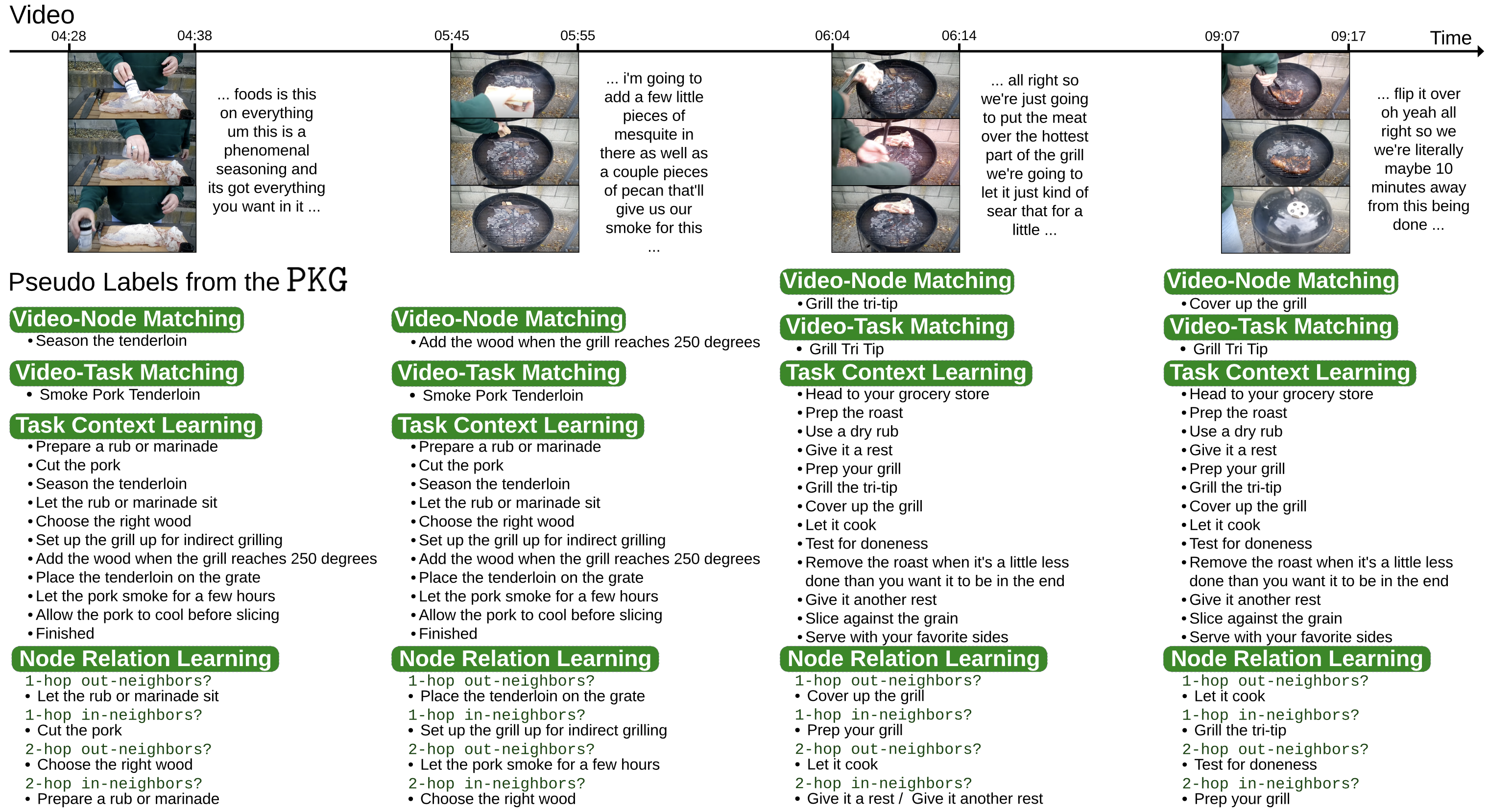}
	\caption{
	\textbf{Pseudo labels generated by the \g of one video} (title is ``Grilling A Tri-Tip ...''). 
Frames and temporally overlapped 
subtitles of four 
segments sampled from this video were shown. 
For a succinct visualization, for each pre-training objective,
we only show the result of the most confident pseudo label.
TCL and NRL provide more procedure-level context information than VNM and VTM. 
Our pseudo labels entail a much higher 
relevance to each segment than the subtitle and
allow \ours to leverage cross-task information sharing.
}
\label{fig:qualitative_one_video}
\end{figure*}

NRL (2 hops) $>$ NRL (1 hop) overall.  
NRL (2 hops) has a worse or close performance than NRL (1 hop) only when the downstream task model $t(\cdot)$ is Transformer (\red{Table~\ref{tab:results}}).
When $t(\cdot)$ is MLP, NRL (2 hops) is always clearly better.
This is because when the capacity of $t(\cdot)$ 
is limited, it desires the input video representations to encode more comprehensive information.
NRL with more hops indicates a larger exploration on the local graph structure of the \g that a video segment belongs to;
it can provide more related neighboring node/step information, and allow the learned video representations to excel at the downstream 
tasks.

We train \ours 
using all 
pre-training objectives without tuning coefficient of each loss term. \ours 
trained using all pre-training objectives
yields the best result on $8$ out of $12$ evaluation settings, which suggests the four pre-training objectives can collaborate to lead to better results. 
Compared with NRL (1 hop), the performance gains brought by VNM, VTM and TCL are relatively small.
This variant also 
fails to outperform NRL (2 hops) on the SF and SR tasks when $t(\cdot)$ is MLP.
These results highlight the superiority of NRL.

We also experiment with the full HowTo100M data. Increasing the size of the pre-training dataset, for both DS and \ourseos,
accuracies are dropped on the COIN dataset when $t(\cdot)$ is MLP (due to MLP's limited capacity to exploit the features pre-trained on the large dataset and scale well),
but we observe performance improvement of \ours on the rest $9$ out of $12$ evaluation settings.

\subsubsection{Comparison to the State of the Art (SOTA)}
\label{subsec:compare_to_sota}
\vspace{-5pt}

We have kept the model architectures and experimental setups the same between \ours and the SOTA baselines.

\noindent \textbf{MIL-NCE}~\cite{s3d} is a pre-training objective
based on
video-subtitle matching, and the subtitle can be weakly aligned to the video segment. 
This pre-training objective is widely used by video foundation models.
We use the 
frozen S3D model released by the authors as $e(\cdot)$ in our framework (and to build the \geos). 
Our reported MIL-NCE results can be interpreted as the results of removing $f(\cdot)$ in our framework.

\noindent \textbf{DS}~\cite{distant}  
proposes 
to match a video segment's subtitle to a step \textit{headline} in wikiHow by leveraging a pre-trained language model, i.e., MPNet~\cite{song2020mpnet}; and the matching results  are used as the pre-training supervisory signals. We use their proposed objective to train $f(\cdot)$--the same MLP-based architecture used by \ourseos--in our experiments.

\begin{figure*}[thb]
	\centering
	\vspace{-10pt}
        \includegraphics[scale=0.285]{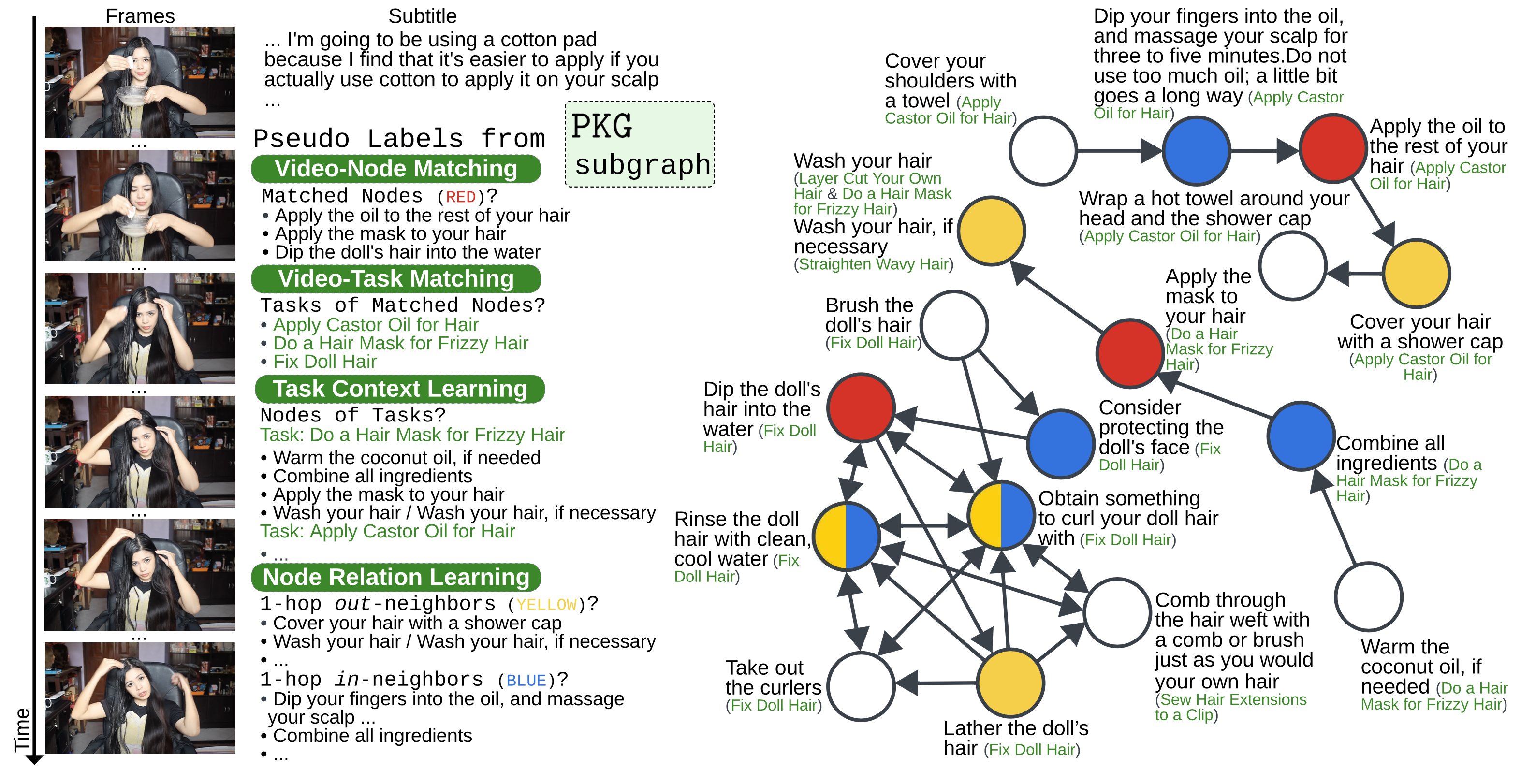}
	\vspace{-3pt}
	\caption{
	\textbf{Pseudo labels of one segment and the subgraph of the \g that this
 segment belongs to.}
	The \g encodes  the 
	procedural
	knowledge of the general
	order and relation of steps from multiple tasks. 
	This is because a node's $k$-hop neighbors can come from multiple tasks, and the edge direction encodes the general execution order of 
 steps (the order that was observed in data -- not in one specific video).
	}
\label{fig:qualitative_one_segment}
\end{figure*}

Our 
\ours
outperforms the SOTA (\red{Table~\ref{tab:results}}). 
The 
large 
performance improvement of ours compared to MIL-NCE highlights the 
ability of our \ours model
in adapting the inferior video features to be instead competent at the procedure understanding tasks.
\ours
also 
outperforms DS.
Among our proposed pre-training objectives,
VNM has the closest results to DS
because both focus on learning step knowledge; the better results of VNM attribute to multimodal matching -- matching the video \textit{frames} to the step \textit{nodes} that 
summarize and unite different step headlines in the same action. 
We perform ablation to match video frames to the wikiHow step headlines (VSM). VSM has a slightly better overall performance than DS, but worse than VNM.
VTM, TCL and NRL learn more advanced procedural knowledge from the \geos,
and therefore their gains over DS are even more obvious.

\ours
pre-trained with all four pre-training objectives obtains the highest 
gain over DS, which is 
$\mathbf{11.23\%}$ improvement in accuracy
on the COIN task recognition
task 
when $t(\cdot)$ is MLP and the HowTo100M subset is the pre-training dataset.
Overall, the 
gains 
are larger when $t(\cdot)$ is MLP than Transformer.
A shallow MLP downstream model, learned with features from \ours  
pre-trained using our full pre-training objectives, even outperforms the Transformer downstream model learned with input features from the SOTA 
pre-trained models. 
This is because our proposed method allows the video feature to early encode relation information to address the limitation of a MLP model in lacking the relational reasoning capability.

\subsection{Qualitative Results}
\label{subsec:quali_results}
\vspace{-5pt}

We present the pseudo labels generated by the \g of one 
long 
video 
in \red{Fig.~\ref{fig:qualitative_one_video}}.
Compared to the subtitles, the source of information that prior pre-training methods 
often use for supervision, pseudo labels generated by the \g entail a 
higher 
relevance to each segment. Subtitles are noisy because the narrator may not directly describe the step. 
E.g., in the $1$st segment, the narrator only mentions ``this is a phenomenal seasoning'' without explicitly 
describing the step ``seasoning the tri-tip''. 
When the camera records how a narrator is performing a step, the narrator may
omit verbally or formally describing the step.
Out of this observation, we leverage a multimodal matching function.

Assigning wikiHow steps to a video, allows one video to leverage cross-task information sharing.
As shown in the pseudo labels of VNM, the matched step headlines can come from another task. E.g., ``Season the tenderloin'' and ``Add the wood ...'' are step headlines of the task ``Smoke Pork Tenderloin'', but the task of the video is ``Grill Tri-Tip''. 
In the wikiHow article of the task ``Grill Tri-Tip'' (shown in the TCL blocks of the $3$rd and $4$th segments), the step headline corresponds to the action ``seasoning'' is ``Prep the roast'', which is vague, and no step headlines describe the action ``adding wood''.
Instead, ``seasoning'' and ``adding wood'' have a clearer step headlines to describe them in the wikiHow article of ``Smoke Pork Tenderloin''.

TCL and NRL provide more procedure-level context information 
as shown in \red{Fig.~\ref{fig:qualitative_one_video}}. The procedural knowledge conveyed by TCL and NRL is the \textit{general} prior  knowledge about the step and task of the current segment, and the knowledge is not constrained to the current step, task, or video. In other words, steps shown in the TCL or the NRL blocks can be absent in this video demonstration.

In \red{Fig.~\ref{fig:qualitative_one_segment}}, we show pseudo labels of one video segment and the subgraph of the \g that the segment belongs to.
The top $3$ matched nodes's step headlines come from different tasks, and especially the top $2$ well describe the step of the video segment.
NRL allows \ours to learn the knowledge on order and relation of cross-task steps because pseudo labels of NRL are led by the structure of the \geos.

\vspace{-5pt}
\section{Conclusion}
\label{sec:conclusion}
\vspace{-5pt}

We show how to learn 
a video representation 
for procedure understanding in instructional videos that encodes
procedural knowledge. 
The key is to leverage a Procedural Knowledge Graph (\geos) to inject procedural knowledge into the video 
representation,  
which improves the state-of-the-art performance on several tasks.

\noindent \textbf{Limitations \& Future Directions:} 
Our model is built on top of frozen video and language encoders. Future work should explore jointly updating these deep visual and text representations while also learning the procedural knowledge model.
Future work should also extend our methodology beyond the existing downstream tasks to more complex procedure understanding benchmarks.

\noindent \textbf{Social Impact:}
The final models may also be limited to perform video understanding on tasks not represented in training. These datasets primarily reflect the culture of only a portion of the world’s population, and may contain that culture’s socioeconomic biases on gender, race, ethnicity, or other features. These biases may be present in the generated pseudo labels, subgraphs, and/or overall video understanding capabilities of the resulting system.

\vspace{5pt}
\textbf{Acknowledgments:} 
Prof.
Mubbasir Kapadia was supported in part by NSF awards: IIS-1703883, IIS-1955404, IIS-1955365, RETTL-2119265, and EAGER-2122119.

{\small
\bibliographystyle{ieee_fullname}
\bibliography{main}

\begin{thebibliography}{10}\itemsep=-1pt

\bibitem{vatt}
Hassan Akbari, Liangzhe Yuan, Rui Qian, Wei-Hong Chuang, Shih-Fu Chang, Yin
  Cui, and Boqing Gong.
\newblock Vatt: Transformers for multimodal self-supervised learning from raw
  video, audio and text.
\newblock {\em Advances in Neural Information Processing Systems},
  34:24206--24221, 2021.

\bibitem{alayrac2016unsupervised}
Jean-Baptiste Alayrac, Piotr Bojanowski, Nishant Agrawal, Josef Sivic, Ivan
  Laptev, and Simon Lacoste-Julien.
\newblock Unsupervised learning from narrated instruction videos.
\newblock In {\em Proceedings of the IEEE Conference on Computer Vision and
  Pattern Recognition}, pages 4575--4583, 2016.

\bibitem{flamingo}
Jean-Baptiste Alayrac, Jeff Donahue, Pauline Luc, Antoine Miech, Iain Barr,
  Yana Hasson, Karel Lenc, Arthur Mensch, Katie Millican, Malcolm Reynolds,
  et~al.
\newblock Flamingo: a visual language model for few-shot learning.
\newblock {\em arXiv preprint arXiv:2204.14198}, 2022.

\bibitem{mmv}
Jean-Baptiste Alayrac, Adria Recasens, Rosalia Schneider, Relja
  Arandjelovi{\'c}, Jason Ramapuram, Jeffrey De~Fauw, Lucas Smaira, Sander
  Dieleman, and Andrew Zisserman.
\newblock Self-supervised multimodal versatile networks.
\newblock {\em Advances in Neural Information Processing Systems}, 33:25--37,
  2020.

\bibitem{alwassel2020self}
Humam Alwassel, Dhruv Mahajan, Bruno Korbar, Lorenzo Torresani, Bernard Ghanem,
  and Du Tran.
\newblock Self-supervised learning by cross-modal audio-video clustering.
\newblock {\em Advances in Neural Information Processing Systems},
  33:9758--9770, 2020.

\bibitem{benaim2020speednet}
Sagie Benaim, Ariel Ephrat, Oran Lang, Inbar Mosseri, William~T Freeman,
  Michael Rubinstein, Michal Irani, and Tali Dekel.
\newblock Speednet: Learning the speediness in videos.
\newblock In {\em Proceedings of the IEEE/CVF Conference on Computer Vision and
  Pattern Recognition}, pages 9922--9931, 2020.

\bibitem{bertasius2021space}
Gedas Bertasius, Heng Wang, and Lorenzo Torresani.
\newblock Is space-time attention all you need for video understanding?
\newblock In {\em ICML}, volume~2, page~4, 2021.

\bibitem{bi2021procedure}
Jing Bi, Jiebo Luo, and Chenliang Xu.
\newblock Procedure planning in instructional videos via contextual modeling
  and model-based policy learning.
\newblock In {\em Proceedings of the IEEE/CVF International Conference on
  Computer Vision}, pages 15611--15620, 2021.

\bibitem{bommasani2021opportunities}
Rishi Bommasani, Drew~A Hudson, Ehsan Adeli, Russ Altman, Simran Arora, Sydney
  von Arx, Michael~S Bernstein, Jeannette Bohg, Antoine Bosselut, Emma
  Brunskill, et~al.
\newblock On the opportunities and risks of foundation models.
\newblock {\em arXiv preprint arXiv:2108.07258}, 2021.

\bibitem{chang2020procedure}
Chien-Yi Chang, De-An Huang, Danfei Xu, Ehsan Adeli, Li Fei-Fei, and
  Juan~Carlos Niebles.
\newblock Procedure planning in instructional videos.
\newblock In {\em European Conference on Computer Vision}, pages 334--350.
  Springer, 2020.

\bibitem{mcn}
Brian Chen, Andrew Rouditchenko, Kevin Duarte, Hilde Kuehne, Samuel Thomas,
  Angie Boggust, Rameswar Panda, Brian Kingsbury, Rogerio Feris, David Harwath,
  et~al.
\newblock Multimodal clustering networks for self-supervised learning from
  unlabeled videos.
\newblock In {\em Proceedings of the IEEE/CVF International Conference on
  Computer Vision}, pages 8012--8021, 2021.

\bibitem{doughty2020action}
Hazel Doughty, Ivan Laptev, Walterio Mayol-Cuevas, and Dima Damen.
\newblock Action modifiers: Learning from adverbs in instructional videos.
\newblock In {\em Proceedings of the IEEE/CVF Conference on Computer Vision and
  Pattern Recognition}, pages 868--878, 2020.

\bibitem{dvornik2022graph2vid}
Nikita Dvornik, Isma Hadji, Hai Pham, Dhaivat Bhatt, Brais Martinez, Afsaneh
  Fazly, and Allan~D Jepson.
\newblock Graph2vid: Flow graph to video grounding for weakly-supervised
  multi-step localization.
\newblock {\em arXiv preprint arXiv:2210.04996}, 2022.

\bibitem{elhamifar2020self}
Ehsan Elhamifar and Dat Huynh.
\newblock Self-supervised multi-task procedure learning from instructional
  videos.
\newblock In {\em European Conference on Computer Vision}, pages 557--573.
  Springer, 2020.

\bibitem{elhamifar2019unsupervised}
Ehsan Elhamifar and Zwe Naing.
\newblock Unsupervised procedure learning via joint dynamic summarization.
\newblock In {\em Proceedings of the IEEE/CVF International Conference on
  Computer Vision}, pages 6341--6350, 2019.

\bibitem{fernando2017self}
Basura Fernando, Hakan Bilen, Efstratios Gavves, and Stephen Gould.
\newblock Self-supervised video representation learning with odd-one-out
  networks.
\newblock In {\em Proceedings of the IEEE conference on computer vision and
  pattern recognition}, pages 3636--3645, 2017.

\bibitem{ge2022bridging}
Yuying Ge, Yixiao Ge, Xihui Liu, Dian Li, Ying Shan, Xiaohu Qie, and Ping Luo.
\newblock Bridging video-text retrieval with multiple choice questions.
\newblock In {\em Proceedings of the IEEE/CVF Conference on Computer Vision and
  Pattern Recognition}, pages 16167--16176, 2022.

\bibitem{ghadiyaram2019large}
Deepti Ghadiyaram, Du Tran, and Dhruv Mahajan.
\newblock Large-scale weakly-supervised pre-training for video action
  recognition.
\newblock In {\em Proceedings of the IEEE/CVF conference on computer vision and
  pattern recognition}, pages 12046--12055, 2019.

\bibitem{ghoddoosian2022weakly}
Reza Ghoddoosian, Isht Dwivedi, Nakul Agarwal, Chiho Choi, and Behzad Dariush.
\newblock Weakly-supervised online action segmentation in multi-view
  instructional videos.
\newblock In {\em Proceedings of the IEEE/CVF Conference on Computer Vision and
  Pattern Recognition}, pages 13780--13790, 2022.

\bibitem{ghoddoosian2022hierarchical}
Reza Ghoddoosian, Saif Sayed, and Vassilis Athitsos.
\newblock Hierarchical modeling for task recognition and action segmentation in
  weakly-labeled instructional videos.
\newblock In {\em Proceedings of the IEEE/CVF Winter Conference on Applications
  of Computer Vision}, pages 1922--1932, 2022.

\bibitem{girdhar2021anticipative}
Rohit Girdhar and Kristen Grauman.
\newblock Anticipative video transformer.
\newblock In {\em Proceedings of the IEEE/CVF International Conference on
  Computer Vision}, pages 13505--13515, 2021.

\bibitem{houlsby2019parameter}
Neil Houlsby, Andrei Giurgiu, Stanislaw Jastrzebski, Bruna Morrone, Quentin
  De~Laroussilhe, Andrea Gesmundo, Mona Attariyan, and Sylvain Gelly.
\newblock Parameter-efficient transfer learning for nlp.
\newblock In {\em International Conference on Machine Learning}, pages
  2790--2799. PMLR, 2019.

\bibitem{huang2018finding}
De-An Huang, Shyamal Buch, Lucio Dery, Animesh Garg, Li Fei-Fei, and
  Juan~Carlos Niebles.
\newblock Finding" it": Weakly-supervised reference-aware visual grounding in
  instructional videos.
\newblock In {\em Proceedings of the IEEE Conference on Computer Vision and
  Pattern Recognition}, pages 5948--5957, 2018.

\bibitem{perceiverio}
Andrew Jaegle, Sebastian Borgeaud, Jean-Baptiste Alayrac, Carl Doersch, Catalin
  Ionescu, David Ding, Skanda Koppula, Daniel Zoran, Andrew Brock, Evan
  Shelhamer, et~al.
\newblock Perceiver io: A general architecture for structured inputs \&
  outputs.
\newblock {\em arXiv preprint arXiv:2107.14795}, 2021.

\bibitem{ji2022learning}
Lei Ji, Chenfei Wu, Daisy Zhou, Kun Yan, Edward Cui, Xilin Chen, and Nan Duan.
\newblock Learning temporal video procedure segmentation from an automatically
  collected large dataset.
\newblock In {\em Proceedings of the IEEE/CVF Winter Conference on Applications
  of Computer Vision}, pages 1506--1515, 2022.

\bibitem{jia2022egotaskqa}
Baoxiong Jia, Ting Lei, Song-Chun Zhu, and Siyuan Huang.
\newblock Egotaskqa: Understanding human tasks in egocentric videos.
\newblock {\em arXiv preprint arXiv:2210.03929}, 2022.

\bibitem{kazakos2021little}
Evangelos Kazakos, Jaesung Huh, Arsha Nagrani, Andrew Zisserman, and Dima
  Damen.
\newblock With a little help from my temporal context: Multimodal egocentric
  action recognition.
\newblock {\em arXiv preprint arXiv:2111.01024}, 2021.

\bibitem{kim2019self}
Dahun Kim, Donghyeon Cho, and In~So Kweon.
\newblock Self-supervised video representation learning with space-time cubic
  puzzles.
\newblock In {\em Proceedings of the AAAI conference on artificial
  intelligence}, volume~33, pages 8545--8552, 2019.

\bibitem{kingma2014adam}
Diederik~P Kingma and Jimmy Ba.
\newblock Adam: A method for stochastic optimization.
\newblock {\em arXiv preprint arXiv:1412.6980}, 2014.

\bibitem{wikihow}
Mahnaz Koupaee and William~Yang Wang.
\newblock Wikihow: A large scale text summarization dataset.
\newblock {\em arXiv preprint arXiv:1810.09305}, 2018.

\bibitem{lee2017unsupervised}
Hsin-Ying Lee, Jia-Bin Huang, Maneesh Singh, and Ming-Hsuan Yang.
\newblock Unsupervised representation learning by sorting sequences.
\newblock In {\em Proceedings of the IEEE international conference on computer
  vision}, pages 667--676, 2017.

\bibitem{clipbert}
Jie Lei, Linjie Li, Luowei Zhou, Zhe Gan, Tamara~L Berg, Mohit Bansal, and
  Jingjing Liu.
\newblock Less is more: Clipbert for video-and-language learning via sparse
  sampling.
\newblock In {\em Proceedings of the IEEE/CVF Conference on Computer Vision and
  Pattern Recognition}, pages 7331--7341, 2021.

\bibitem{alpro}
Dongxu Li, Junnan Li, Hongdong Li, Juan~Carlos Niebles, and Steven~CH Hoi.
\newblock Align and prompt: Video-and-language pre-training with entity
  prompts.
\newblock In {\em Proceedings of the IEEE/CVF Conference on Computer Vision and
  Pattern Recognition}, pages 4953--4963, 2022.

\bibitem{hero}
Linjie Li, Yen-Chun Chen, Yu Cheng, Zhe Gan, Licheng Yu, and Jingjing Liu.
\newblock Hero: Hierarchical encoder for video+ language omni-representation
  pre-training.
\newblock In {\em Proceedings of the 2020 Conference on Empirical Methods in
  Natural Language Processing (EMNLP)}, pages 2046--2065, 2020.

\bibitem{bridgeprompt}
Muheng Li, Lei Chen, Yueqi Duan, Zhilan Hu, Jianjiang Feng, Jie Zhou, and Jiwen
  Lu.
\newblock Bridge-prompt: Towards ordinal action understanding in instructional
  videos.
\newblock In {\em Proceedings of the IEEE/CVF Conference on Computer Vision and
  Pattern Recognition}, pages 19880--19889, 2022.

\bibitem{li2022order}
Muheng Li, Lei Chen, Jiwen Lu, Jianjiang Feng, and Jie Zhou.
\newblock Order-constrained representation learning for instructional video
  prediction.
\newblock {\em IEEE Transactions on Circuits and Systems for Video Technology},
  2022.

\bibitem{kevin2022egovlp}
Kevin~Qinghong Lin, Alex~Jinpeng Wang, Mattia Soldan, Michael Wray, Rui Yan,
  Eric~Zhongcong Xu, Difei Gao, Rongcheng Tu, Wenzhe Zhao, Weijie Kong,
  Chengfei Cai, Hongfa Wang, Dima Damen, Bernard Ghanem, Wei Liu, and
  Mike~Zheng Shou.
\newblock Egocentric video-language pretraining.
\newblock {\em arXiv preprint arXiv:2206.01670}, 2022.

\bibitem{distant}
Xudong Lin, Fabio Petroni, Gedas Bertasius, Marcus Rohrbach, Shih-Fu Chang, and
  Lorenzo Torresani.
\newblock Learning to recognize procedural activities with distant supervision.
\newblock In {\em Proceedings of the IEEE/CVF Conference on Computer Vision and
  Pattern Recognition}, pages 13853--13863, 2022.

\bibitem{lotter2016deep}
William Lotter, Gabriel Kreiman, and David Cox.
\newblock Deep predictive coding networks for video prediction and unsupervised
  learning.
\newblock {\em arXiv preprint arXiv:1605.08104}, 2016.

\bibitem{lu2021weakly}
Zijia Lu and Ehsan Elhamifar.
\newblock Weakly-supervised action segmentation and alignment via
  transcript-aware union-of-subspaces learning.
\newblock In {\em Proceedings of the IEEE/CVF International Conference on
  Computer Vision}, pages 8085--8095, 2021.

\bibitem{lu2022set}
Zijia Lu and Ehsan Elhamifar.
\newblock Set-supervised action learning in procedural task videos via pairwise
  order consistency.
\newblock In {\em Proceedings of the IEEE/CVF Conference on Computer Vision and
  Pattern Recognition}, pages 19903--19913, 2022.

\bibitem{luo2020univl}
Huaishao Luo, Lei Ji, Botian Shi, Haoyang Huang, Nan Duan, Tianrui Li, Jason
  Li, Taroon Bharti, and Ming Zhou.
\newblock Univl: A unified video and language pre-training model for multimodal
  understanding and generation.
\newblock {\em arXiv preprint arXiv:2002.06353}, 2020.

\bibitem{s3d}
Antoine Miech, Jean-Baptiste Alayrac, Lucas Smaira, Ivan Laptev, Josef Sivic,
  and Andrew Zisserman.
\newblock End-to-end learning of visual representations from uncurated
  instructional videos.
\newblock In {\em Proceedings of the IEEE/CVF Conference on Computer Vision and
  Pattern Recognition}, pages 9879--9889, 2020.

\bibitem{miech2019leveraging}
Antoine Miech, Ivan Laptev, Josef Sivic, Heng Wang, Lorenzo Torresani, and Du
  Tran.
\newblock Leveraging the present to anticipate the future in videos.
\newblock In {\em Proceedings of the IEEE/CVF Conference on Computer Vision and
  Pattern Recognition Workshops}, pages 0--0, 2019.

\bibitem{howto100m}
Antoine Miech, Dimitri Zhukov, Jean-Baptiste Alayrac, Makarand Tapaswi, Ivan
  Laptev, and Josef Sivic.
\newblock How{T}o100{M}: {L}earning a {T}ext-{V}ideo {E}mbedding by {W}atching
  {H}undred {M}illion {N}arrated {V}ideo {C}lips.
\newblock In {\em ICCV}, 2019.

\bibitem{morgado2021audio}
Pedro Morgado, Nuno Vasconcelos, and Ishan Misra.
\newblock Audio-visual instance discrimination with cross-modal agreement.
\newblock In {\em Proceedings of the IEEE/CVF Conference on Computer Vision and
  Pattern Recognition}, pages 12475--12486, 2021.

\bibitem{paszke2019pytorch}
Adam Paszke, Sam Gross, Francisco Massa, Adam Lerer, James Bradbury, Gregory
  Chanan, Trevor Killeen, Zeming Lin, Natalia Gimelshein, Luca Antiga, et~al.
\newblock Pytorch: An imperative style, high-performance deep learning library.
\newblock {\em Advances in neural information processing systems}, 32, 2019.

\bibitem{piergiovanni2021unsupervised}
AJ Piergiovanni, Anelia Angelova, Michael~S Ryoo, and Irfan Essa.
\newblock Unsupervised discovery of actions in instructional videos.
\newblock {\em arXiv preprint arXiv:2106.14733}, 2021.

\bibitem{qian2021spatiotemporal}
Rui Qian, Tianjian Meng, Boqing Gong, Ming-Hsuan Yang, Huisheng Wang, Serge
  Belongie, and Yin Cui.
\newblock Spatiotemporal contrastive video representation learning.
\newblock In {\em Proceedings of the IEEE/CVF Conference on Computer Vision and
  Pattern Recognition}, pages 6964--6974, 2021.

\bibitem{clip}
Alec Radford, Jong~Wook Kim, Chris Hallacy, Aditya Ramesh, Gabriel Goh,
  Sandhini Agarwal, Girish Sastry, Amanda Askell, Pamela Mishkin, Jack Clark,
  et~al.
\newblock Learning transferable visual models from natural language
  supervision.
\newblock In {\em International Conference on Machine Learning}, pages
  8748--8763. PMLR, 2021.

\bibitem{recasens2021broaden}
Adria Recasens, Pauline Luc, Jean-Baptiste Alayrac, Luyu Wang, Florian Strub,
  Corentin Tallec, Mateusz Malinowski, Viorica P{\u{a}}tr{\u{a}}ucean, Florent
  Altch{\'e}, Michal Valko, et~al.
\newblock Broaden your views for self-supervised video learning.
\newblock In {\em Proceedings of the IEEE/CVF International Conference on
  Computer Vision}, pages 1255--1265, 2021.

\bibitem{ruan2022survey}
Ludan Ruan and Qin Jin.
\newblock Survey: Transformer based video-language pre-training.
\newblock {\em AI Open}, 2022.

\bibitem{schiappa2022self}
Madeline~C Schiappa, Yogesh~S Rawat, and Mubarak Shah.
\newblock Self-supervised learning for videos: A survey.
\newblock {\em arXiv preprint arXiv:2207.00419}, 2022.

\bibitem{sener2021learning}
Fadime Sener, Rishabh Saraf, and Angela Yao.
\newblock Learning video models from text: Zero-shot anticipation for
  procedural actions.
\newblock {\em arXiv preprint arXiv:2106.03158}, 2021.

\bibitem{seo2022end}
Paul~Hongsuck Seo, Arsha Nagrani, Anurag Arnab, and Cordelia Schmid.
\newblock End-to-end generative pretraining for multimodal video captioning.
\newblock In {\em Proceedings of the IEEE/CVF Conference on Computer Vision and
  Pattern Recognition}, pages 17959--17968, 2022.

\bibitem{shen2022semi}
Yuhan Shen and Ehsan Elhamifar.
\newblock Semi-weakly-supervised learning of complex actions from instructional
  task videos.
\newblock In {\em Proceedings of the IEEE/CVF Conference on Computer Vision and
  Pattern Recognition}, pages 3344--3354, 2022.

\bibitem{shen2021learning}
Yuhan Shen, Lu Wang, and Ehsan Elhamifar.
\newblock Learning to segment actions from visual and language instructions via
  differentiable weak sequence alignment.
\newblock In {\em Proceedings of the IEEE/CVF Conference on Computer Vision and
  Pattern Recognition}, pages 10156--10165, 2021.

\bibitem{song2020mpnet}
Kaitao Song, Xu Tan, Tao Qin, Jianfeng Lu, and Tie-Yan Liu.
\newblock Mpnet: Masked and permuted pre-training for language understanding.
\newblock {\em Advances in Neural Information Processing Systems},
  33:16857--16867, 2020.

\bibitem{souvcek2022look}
Tom{\'a}{\v{s}} Sou{\v{c}}ek, Jean-Baptiste Alayrac, Antoine Miech, Ivan
  Laptev, and Josef Sivic.
\newblock Look for the change: Learning object states and state-modifying
  actions from untrimmed web videos.
\newblock In {\em Proceedings of the IEEE/CVF Conference on Computer Vision and
  Pattern Recognition}, pages 13956--13966, 2022.

\bibitem{srivastava2015unsupervised}
Nitish Srivastava, Elman Mansimov, and Ruslan Salakhudinov.
\newblock Unsupervised learning of video representations using lstms.
\newblock In {\em International conference on machine learning}, pages
  843--852. PMLR, 2015.

\bibitem{cbt}
Chen Sun, Fabien Baradel, Kevin Murphy, and Cordelia Schmid.
\newblock Learning video representations using contrastive bidirectional
  transformer.
\newblock {\em arXiv preprint arXiv:1906.05743}, 2019.

\bibitem{videobert}
Chen Sun, Austin Myers, Carl Vondrick, Kevin Murphy, and Cordelia Schmid.
\newblock Videobert: A joint model for video and language representation
  learning.
\newblock In {\em Proceedings of the IEEE/CVF International Conference on
  Computer Vision}, pages 7464--7473, 2019.

\bibitem{sun2022plate}
Jiankai Sun, De-An Huang, Bo Lu, Yun-Hui Liu, Bolei Zhou, and Animesh Garg.
\newblock Plate: Visually-grounded planning with transformers in procedural
  tasks.
\newblock {\em IEEE Robotics and Automation Letters}, 7(2):4924--4930, 2022.

\bibitem{sung2022vl}
Yi-Lin Sung, Jaemin Cho, and Mohit Bansal.
\newblock Vl-adapter: Parameter-efficient transfer learning for
  vision-and-language tasks.
\newblock In {\em Proceedings of the IEEE/CVF Conference on Computer Vision and
  Pattern Recognition}, pages 5227--5237, 2022.

\bibitem{coincvpr}
Yansong Tang, Dajun Ding, Yongming Rao, Yu Zheng, Danyang Zhang, Lili Zhao,
  Jiwen Lu, and Jie Zhou.
\newblock Coin: A large-scale dataset for comprehensive instructional video
  analysis.
\newblock In {\em Proceedings of the IEEE/CVF Conference on Computer Vision and
  Pattern Recognition}, pages 1207--1216, 2019.

\bibitem{coin}
Yansong Tang, Jiwen Lu, and Jie Zhou.
\newblock Comprehensive instructional video analysis: The coin dataset and
  performance evaluation.
\newblock {\em IEEE transactions on pattern analysis and machine intelligence},
  43(9):3138--3153, 2020.

\bibitem{vaswani2017attention}
Ashish Vaswani, Noam Shazeer, Niki Parmar, Jakob Uszkoreit, Llion Jones,
  Aidan~N Gomez, {\L}ukasz Kaiser, and Illia Polosukhin.
\newblock Attention is all you need.
\newblock {\em Advances in neural information processing systems}, 30, 2017.

\bibitem{vondrick2016anticipating}
Carl Vondrick, Hamed Pirsiavash, and Antonio Torralba.
\newblock Anticipating visual representations from unlabeled video.
\newblock In {\em Proceedings of the IEEE conference on computer vision and
  pattern recognition}, pages 98--106, 2016.

\bibitem{wang2021temporal}
Dong Wang, Di Hu, Xingjian Li, and Dejing Dou.
\newblock Temporal relational modeling with self-supervision for action
  segmentation.
\newblock In {\em Proceedings of the AAAI Conference on Artificial
  Intelligence}, volume~35, pages 2729--2737, 2021.

\bibitem{wang2019self}
Jiangliu Wang, Jianbo Jiao, Linchao Bao, Shengfeng He, Yunhui Liu, and Wei Liu.
\newblock Self-supervised spatio-temporal representation learning for videos by
  predicting motion and appearance statistics.
\newblock In {\em Proceedings of the IEEE/CVF Conference on Computer Vision and
  Pattern Recognition}, pages 4006--4015, 2019.

\bibitem{wang2020self}
Jiangliu Wang, Jianbo Jiao, and Yun-Hui Liu.
\newblock Self-supervised video representation learning by pace prediction.
\newblock In {\em European conference on computer vision}, pages 504--521.
  Springer, 2020.

\bibitem{actionclip}
Mengmeng Wang, Jiazheng Xing, and Yong Liu.
\newblock Actionclip: A new paradigm for video action recognition.
\newblock {\em arXiv preprint arXiv:2109.08472}, 2021.

\bibitem{wang2022multimedia}
Qingyun Wang, Manling Li, Hou~Pong Chan, Lifu Huang, Julia Hockenmaier, Girish
  Chowdhary, and Heng Ji.
\newblock Multimedia generative script learning for task planning.
\newblock {\em arXiv preprint arXiv:2208.12306}, 2022.

\bibitem{instructionalvqa}
Shaojie Wang, Wentian Zhao, Ziyi Kou, Jing Shi, and Chenliang Xu.
\newblock How to make a blt sandwich? learning vqa towards understanding web
  instructional videos.
\newblock In {\em Proceedings of the IEEE/CVF Winter Conference on Applications
  of Computer Vision}, pages 1130--1139, 2021.

\bibitem{xu2019self}
Dejing Xu, Jun Xiao, Zhou Zhao, Jian Shao, Di Xie, and Yueting Zhuang.
\newblock Self-supervised spatiotemporal learning via video clip order
  prediction.
\newblock In {\em Proceedings of the IEEE/CVF Conference on Computer Vision and
  Pattern Recognition}, pages 10334--10343, 2019.

\bibitem{xu2020benchmark}
Frank~F Xu, Lei Ji, Botian Shi, Junyi Du, Graham Neubig, Yonatan Bisk, and Nan
  Duan.
\newblock A benchmark for structured procedural knowledge extraction from
  cooking videos.
\newblock {\em arXiv preprint arXiv:2005.00706}, 2020.

\bibitem{xu2021vlm}
Hu Xu, Gargi Ghosh, Po-Yao Huang, Prahal Arora, Masoumeh Aminzadeh, Christoph
  Feichtenhofer, Florian Metze, and Luke Zettlemoyer.
\newblock Vlm: Task-agnostic video-language model pre-training for video
  understanding.
\newblock {\em arXiv preprint arXiv:2105.09996}, 2021.

\bibitem{videoclip}
Hu Xu, Gargi Ghosh, Po-Yao Huang, Dmytro Okhonko, Armen Aghajanyan, Florian
  Metze, Luke Zettlemoyer, and Christoph Feichtenhofer.
\newblock Videoclip: Contrastive pre-training for zero-shot video-text
  understanding.
\newblock In {\em Proceedings of the 2021 Conference on Empirical Methods in
  Natural Language Processing}, pages 6787--6800, 2021.

\bibitem{yang2021induce}
Yue Yang, Joongwon Kim, Artemis Panagopoulou, Mark Yatskar, and Chris
  Callison-Burch.
\newblock Induce, edit, retrieve: Language grounded multimodal schema for
  instructional video retrieval.
\newblock {\em arXiv preprint arXiv:2111.09276}, 2021.

\bibitem{yang2021visual}
Yue Yang, Artemis Panagopoulou, Qing Lyu, Li Zhang, Mark Yatskar, and Chris
  Callison-Burch.
\newblock Visual goal-step inference using wikihow.
\newblock {\em arXiv preprint arXiv:2104.05845}, 2021.

\bibitem{yang2022cross}
Zhengyuan Yang, Jingen Liu, Jing Huang, Xiaodong He, Tao Mei, Chenliang Xu, and
  Jiebo Luo.
\newblock Cross-modal contrastive distillation for instructional activity
  anticipation.
\newblock {\em arXiv preprint arXiv:2201.06734}, 2022.

\bibitem{merlotreserve}
Rowan Zellers, Jiasen Lu, Ximing Lu, Youngjae Yu, Yanpeng Zhao, Mohammadreza
  Salehi, Aditya Kusupati, Jack Hessel, Ali Farhadi, and Yejin Choi.
\newblock Merlot reserve: Neural script knowledge through vision and language
  and sound.
\newblock In {\em Proceedings of the IEEE/CVF Conference on Computer Vision and
  Pattern Recognition}, pages 16375--16387, 2022.

\bibitem{merlot}
Rowan Zellers, Ximing Lu, Jack Hessel, Youngjae Yu, Jae~Sung Park, Jize Cao,
  Ali Farhadi, and Yejin Choi.
\newblock Merlot: Multimodal neural script knowledge models.
\newblock {\em Advances in Neural Information Processing Systems},
  34:23634--23651, 2021.

\bibitem{zhang2020intent}
Li Zhang, Qing Lyu, and Chris Callison-Burch.
\newblock Intent detection with wikihow.
\newblock {\em arXiv preprint arXiv:2009.05781}, 2020.

\bibitem{zhang2020reasoning}
Li Zhang, Qing Lyu, and Chris Callison-Burch.
\newblock Reasoning about goals, steps, and temporal ordering with wikihow.
\newblock {\em arXiv preprint arXiv:2009.07690}, 2020.

\bibitem{zhou2017procnets}
Luowei Zhou, Chenliang Xu, and Jason~J Corso.
\newblock Procnets: Learning to segment procedures in untrimmed and
  unconstrained videos.
\newblock {\em arXiv preprint arXiv:1703.09788}, 2(6):7, 2017.

\bibitem{youcook2}
Luowei Zhou, Chenliang Xu, and Jason~J Corso.
\newblock Towards automatic learning of procedures from web instructional
  videos.
\newblock In {\em Thirty-Second AAAI Conference on Artificial Intelligence},
  2018.

\bibitem{zhou2022show}
Shuyan Zhou, Li Zhang, Yue Yang, Qing Lyu, Pengcheng Yin, Chris Callison-Burch,
  and Graham Neubig.
\newblock Show me more details: Discovering hierarchies of procedures from
  semi-structured web data.
\newblock {\em arXiv preprint arXiv:2203.07264}, 2022.

\bibitem{crosstask}
Dimitri Zhukov, Jean-Baptiste Alayrac, Ramazan~Gokberk Cinbis, David Fouhey,
  Ivan Laptev, and Josef Sivic.
\newblock Cross-task weakly supervised learning from instructional videos.
\newblock In {\em Proceedings of the IEEE/CVF Conference on Computer Vision and
  Pattern Recognition}, pages 3537--3545, 2019.

\end{thebibliography}
}

\clearpage
\section*{Supplementary Material}

\noindent This Supplementary Material is organized as follows:

\noindent A. Additional Results

\quad A.1. More Qualitative Results

\quad A.2. Results of \ours with 2 Hops

\quad A.3. Leveraging the \g for Downstream Tasks

\noindent B. Implementation Details

\quad B.1. Node \& Edge Construction of the \geos

\quad B.2. Pseudo Label Generation

\quad B.3. Pretraining \ours

\quad B.4. Downstream Evaluation

\quad B.5. Implementation of Baselines

\bigskip
\bigskip

\begin{figure*}[thb]
	\centering
     \includegraphics[scale=0.31]{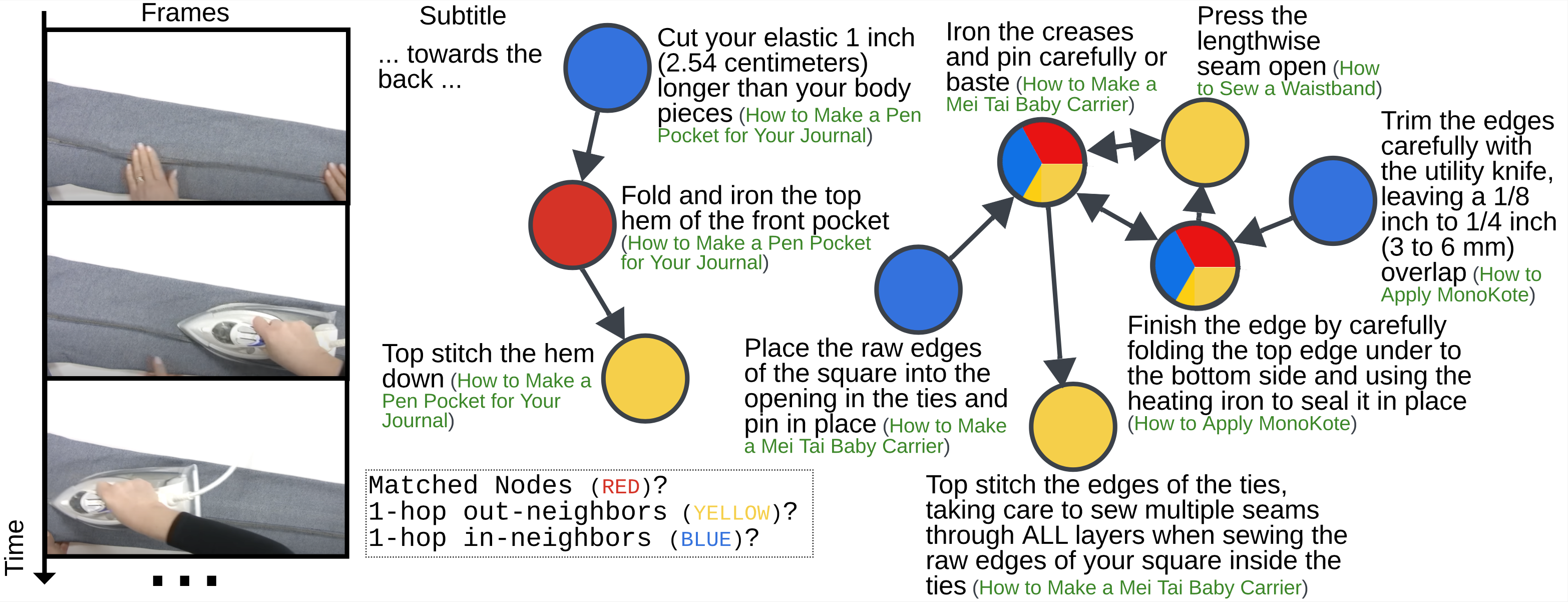}
	\caption{\textbf{The \g subgraph that a video segment belongs to.} 
	The temporally overlapped subtitle of this video segment fails to describe the step of the segment because the action of the step are recorded by the camera. The matched nodes of the \g capture the action ``ironing'' of the step. According to the \g subgraph, the action before ``ironing'' can be ``trimming'' or ``cutting'', and the step after this video segment can be ``stitch the hem/edges''.
    The steps of the \g subgraph that the video segment belongs to can come from multiple tasks (e.g., ``How to Sew a Waistband'', ``How to Apply MonoKote'', ``How to Make a Mei Tai Baby Carrier'', etc).
}
\label{appendix_fig:qua_3515}
\end{figure*}

\begin{figure*}[thb]
	\centering
        \includegraphics[scale=0.31]{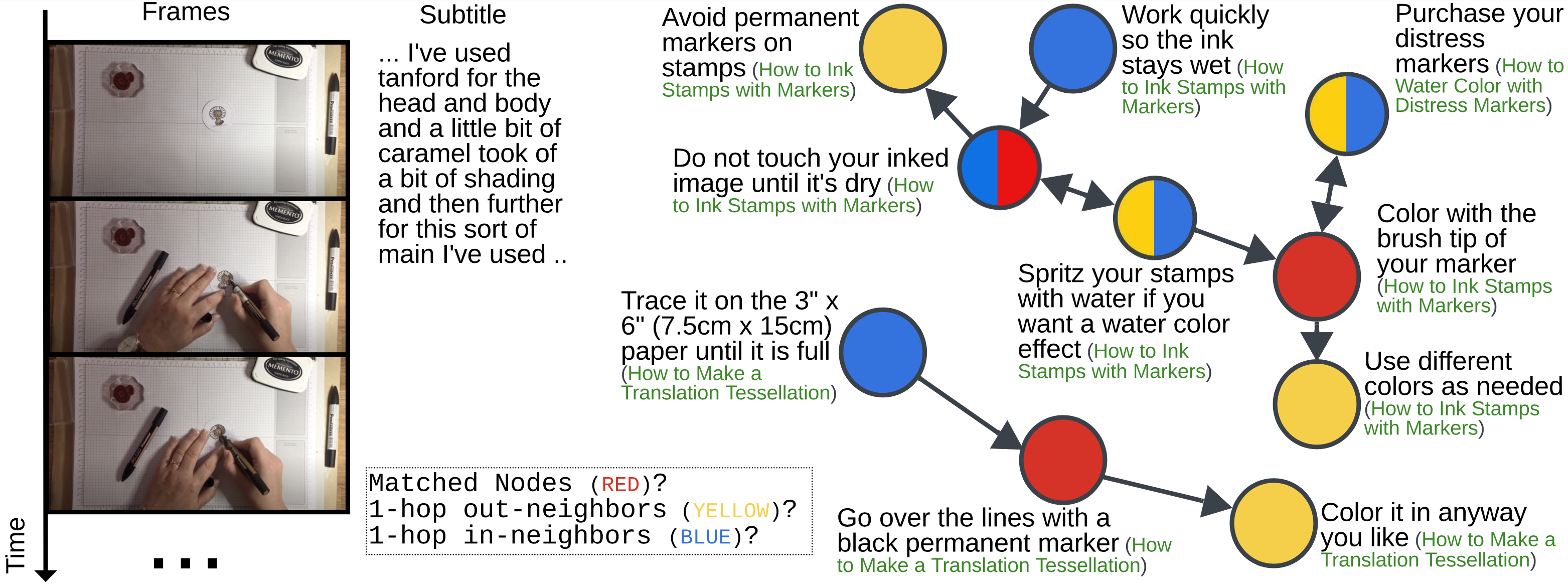}
	\caption{\textbf{The \g subgraph that a video segment belongs to.} 
	The \g subgraph entails a high relevance to the video segment (which is about coloring with a marker), and the graph structure encodes the 
	procedural knowledge of the general
	order and relation of steps from multiple tasks. 
}
\label{appendix_fig:qua_16818}
\end{figure*}

\begin{figure*}[thb]
	\centering
	\includegraphics[scale=0.31]{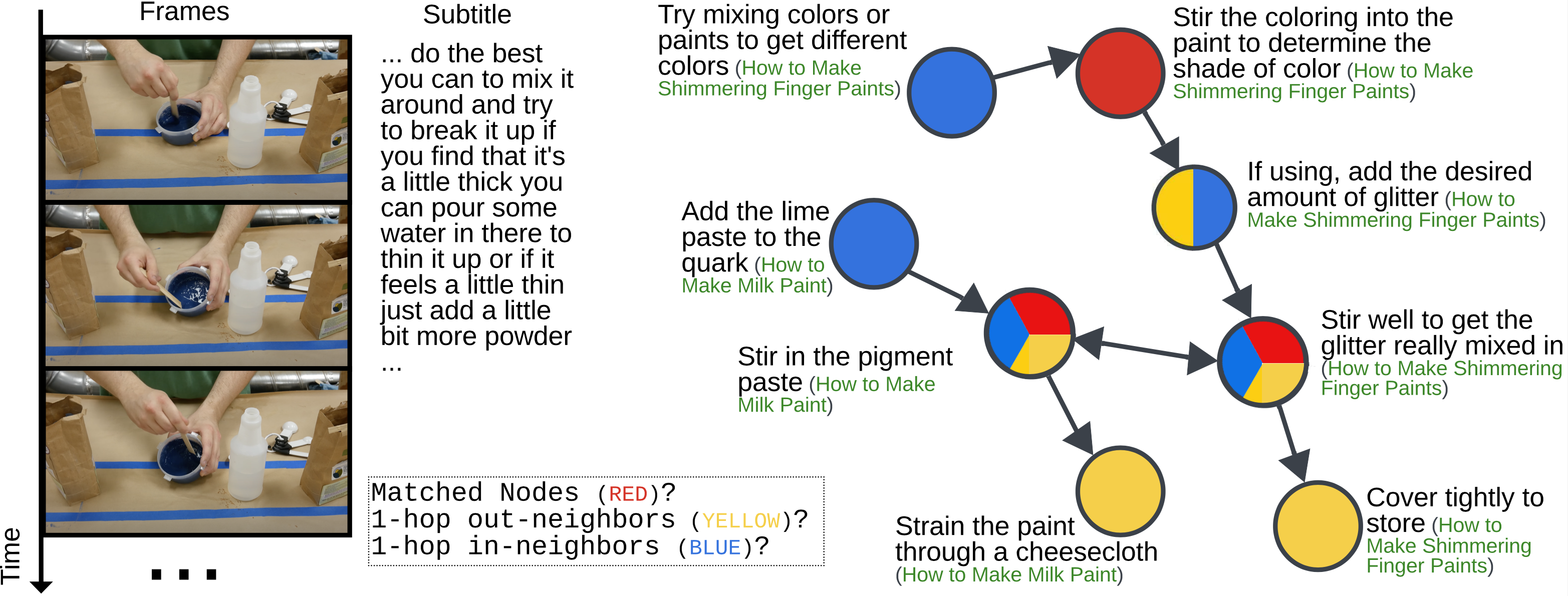}
	\caption{\textbf{The \g subgraph that a video segment belongs to.} 
    The matched step nodes of the \g are about ``stirring''/``mixing'' the ``pigment''/``paint'', which reflects the action in the video segment.  
	The \g subgraph entails a high relevance to the video segment, and the graph structure encodes the 
	procedural knowledge of the general
	order and relation of steps from multiple tasks. For example, the next step ($1$-hop out neighbor) can be ``cover tightly to store'' , and the steps of the \g subgraph come from the wikiHow tasks ``How to Make Milk Paint'' and ``How to Make Shimmering Finger Paints''.
}
\label{appendix_fig:qua_62797}
\end{figure*}

\begin{figure*}[thb]
	\centering
	\includegraphics[scale=0.31]{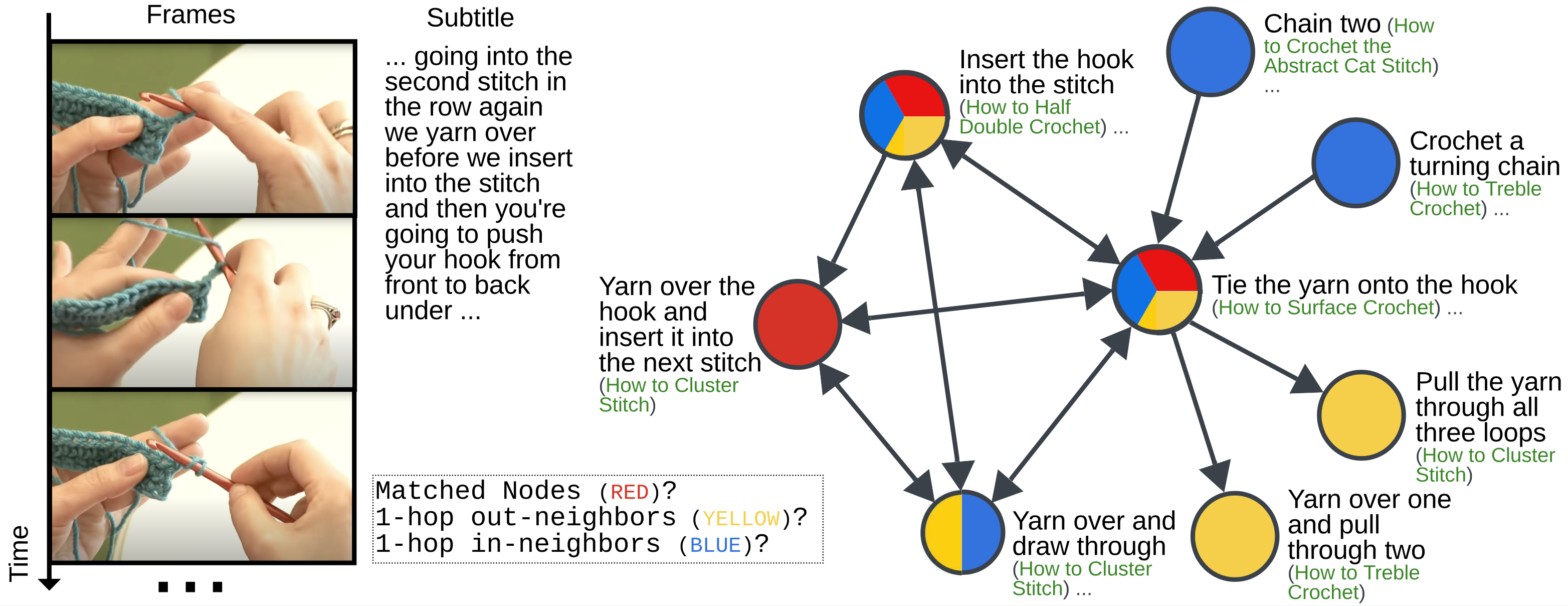}
	\caption{\textbf{The \g subgraph that a video segment belongs to.} 
    In this example, the left most four step nodes are quite densely connected, which might be attributed to the visually similar frames as well as complex semantic implication and subtle difference between step headlines of these step nodes.
    Overall the \g subgraph entails a high relevance to the video segment.
}
\label{appendix_fig:qua_1680}
\end{figure*}

\begin{figure*}[thb]
	\centering
	\includegraphics[scale=0.30]{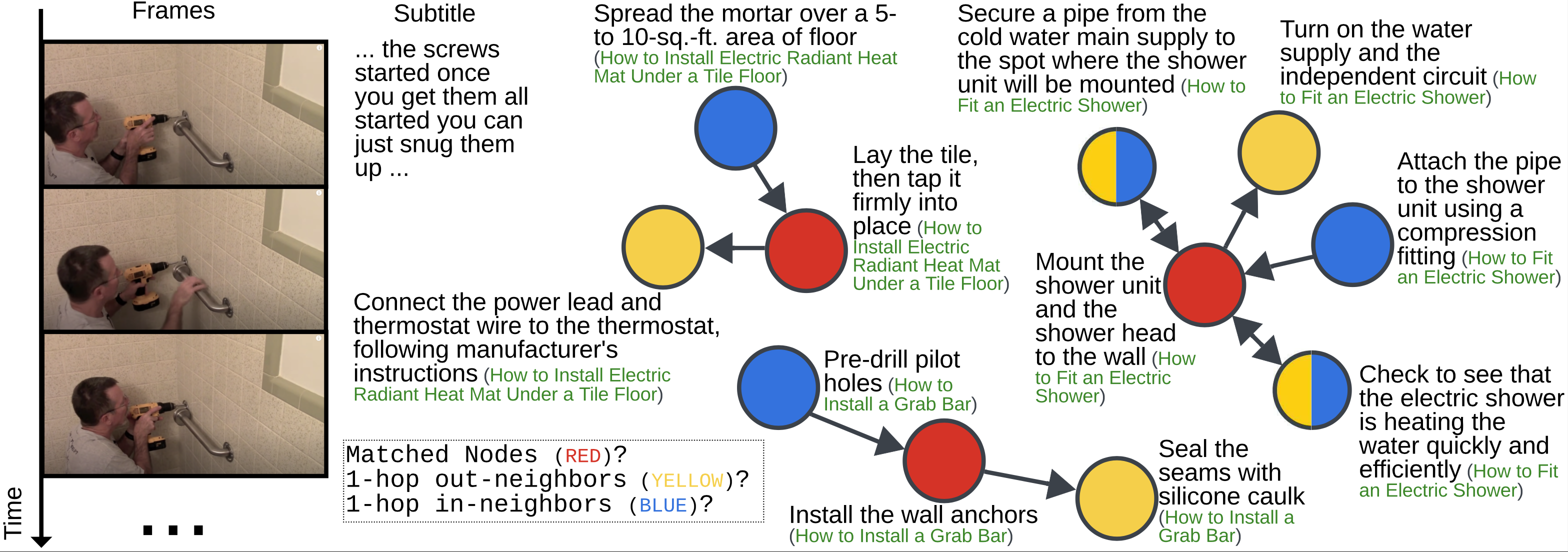}
	\caption{\textbf{The \g subgraph that a video segment belongs to.} 
    ``Install the wall anchors'' is the top $2$ matched node of this video segment, which well describes the step of the segment. The top $1$ matched node ``mount the shower unit and the shower head to the wall'' is also related to the shower room, but the objects ``shower unit'' and ``shower head'' are actually missing in the video segment. The top $3$ matched node ``lay the tile, then tap it firmly into place'' is mainly focusing on the object ``tile'' shown in the video segment. 
    Using a larger version of the wikiHow dataset and a stronger pre-trained multimodal video foundation model would lead to a even better quality of pseudo labels.
}
\label{appendix_fig:qua_2345}
\end{figure*}

\begin{figure*}[thb]
	\centering
	\includegraphics[scale=0.32]{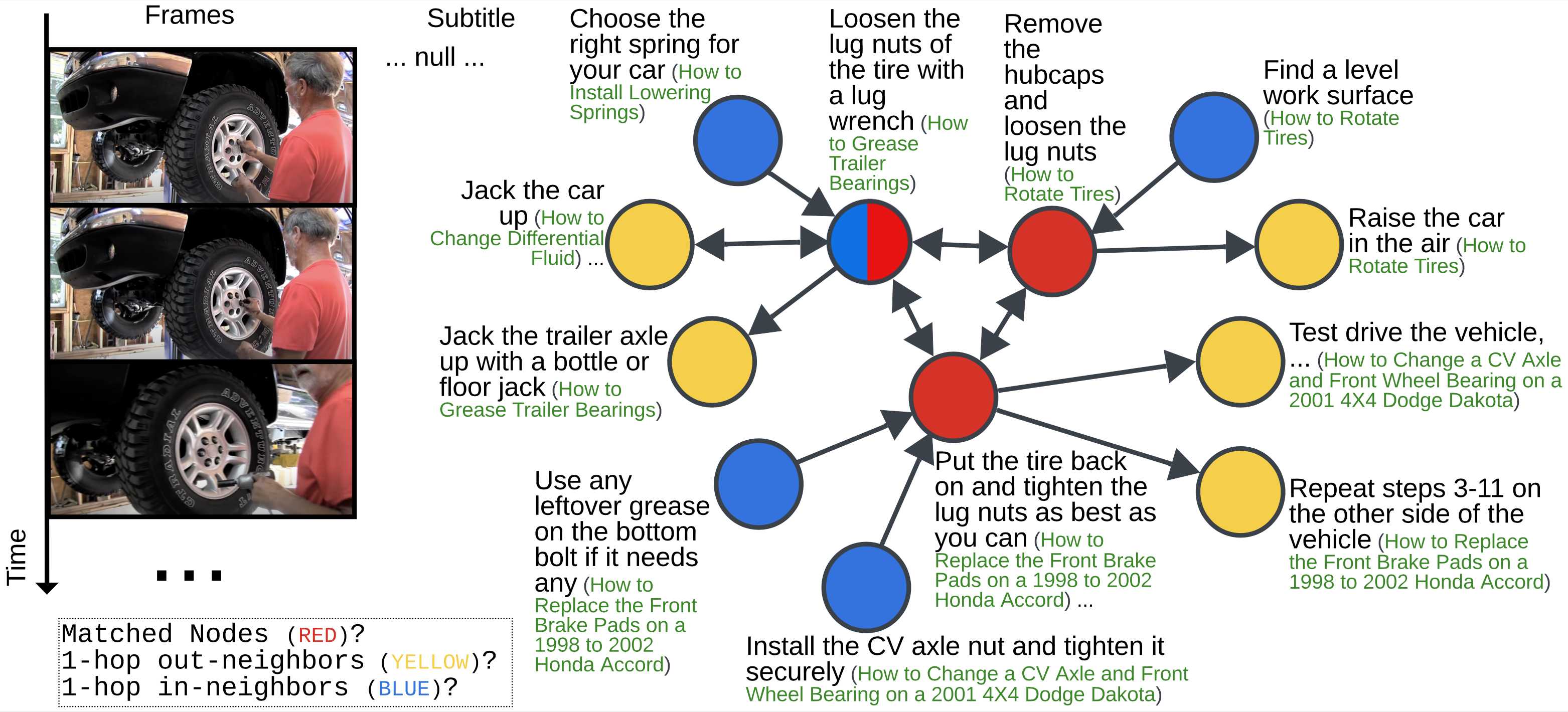}
	\caption{\textbf{The \g subgraph that a video segment belongs to.} 
    This \g subgraph fails to conform to the temporal signals of the video segment: 
    a man is ``tightening the lug nuts'' but two of the matched nodes are about ``loosening the lug nuts''.  In addition, the car has been jacked up in the video segment, but according to the \g subgraph, ``jack the car up'' is the next step (one of the top $5$ $1$-hop out-neighbors).
     The \g graph encodes the \textit{general} order and relation of steps -- the knowledge is not conditioned on a specific video segment. 
}
\label{appendix_fig:qua_1032}
\end{figure*}

\renewcommand{\thesection}{A}
 \section{Additional Results}
\subsection{More Qualitative Results}

\begin{figure*}[thb]
	\centering
	\includegraphics[scale=0.31]{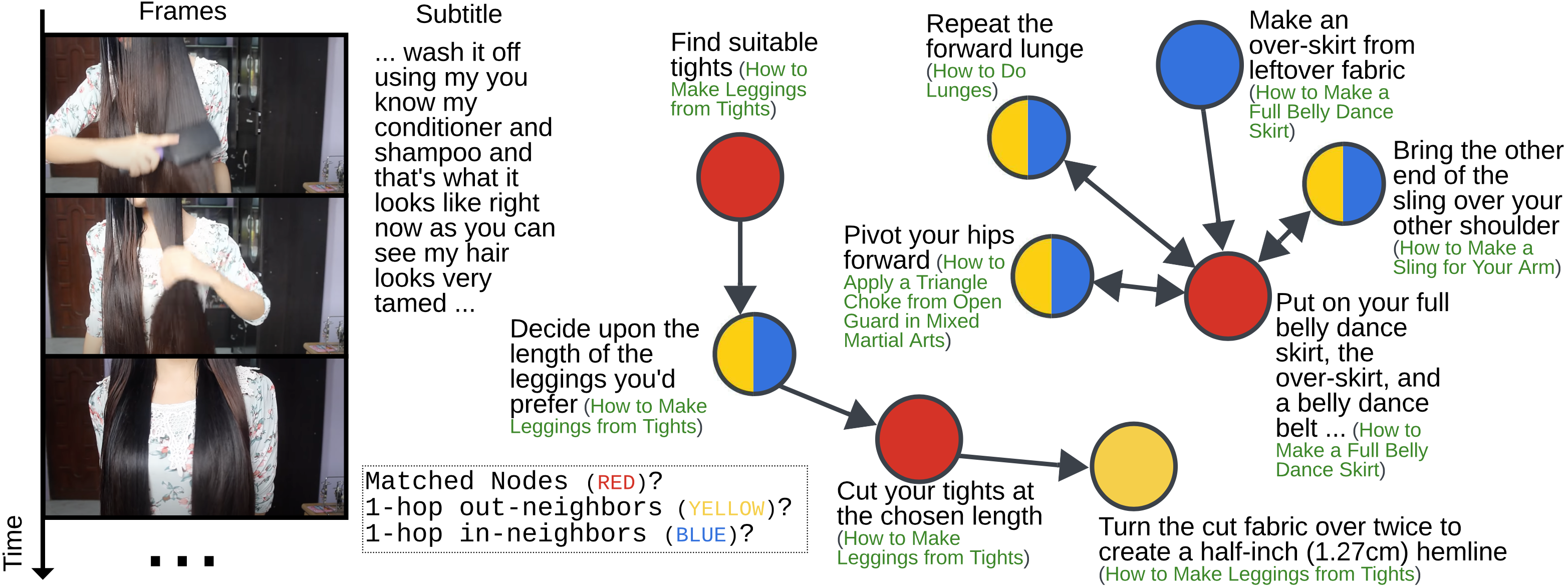}
	\caption{\textbf{The \g subgraph that a video segment belongs to.} 
The \g subgraph is not relevant to the video segment in this example because
the narrator is mentioning the step ``wash hair'', but the frames are not showing the step
``wash hair''. 
Subtitles can be useful when the action of the step is not demonstrated but only verbally described by the narrator.
}
\label{appendix_fig:qua_620069}
\end{figure*}

We present more qualitative results shown in \red{Fig.}~\ref{appendix_fig:qua_3515} $\sim$~\ref{appendix_fig:qua_620069}.
In these visualizations, we illustrate the \g subgraph of multiple video segments.
Specifically, we show the step nodes that are matched to the video segment,
a step headline member of the step node, the wikiHow task name of the step headline member (green fonts), the $1$-hop in- and out- neighbors of the matched step nodes -- the exact neighboring information used by our \ours model (\red{Table 1} in the main paper) to perform Node Relation Learning during pre-training, and the edge connections between these step nodes.
For a clear exposition, we did not plot nodes or edges of the \g that were not leveraged by \ours pre-training. 
In order to help the readers to have a concrete idea of the pseudo labels generated by the \geos, we list the complete pseudo labels of these video segments that were used by our \ours model pre-training
to perform Video-Node Matching, Video-Task Matching, Task Context Learning and Node Relation Learning in \red{Table~}\ref{tab:qua_3515_vnm} $\sim$~\ref{tab:qua_620069_tcl}.
For pseudo labels of Video-Node Matching and Node Relation Learning, node IDs are listed in the descending ranking order 
of their confidence scores in the tables.

As shown in the figures, the \g subgraph that a video segment belongs to entails a high relevance to the video segment, and the graph structure encodes the procedural knowledge of the general order and relation of steps from multiple tasks. 
Qualitative results also suggest that using a larger version of wikiHow dataset (e.g., \cite{zhang2020reasoning,zhang2020intent,yang2021visual,zhou2022show}) or stronger video and language encoders to either build the \g or generate pseudo labels would be beneficial (e.g., to obtain more relevant steps for video segments).

\begin{table}[t]
\footnotesize
\begin{center}
\begin{tabular}{ll}
\toprule
\multirow{6}{*}{Top 1 ($40$)} & Step Headline: Finished                                                               \\
           & Task: How to Make Pine Needle Tea                                                     \\ \cline{2-2} 
           & Step Headline: Finished                                                               \\
           & Task: How to Use a Bone Folder                                                        \\ \cline{2-2} 
           & Step Headline: Finished 
            \\ 
           & Task: How to Sew Hair Extensions to a Clip \\ \midrule
\multirow{6}{*}{Top 2 ($22$)}  & Step Headline: Disconnect the battery                                                 \\
           & Task: How to Install a Car Starter                                                    \\ \cline{2-2} 
           & Step Headline: Disconnect the battery cables                                        \\
           & Task: How to Change a Timing Chain                                                    \\ \cline{2-2} 
           & Step Headline: Disconnect the battery
           \\
           & Task: How to Change Radiator Fluid                                                    \\ \midrule
\multirow{6}{*}{Top 3 ($15$)} & Step Headline: Gather the necessary supplies                                                    \\
           & Task: How to Chlorinate a Well                                                     \\ \cline{2-2} 
           & Step Headline: Gather miscellaneous supplies                                                              \\
           & Task: How to Make a Wargaming Table                                                        \\ \cline{2-2} 
           & Step Headline: Gather your supplies
            \\ 
           & Task: How to Apply a Horse Tail Bandage \\ 
\bottomrule
\end{tabular}
\end{center}
\vspace{-8pt}
\scriptsize{Note: The number inside the bracket is the number of step headline members  of the step node. For each step node, we list $3$ random members. }
\caption{\textbf{The top $3$ largest step nodes of the \geos.}
Steps from different tasks may be described in the same or slightly different manner, but share the same semantic meaning. }
\label{tab:step_nodes}
\end{table}

\subsection{Results of \textbf{\ours}with 2 Hops}
In \red{Table 1} of the main paper, we report the results of \ours 
that uses the best setting of VNM, VTM and TCL from the ablation study,
but for NRL, $K=1$ instead of $2$.
This is because computing the NRL pseudo labels with $K=2$ is resource-wise expensive for the \textit{full} HowTo100M dataset due to
$51$M video segment samples (see \red{Sec.}~\ref{appendix_subsec:node_edge_construct}).

In \red{Table~\ref{tab:ours_2hop_fullresults}} of the Supplementary Material, we report results of \ours (VNM+VTM+TCL+NRL) that uses $K=2$ for NRL and the HowTo100M \textit{subset} as the pre-training data. $K=2$ for NRL has a 
better overall performance than $K=1$: on $8$ out of the $12$ evaluation settings, the performance of $K=2$ $\geq$ the performance of $K=1$ for \ours (VNM+VTM+TCL+NRL). Considering the trade-off between computation and performance, we recommend experimenting with a larger $K$ when the size of the training video data is small.

\begin{figure*}[thb]
	\centering
	\includegraphics[scale=0.31]{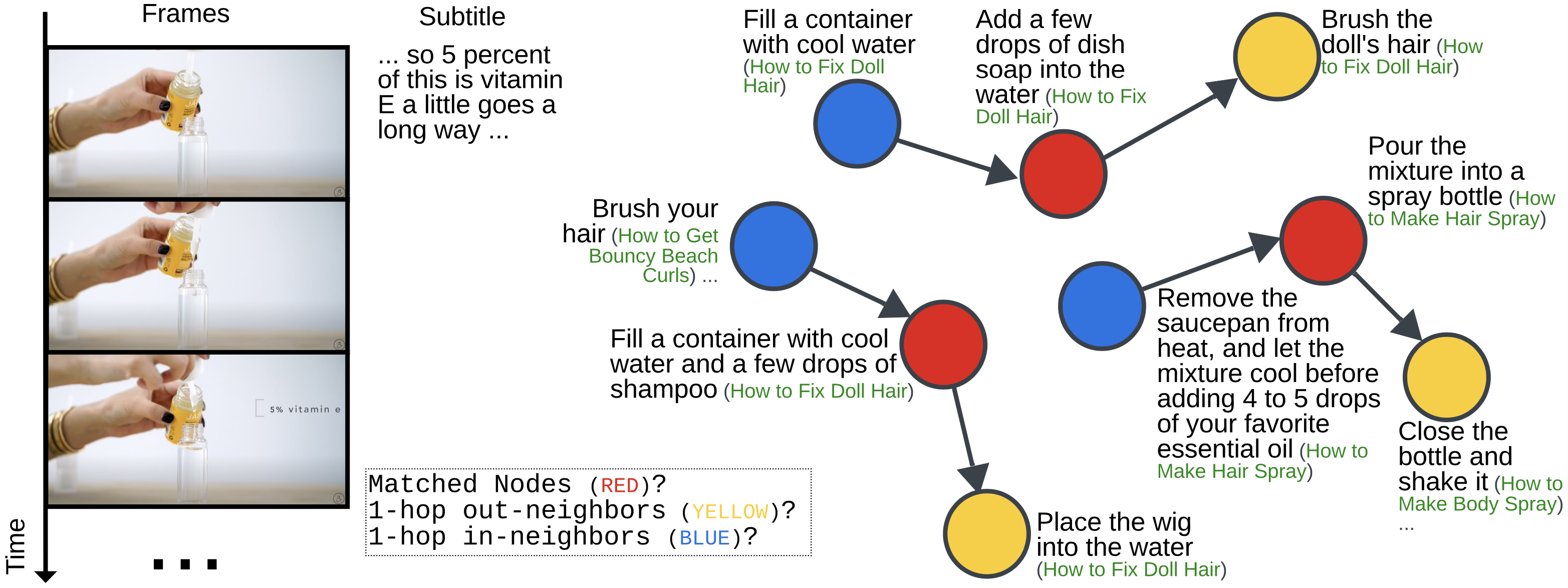}
	\caption{\textbf{The \g subgraph that a video segment belongs to.}  
   The step of the video segment is ``adding a few drops of Vitamin E oil''. ``Vitamin E'' is in the subtitle, but the \g subgraph fails to capture the ``Vitamin E'' information, because the visual frame signals were used to match the video segment to nodes. This example suggests subtitles can be useful in cases such as when the objects are small or hard to be recognized from frames.
}
\label{appendix_fig:qua_1748}
\end{figure*}

\subsection{Leveraging the \textbf{\g}for Downstream Tasks}
Prior work DS~\cite{distant} experimented with incorporating partial procedural knowledge at the downstream stage, i.e., at the inference time of $f(\cdot)$. Here, we conduct similar experiments to directly leverage the \g for the downstream model. We then compare the effectiveness of our proposed \g with $\mathbb{B}$ (i.e., wikiHow articles) 
for the downstream task of Step Forecasting on the COIN dataset. Below, we describe the details of the comparative experiments. The mathematical notations follow the ones
in the main paper.

\noindent \textbf{W/o Incorporating Knowledge Base}. 
Eq.~\ref{equa:downstream_no_knowledge} formulates the input sequence to the downstream Transformer model $\mathcal{T}$ without incorporating procedural knowledge from any knowledge database for the downstream Step Forecasting task (suppose historical steps contain $L$ segments):
{\small \begin{equation}
\mathcal{T}\left(f(e(x_1)), f(e(x_2)), \cdots, 
f(e(x_L))\right)
\label{equa:downstream_no_knowledge}
\end{equation}
}

\noindent \textbf{W/ Incorporating $\mathbb{B}$ (i.e., the method in DS~\cite{distant})}. 
In order to incorporate the knowledge retrieved from $\mathbb{B}$ for each segment into the input provided to the Transformer downstream model, a retrieval approach is firstly adopted to find for each segment $x_l$, the step headline from $\mathbb{B}$ that best explains the segment according to the pre-trained video model $f(\cdot)$.
Specifically, given the trained $f(\cdot)$ along with the trained segment-step matching classifier $a(\cdot)$, the retrieved step headline $\hat{s}_{i_{x_l}}^{(t_{x_l})}$ of segment $x_l$ is the one that corresponds to the step class ID yielding the maximum classification score according to the inference of $a(f(e(x_l)))$.
Then, $\mathcal{T}$ accepts a different input sequence as shown in Eq.~\ref{equa:downstream_use_wikihow}:
{\scriptsize \begin{equation}
\mathcal{T}\left(f(e(x_1)), m(\hat{s}_{i_{x_1}+1}^{(t_{x_1})}), f(e(x_2)), m(\hat{s}_{i_{x_2}+1}^{(t_{x_2})}) \cdots, f(e(x_L)), m(\hat{s}_{i_{x_L}+1}^{(t_{x_L})})\right)
\label{equa:downstream_use_wikihow}
\end{equation}
}
$m(\cdot)$ represents the feature extractor of step headlines. 

In other words, for each segment $x_l$, we obtain its representation using $f(e(x_l)))$. Here, the  model $f(\cdot)$ was trained using DS~\cite{distant} and $e(\cdot)$ is the MIL-NCE model~\cite{s3d}. The 
answer head $a(\cdot)$ predicts the most likely current step $\hat{s}_{i_{x_l}}^{(t_{x_l})}$. We then look up wikiHow article $t_{x_l}$ to find the next step $\hat{s}_{i_{x_l}+1}^{(t_{x_l})}$. We obtain the next step's headline feature produced by the step headline feature extractor $m(\cdot)$, and the next step's headline feature follows the representation of the segment $x_l$ in the input sequence to $\mathcal{T}$.

We call this downstream Transformer variant as ``Transformer w/ KB Transfer from $\mathbb{B}$''. Segment features produced from $f(\cdot)$ and the step headline features of retrieved next steps according to $a(f(\cdot))$ and $\mathbb{B}$, together they form the input sequence to the Transformer.

\begin{table}[h]
\footnotesize
  \aboverulesep=0ex
  \belowrulesep=0ex 
\begin{center}
\begin{tabular}{ll|c}
\toprule
Pre-training  & Downstream Method &  Accuracy   \\ \midrule
     DS~\cite{distant}         &   Transformer w/ KB Transfer from $\mathbb{B}$~\cite{distant}      &    $22.15$   \\ \hline
 \ours &   Transformer w/ KB Transfer from the \g    &   $\mathbf{35.46}$ \\ \bottomrule
\end{tabular}
\end{center}
\vspace{-15pt}
\caption{Results of Step Forecasting on COIN by directly 
incorporating the procedural knowledge database at the downstream stage.  }
\label{tab:downstream_with_knowledge}
\end{table}

\noindent \textbf{W/ Incorporating the \geos}~\cite{distant}. 
Similar to the baseline ``Transformer w/ KB Transfer from $\mathbb{B}$'' proposed by~\cite{distant}, we propose ``Transformer w/ KB Transfer from the \geos''.

For each segment $x_l$, we obtain its representation using $f(e(x_l)))$; here, $f(\cdot)$ is \ours pre-trained using our proposed objectives that leverage the \geos, and $e(\cdot)$ is the MIL-NCE model~\cite{s3d}.
The Video-Node Matching answer head $a(\cdot)$ predicts the step node $\hat{v}_{x_l}$ that is most likely to be matched to the video segment $x_l$.
We then look up the \g to obtain the $1$-hop out-neighboring nodes $\mathcal{N}(\hat{v}_{x_l})$ of the node $\hat{v}_{x_l}$.
Given the step headlines of $\mathcal{N}(\hat{v}_{x_l})$,
for each step headline, we obtain the step's headline feature produced by the step headline feature extractor $m(\cdot)$. The mean of these feature vectors is considered as the feature of the most likely next node, which follows the representation of the segment $x_l$ to form the input sequence to $\mathcal{T}$:
{\scriptsize \begin{equation}
\mathcal{T}\left(\cdots, f(e(x_l)), \frac{1}{|\mathcal{N}(\hat{v}_{x_l})|}\sum_{i \in \mathcal{N}(\hat{v}_{x_l})} \left(\frac{1}{|\mathcal{S}(i)|}\sum_{j \in \mathcal{S}(i)} m(j)\right), \cdots \right)
\label{equa:downstream_use_pkg}
\end{equation}
}
where $\mathcal{S}(i)$ denotes the set of step headline members of node $i$.
Therefore,
\ours produced segment features and features of the \g retrieved next nodes, together they form the input sequence to the Transformer.

Results are presented in Tab.~\ref{tab:downstream_with_knowledge}. We observe a performance drop after integrating procedural knowledge into the downstream models.
This suggests a future
research direction towards effectively integrating procedural knowledge into procedure-understanding-related downstream model training and/or inference.
However, 
\ours still outperforms the prior work DS~\cite{distant} under this setting, which further reinforces our key insight that using procedural information during pre-training is beneficial. 
We highlight a crucial difference between our work and DS~\cite{distant}: DS uses partial procedural knowledge in the downstream task, while we are the first to show that using graph-structured procedural knowledge in \textit{pre-training} is potentially beneficial for \textit{any} procedure understanding downstream task.

\begin{table*}[ht]
\setlength{\tabcolsep}{2.8pt}
\footnotesize
  \aboverulesep=0ex
  \belowrulesep=0ex 
\begin{center}
\begin{tabular}{l|lll|lll|lll|lll}
\toprule
\multicolumn{1}{l|}{\multirow{3}{*}{\textbf{Pre-training Method}}}      & \multicolumn{6}{c|}{\textbf{Downstream Transformer}}                                                                                                                             & \multicolumn{6}{c}{\textbf{Downstream MLP}}                                                                                                                     \\ \cline{2-13} 
\multicolumn{1}{l|}{}                            & \multicolumn{3}{c|}{\textbf{COIN}}                                                 & \multicolumn{3}{c|}{\textbf{CrossTask}}                                            & \multicolumn{3}{c|}{\textbf{COIN}}                                                 & \multicolumn{3}{c}{\textbf{CrossTask}}                                            \\
\multicolumn{1}{l|}{}                            & \multicolumn{1}{c}{\textbf{SF}} & \multicolumn{1}{c}{\textbf{SR}} & \multicolumn{1}{c|}{\textbf{TR}} & \multicolumn{1}{c}{\textbf{SF}} & \multicolumn{1}{c}{\textbf{SR}} & \multicolumn{1}{c|}{\textbf{TR}} & \multicolumn{1}{c}{\textbf{SF}} & \multicolumn{1}{c}{\textbf{SR}} & \multicolumn{1}{c|}{\textbf{TR}} & \multicolumn{1}{c}{\textbf{SF}} & \multicolumn{1}{c}{\textbf{SR}} & \multicolumn{1}{c}{\textbf{TR}} \\
\midrule 

\multicolumn{1}{l|}{\ours (VNM + VTM + TCL + NRL)}    & $\mathbf{42.91}$  & $\mathbf{50.62}$  &  $85.02$    &   $61.29$  &    $\mathbf{62.48}$  &  $\mathbf{67.41}$ & $39.75$   & $44.63$   &  $\mathbf{83.88}$ &   $\mathbf{59.53}$ &  $\mathbf{60.76}$   &    $\mathbf{67.51}$  \\ 
\bottomrule
\end{tabular}
\end{center}
\vspace{-8pt}
\scriptsize{Note: 
\textbf{SF}: Step Forecasting; \textbf{SR}: Step Recognition; \textbf{TR}: Task Recognition.\\
}
\vspace{-10pt}
\caption{Downstream procedure understanding evaluation results of \ours (VNM + VTM + TCL + NRL) \textbf{when $\mathbf{K=2}$ for NRL}. Bolded ones are the cases where the accuracy of $K=2$ $\geq$ the accuracy of $K=1$ for \ours (VNM + VTM + TCL + NRL). 
}
\label{tab:ours_2hop_fullresults}
\end{table*}

\begin{table*}[t]
\footnotesize
\begin{center}
\begin{tabular}{ll}
\toprule
 \textbf{Node ID} &\textbf{Step Headline} \\ \midrule
 \multirow{1}{*}{1982} & Finish the edge by carefully folding the top edge under to the bottom side and using the heating iron to seal it in place. \\ \hline
 \multirow{1}{*}{443}   & Fold and iron the top hem of the front pocket. \\\hline
  \multirow{1}{*}{1655}    & Iron the creases and pin carefully or baste. \\ 
\bottomrule
\end{tabular}
\end{center}
\vspace{-10pt}
\caption{Pseudo labels of \textbf{Video-Node Matching} produced by the \g for the video segment shown in \red{Fig.~\ref{appendix_fig:qua_3515}}.}
\label{tab:qua_3515_vnm}
\end{table*}

\begin{table*}[t]
\footnotesize
\begin{center}
\begin{tabular}{l}
\toprule
 \textbf{Task Name} \\ \midrule
 How to Apply MonoKote \\ 
 How to Make a Pen Pocket for Your Journal \\
 How to Make a Mei Tai Baby Carrier \\
\bottomrule
\end{tabular}
\end{center}
\vspace{-10pt}
\caption{Pseudo labels of \textbf{Video-Task Matching} (using wikiHow task names) produced by the \g for the video segment shown in \red{Fig.~\ref{appendix_fig:qua_3515}}.}
\label{tab:qua_2515_vtm}
\end{table*}

\begin{table*}[t]
\footnotesize
  \aboverulesep=0ex
  \belowrulesep=0ex 
\begin{center}
\begin{tabularx}{\linewidth}{l|lX}
\toprule
\textbf{Relation Type} & \textbf{Node ID} &\textbf{Step Headline} \\ \midrule
\multirow{6}{*}{1 Hop In-Neighbor} & \multirow{1}{*}{3139} & Trim the edges carefully with the utility knife, leaving a 1/8 inch to 1/4 inch (3 to 6 mm) overlap. \\ \cline{2-3}
 & \multirow{1}{*}{591}   & Cut your elastic 1 inch (2.54 centimeters) longer than your body pieces. \\ \cline{2-3}
 &  \multirow{1}{*}{1255}   & Place the raw edges of the square into the opening in the ties and pin in place. \\ \cline{2-3}
    &  \multirow{1}{*}{1655}  & Iron the creases and pin carefully or baste. \\  \cline{2-3} 
      & \multirow{1}{*}{1982} & Finish the edge by carefully folding the top edge under to the bottom side and using the heating iron to seal it in place. \\  \hline
 \multirow{6}{*}{1 Hop Out-Neighbor} & \multirow{1}{*}{1087} & Top stitch the hem down. \\  \cline{2-3}
 & \multirow{1}{*}{877}  & Top stitch the edges of the ties, taking care to sew multiple seams through ALL layers when sewing the raw edges of your square inside the ties.  \\  \cline{2-3}
 & \multirow{1}{*}{1982}  & Finish the edge by carefully folding the top edge under to the bottom side and using the heating iron to seal it in place. \\  \cline{2-3}
    &  \multirow{1}{*}{1655}  & Iron the creases and pin carefully or baste. \\   \cline{2-3}
 & \multirow{1}{*}{1006}  & Press the lengthwise seam open. \\ \hline
  \multirow{4}{*}{2 Hops In-Neighbor}   & \multirow{1}{*}{2469} & Unroll a second sheet of MonoKote, and repeat the process on the other side of the aircraft part.  \\ \cline{2-3}
 & \multirow{1}{*}{782}  & Cut a third piece of for the front pocket. \\ \cline{2-3}
   & \multirow{1}{*}{1744} & Repeat measuring, marking and sewing steps for the second tie. \\ 
 \hline
 \multirow{4}{*}{2 Hops Out-Neighbor} & \multirow{1}{*}{233} & Pin the pocket to the front of one of the larger body pieces.\\ \cline{2-3}
 & \multirow{1}{*}{6588} & The pouch is now ready to wear.  \\ \cline{2-3}
  & \multirow{1}{*}{1982} & Finish the edge by carefully folding the top edge under to the bottom side and using the heating iron to seal it in place.  \\  \cline{2-3}
\bottomrule
\end{tabularx}
\end{center}
\vspace{-10pt}
\caption{Pseudo labels of \textbf{Node Relation Learning} produced by the \g for the video segment shown in \red{Fig.~\ref{appendix_fig:qua_3515}}.}
\label{tab:nrl_qua_3515}
\end{table*}

\begin{table*}[t]
\footnotesize
\begin{center}
\begin{tabular}{ll}
\toprule
 \textbf{Node ID} &\textbf{Step Headline} \\ \midrule
 \multirow{1}{*}{2811} & Color with the brush tip of your marker. \\ \hline
 \multirow{1}{*}{8988}   & Do not touch your inked image until it's dry.\\\hline
  \multirow{1}{*}{9399}    & Go over the lines with a black permanent marker. \\ 
\bottomrule
\end{tabular}
\end{center}
\vspace{-10pt}
\caption{Pseudo labels of \textbf{Video-Node Matching} produced by the \g for the video segment shown in \red{Fig.~\ref{appendix_fig:qua_16818}}.}
\label{tab:qua_16818_vnm}
\end{table*}

\begin{table*}[t]
\footnotesize
\begin{center}
\begin{tabular}{l}
\toprule
 \textbf{Task Name} \\ \midrule 
How to Ink Stamps with Markers \\
How to Make a Translation Tessellation \\
\bottomrule
\end{tabular}
\end{center}
\vspace{-10pt}
\caption{Pseudo labels of \textbf{Video-Task Matching} (using wikiHow task names) produced by the \g for the video segment shown in \red{Fig.~\ref{appendix_fig:qua_16818}}.}
\label{tab:qua_16818_vtm}
\end{table*}

\begin{table*}[t]
\footnotesize
  \aboverulesep=0ex
  \belowrulesep=0ex 
\begin{center}
\begin{tabular}{l|ll}
\toprule
\textbf{Relation Type} & \textbf{Node ID} &\textbf{Step Headline} \\ \midrule
\multirow{5}{*}{1 Hop In-Neighbor} & \multirow{1}{*}{8361} & Spritz your stamps with water if you want a water color effect. \\ \cline{2-3}
 & \multirow{1}{*}{4496}   &  Work quickly so the ink stays wet. \\ \cline{2-3}
 &  \multirow{1}{*}{5543}   & Trace it on the 3" x 6" (7.5cm x 15cm) paper until it is full. \\ \cline{2-3}
    &  \multirow{1}{*}{3201}  & Purchase your distress markers. \\  \cline{2-3} 
      & \multirow{1}{*}{8988} & Do not touch your inked image until it's dry. \\  \hline
 \multirow{5}{*}{1 Hop Out-Neighbor} & \multirow{1}{*}{8130} & Use different colors as needed. \\  \cline{2-3}
 & \multirow{1}{*}{8589}  & Avoid permanent markers on stamps.  \\  \cline{2-3}
 & \multirow{1}{*}{8492}  & Color it in anyway you like. \\  \cline{2-3}
    &  \multirow{1}{*}{3201}  & Purchase your distress markers. \\   \cline{2-3}
 & \multirow{1}{*}{8361}  & Spritz your stamps with water if you want a water color effect. \\ \hline
  \multirow{3}{*}{2 Hops In-Neighbor}   & \multirow{1}{*}{8141} & Sand your stamp lightly.  \\ \cline{2-3}
 & \multirow{1}{*}{2985}  & Experiment a little. \\ \cline{2-3}
   & \multirow{1}{*}{8591} & Convert this base tessellation into a more interesting shape. \\ \cline{2-3}
 \hline
 \multirow{3}{*}{2 Hops Out-Neighbor} & \multirow{1}{*}{3336} & Breathe on the stamp to moisten the ink. \\ \cline{2-3}
 & \multirow{1}{*}{6068} & Ink it all over.  \\ \cline{2-3}
  & \multirow{1}{*}{2811} & Color with the brush tip of your marker.  \\  \cline{2-3}
\bottomrule
\end{tabular}
\end{center}
\vspace{-10pt}
\caption{Pseudo labels of \textbf{Node Relation Learning} produced by the \g for the video segment shown in \red{Fig.~\ref{appendix_fig:qua_16818}}.}
\label{tab:nrl_qua_16818}
\end{table*}

\begin{table*}[t]
\footnotesize
\begin{center}
\begin{tabular}{ll}
\toprule
 \textbf{Node ID} &\textbf{Step Headline} \\ \midrule
 \multirow{1}{*}{2374} & Stir in the pigment paste. \\ \hline
 \multirow{1}{*}{6937}   & Stir well to get the glitter really mixed in. \\\hline
  \multirow{1}{*}{9026}    & Stir the coloring into the paint to determine the shade of color.\\ 
\bottomrule
\end{tabular}
\end{center}
\vspace{-10pt}
\caption{Pseudo labels of \textbf{Video-Node Matching} produced by the \g for the video segment shown in \red{Fig.~\ref{appendix_fig:qua_62797}}.}
\label{tab:qua_62797_vnm}
\end{table*}

\begin{table*}[t]
\footnotesize
\begin{center}
\begin{tabular}{l}
\toprule
 \textbf{Task Name} \\ \midrule
How to Make Milk Paint \\
How to Make Shimmering Finger Paints \\
\bottomrule
\end{tabular}
\end{center}
\vspace{-10pt}
\caption{Pseudo labels of \textbf{Video-Task Matching} (using wikiHow task names) produced by the \g for the video segment shown in \red{Fig.~\ref{appendix_fig:qua_62797}}.}
\label{tab:qua_2515_vtm}
\end{table*}

\begin{table*}[t]
\footnotesize
  \aboverulesep=0ex
  \belowrulesep=0ex 
\begin{center}
\begin{tabular}{l|ll}
\toprule
\textbf{Relation Type} & \textbf{Node ID} &\textbf{Step Headline} \\ \midrule
\multirow{5}{*}{1 Hop In-Neighbor} & \multirow{1}{*}{2459} & Add the lime paste to the quark. \\ \cline{2-3}
 & \multirow{1}{*}{7885}   & If using, add the desired amount of glitter. \\ \cline{2-3}
 &  \multirow{1}{*}{9377}   & Try mixing colors or paints to get different colors. \\ \cline{2-3}
    &  \multirow{1}{*}{6937}  & Stir well to get the glitter really mixed in. \\  \cline{2-3} 
      & \multirow{1}{*}{2374} & Stir in the pigment paste. \\  \hline
 \multirow{5}{*}{1 Hop Out-Neighbor} & \multirow{1}{*}{8226} & Strain the paint through a cheesecloth. \\  \cline{2-3}
 & \multirow{1}{*}{5149}  & Cover tightly to store.  \\  \cline{2-3}
 & \multirow{1}{*}{7885}  & If using, add the desired amount of glitter. \\  \cline{2-3}
    &  \multirow{1}{*}{2374}  & Stir in the pigment paste. \\   \cline{2-3}
 & \multirow{1}{*}{6937}  & Stir well to get the glitter really mixed in.  \\ \hline
  \multirow{3}{*}{2 Hops In-Neighbor}   & \multirow{1}{*}{5349} & Transfer the quark into a paint bucket. \\ \cline{2-3}
 & \multirow{1}{*}{9026}  & Stir the coloring into the paint to determine the shade of color. \\ \cline{2-3}
   & \multirow{1}{*}{8702} & Add a few drops of food coloring to each container. \\ \cline{2-3}
 \hline
 \multirow{3}{*}{2 Hops Out-Neighbor} & \multirow{1}{*}{4614} & Prepare the surface you are painting. \\ \cline{2-3}
 & \multirow{1}{*}{6937} & Stir well to get the glitter really mixed in.  \\ \cline{2-3}
  & \multirow{1}{*}{2374} & Stir in the pigment paste. \\  
\bottomrule
\end{tabular}
\end{center}
\vspace{-10pt}
\caption{Pseudo labels of \textbf{Node Relation Learning} produced by the \g for the video segment shown in \red{Fig.~\ref{appendix_fig:qua_62797}}.}
\label{tab:nrl_qua_62797}
\end{table*}

\begin{table*}[t]
\footnotesize
\begin{center}
\begin{tabular}{ll}
\toprule
 \textbf{Node ID} &\textbf{ Step Headline} \\ \midrule
 \multirow{1}{*}{10016} & Yarn over the hook and insert it into the next stitch. \\ \hline
 \multirow{5}{*}{67}   & Tie the yarn onto the hook. \\
     & Tie the yarn onto your hook. \\
     & Yarn over the hook. \\ 
     & Yarn over and insert the hook. \\
     & Yarn over twice and insert hook. \\  \hline
 \multirow{4}{*}{37}   & Insert the hook into the stitch. \\
     & Insert the hook into the first stitch. \\
     & Insert the hook into the first stitch of the group. \\ 
     & Insert the hook into the next stitch. \\
\bottomrule
\end{tabular}
\end{center}
\vspace{-10pt}
\caption{Pseudo labels of \textbf{Video-Node Matching} produced by the \g for the video segment shown in \red{Fig.~\ref{appendix_fig:qua_1680}}.}
\label{tab:qua_1680_vnm}
\end{table*}

\begin{table*}[t]
\footnotesize
\begin{center}
\begin{tabular}{l}
\toprule
 \textbf{Task Name} \\ \midrule
 How to Cluster Stitch \\ 
 How to Crochet Popcorn Stitch \\
 How to Surface Crochet \\
 How to Half Double Crochet \\
 How to Treble Crochet \\
\bottomrule
\end{tabular}
\end{center}
\vspace{-10pt}
\caption{Pseudo labels of \textbf{Video-Task Matching} (using wikiHow task names) produced by the \g for the video segment shown in \red{Fig.~\ref{appendix_fig:qua_1680}}.}
\label{tab:qua_1680_vtm}
\end{table*}

\begin{table*}[t]
\footnotesize
  \aboverulesep=0ex
  \belowrulesep=0ex 
\begin{center}
\begin{tabular}{l|ll}
\toprule
\textbf{Relation Type} & \textbf{Node ID} &\textbf{Step Headline} \\ \midrule
\multirow{13}{*}{1 Hop In-Neighbor} & \multirow{2}{*}{493} & Yarn over and draw through. \\ 
 &  & Yarn over and draw through twice. \\ \cline{2-3}
 & \multirow{4}{*}{37}   & Insert the hook into the stitch. \\
  &    & Insert the hook into the first stitch. \\
 &     & Insert the hook into the first stitch of the group. \\ 
  &    & Insert the hook into the next stitch. \\  \cline{2-3}
 &  \multirow{5}{*}{67}   & Tie the yarn onto the hook. \\
  &    & Tie the yarn onto your hook. \\
  &    & Yarn over the hook. \\ 
  &    & Yarn over and insert the hook. \\
  &    & Yarn over twice and insert hook. \\  \cline{2-3}
  &  \multirow{1}{*}{418}  & Chain two.  \\   \cline{2-3}
  & \multirow{1}{*}{180} & Crochet a turning chain. \\  \hline
 \multirow{13}{*}{1 Hop Out-Neighbor} & \multirow{2}{*}{493} & Yarn over and draw through. \\ 
 &  & Yarn over and draw through twice. \\ \cline{2-3}
 & \multirow{4}{*}{37}  & Insert the hook into the first stitch.  \\
 &  &  Insert the hook into the next stitch. \\
 &  & Insert the hook into the first stitch of the group.  \\
 &  & Insert the hook into the stitch. \\ \cline{2-3}
 & \multirow{5}{*}{67}  & Tie the yarn onto the hook.  \\ 
 &  & Tie the yarn onto your hook. \\
 &  & Yarn over the hook. \\ 
 &  & Yarn over and insert the hook. \\ 
 &  & Yarn over twice and insert hook. \\ \cline{2-3}
 & \multirow{1}{*}{4836}  & Pull the yarn through all three loops. \\ \cline{2-3}
 & \multirow{1}{*}{4922}  & Yarn over one and pull through two. \\ \hline
  \multirow{4}{*}{2 Hops In-Neighbors}  & \multirow{2}{*}{493} & Yarn over and draw through. \\ 
 &  & Yarn over and draw through twice. \\ \cline{2-3}
 & \multirow{1}{*}{403} & Create a foundation chain. \\ \cline{2-3}
 & \multirow{1}{*}{4836}  & Pull the yarn through all three loops. \\ \cline{2-3}
\hline
 \multirow{7}{*}{2 Hops Out-Neighbor} & \multirow{2}{*}{493} & Yarn over and draw through. \\ 
 &  & Yarn over and draw through twice. \\ \cline{2-3}
 & \multirow{1}{*}{8289} & Insert the hook into the next spot.  \\ \cline{2-3}
 & \multirow{4}{*}{17} & Yarn over and pull through twice.\\
 &  & Yarn over and pull through. \\
 &  & Wrap the yarn around and pull through twice. \\
 &  & Loop the yarn over and pull through. \\ 
\bottomrule
\end{tabular}
\end{center}
\vspace{-10pt}
\caption{Pseudo labels of \textbf{Node Relation Learning} produced by the \g for the video segment shown in \red{Fig.~\ref{appendix_fig:qua_1680}}.}
\label{tab:nrl_qua_1680}
\end{table*}

\begin{table*}[t]
\footnotesize
\begin{center}
\begin{tabular}{ll}
\toprule
 \textbf{Node ID} &\textbf{Step Headline} \\ \midrule
 \multirow{1}{*}{3765} & Mount the shower unit and the shower head to the wall. \\ \hline
 \multirow{1}{*}{3616}   & Install the wall anchors. \\\hline
  \multirow{1}{*}{2823}    & Lay the tile, then tap it firmly into place. \\ 
\bottomrule
\end{tabular}
\end{center}
\vspace{-10pt}
\caption{Pseudo labels of \textbf{Video-Node Matching} produced by the \g for the video segment shown in \red{Fig.~\ref{appendix_fig:qua_2345}}.}
\label{tab:qua_2345_vnm}
\end{table*}

\begin{table*}[t]
\footnotesize
\begin{center}
\begin{tabular}{l}
\toprule
 \textbf{Task Name} \\ \midrule
 How to Fit an Electric Shower \\ 
 How to Install a Grab Bar \\
 How to Install Electric Radiant Heat Mat Under a Tile Floor \\
\bottomrule
\end{tabular}
\end{center}
\vspace{-10pt}
\caption{Pseudo labels of \textbf{Video-Task Matching} (using wikiHow task names) produced by the \g for the video segment shown in \red{Fig.~\ref{appendix_fig:qua_2345}}.}
\label{tab:qua_2515_vtm}
\end{table*}

\begin{table*}[t]
\footnotesize
  \aboverulesep=0ex
  \belowrulesep=0ex 
\begin{center}
\begin{tabularx}{\linewidth}{l|lX}
\toprule
\textbf{Relation Type} & \textbf{Node ID} &\textbf{Step Headline} \\ \midrule
\multirow{5}{*}{1 Hop In-Neighbor} & \multirow{1}{*}{8332} & Attach the pipe to the shower unit using a compression fitting. \\ \cline{2-3}
 & \multirow{1}{*}{6772}   & Pre-drill pilot holes. \\ \cline{2-3}
 &  \multirow{1}{*}{1057}   & Spread the mortar over a 5- to 10-sq.-ft. area of floor. \\ \cline{2-3}
    &  \multirow{1}{*}{2931}  & Check to see that the electric shower is heating the water quickly and efficiently. \\  \cline{2-3} 
      & \multirow{1}{*}{8721} & Secure a pipe from the cold water main supply to the spot where the shower unit will be mounted. \\  \hline
 \multirow{5}{*}{1 Hop Out-Neighbor} & \multirow{1}{*}{4498} & Turn on the water supply and the independent circuit. \\  \cline{2-3}
 & \multirow{1}{*}{1846}  & Seal the seams with silicone caulk.  \\  \cline{2-3}
 & \multirow{1}{*}{837}  & Connect the power lead and thermostat wire to the thermostat, following manufacturer's instructions. \\  \cline{2-3}
    &  \multirow{1}{*}{2931}  & Check to see that the electric shower is heating the water quickly and efficiently. \\   \cline{2-3}
 & \multirow{1}{*}{8721}  & Secure a pipe from the cold water main supply to the spot where the shower unit will be mounted.\\ \hline
  \multirow{3}{*}{2 Hops In-Neighbor}   & \multirow{1}{*}{4735} & Attach a non-return valve or stop tap to the pipe to isolate the shower's water supply from the rest of the building. \\ \cline{2-3}
 & \multirow{1}{*}{6528}  & Mark the location of the stud.\\ \cline{2-3}
   & \multirow{1}{*}{1725} & Install conduit connectors to both ends of two pieces of 1/2-in. \\ \cline{2-3}
 \hline
 \multirow{3}{*}{2 Hops Out-Neighbor} & \multirow{1}{*}{2931} & Check to see that the electric shower is heating the water quickly and efficiently. \\ \cline{2-3}
 & \multirow{1}{*}{3147} & Test at the end by pulling on it.  \\ \cline{2-3}
  & \multirow{1}{*}{8894} & Connect the stop tee to the fill valve.  \\  \cline{2-3}
\bottomrule
\end{tabularx}
\end{center}
\vspace{-10pt}
\caption{Pseudo labels of \textbf{Node Relation Learning} produced by the \g for the video segment shown in \red{Fig.~\ref{appendix_fig:qua_2345}}.}
\label{tab:nrl_qua_2345}
\end{table*}

\begin{table*}[t]
\footnotesize
\begin{center}
\begin{tabular}{ll}
\toprule
 \textbf{Node ID} &\textbf{Step Headline} \\ \midrule
 \multirow{1}{*}{599} & Flip the paper to the front/printed side and use back light to check the placement. \\ \hline
 \multirow{2}{*}{68}   & Put the tire back on and tighten the lug nuts as best as you can. \\
  & Put your tire back on the truck and tighten the lug nuts. \\ \hline
  \multirow{1}{*}{9667}    & Remove the hubcaps and loosen the lug nuts. \\ 
\bottomrule
\end{tabular}
\end{center}
\vspace{-10pt}
\caption{Pseudo labels of \textbf{Video-Node Matching} produced by the \g for the video segment shown in \red{Fig.~\ref{appendix_fig:qua_1032}}.}
\label{tab:qua_1032_vnm}
\end{table*}

\begin{table*}[t]
\footnotesize
\begin{center}
\begin{tabular}{l}
\toprule
 \textbf{Task Name} \\ \midrule
How to Foundation Piece a Quilt Block \\ 
How to Replace the Front Brake Pads on a 1998 to 2002 Honda Accord \\
How to Change a CV Axle and Front Wheel Bearing on a 2001 4X4 Dodge Dakota \\
How to Rotate Tires \\
\bottomrule
\end{tabular}
\end{center}
\vspace{-10pt}
\caption{Pseudo labels of \textbf{Video-Task Matching} (using wikiHow task names) produced by the \g for the video segment shown in \red{Fig.~\ref{appendix_fig:qua_1032}}.}
\label{tab:qua_2515_vtm}
\end{table*}

\begin{table*}[t]
\footnotesize
  \aboverulesep=0ex
  \belowrulesep=0ex 
\begin{center}
\begin{tabularx}{\linewidth}{l|lX}
\toprule
\textbf{Relation Type} & \textbf{Node ID} &\textbf{Step Headline} \\ \midrule
\multirow{6}{*}{1 Hop In-Neighbor} & \multirow{1}{*}{3209} & Choose the right spring for your car. \\ \cline{2-3}
 & \multirow{1}{*}{9593}   & Install the CV axle nut and tighten it securely. \\ \cline{2-3}
 &  \multirow{1}{*}{5567}   & Use any leftover grease on the bottom bolt if it needs any. \\ \cline{2-3}
    &  \multirow{1}{*}{9857}  & Find a level work surface. \\  \cline{2-3} 
      & \multirow{2}{*}{599} & Loosen the lug nuts of the tire with a lug wrench. \\
      & & Loosen the lug nuts with a lug wrench (tire iron) or impact wrench. \\ \hline
 \multirow{13}{*}{1 Hop Out-Neighbor} & \multirow{1}{*}{5882} & Jack the trailer axle up with a bottle or floor jack. \\  \cline{2-3}
 & \multirow{8}{*}{72}  & Jack up the vehicle if need be.  \\ 
 & & Jack the car up if necessary. \\ 
 & & Jack up the vehicle if need be. \\
 & & Jack the car up. \\
 & & Jack up the vehicle. \\
 & & Jack up the car if necessary. \\
 & & Jack up the car. \\
 & & Jack the vehicle up. \\ \cline{2-3}
 & \multirow{1}{*}{5197}  & Repeat steps 3-11 on the other side of the vehicle. \\  \cline{2-3}
    &  \multirow{2}{*}{5659}  & Test drive the vehicle, making sure the steering feels tight, and there are no unusual sounds from the new parts which could indicate improper installation or other damaged parts.  \\   \cline{2-3}
 & \multirow{1}{*}{8608}  & Raise the car in the air. \\ \hline
  \multirow{5}{*}{2 Hops In-Neighbor}   & \multirow{1}{*}{8066} & Install the new brake pads in place. \\ \cline{2-3}
 & \multirow{1}{*}{6789}  & Get some jack stands. \\ \cline{2-3}
   & \multirow{1}{*}{4680} & Install the antilock sensor cable, anchoring it to the brake fluid line and frame in the same locations the previous cable was anchored, if you have replaced the wheel bearing hub assembly and your vehicle is equipped with four wheel antilock brakes. \\ \cline{2-3}
 \hline
 \multirow{6}{*}{2 Hops Out-Neighbor} & \multirow{1}{*}{9351} & Remove the lug nuts and wheel with the lug wrench and look at the back of the wheel and hub for grease. \\ \cline{2-3}
 & \multirow{4}{*}{75} & Remove the lug nuts and the wheel.  \\ 
  & & Remove the lug nuts from the wheel to remove it. \\ 
  & & Remove the lug nuts and pull the wheel off of the hub. \\
  & & Remove the lug nuts from each wheel and remove the wheels from the hubs. \\ \cline{2-3}
  & \multirow{1}{*}{8221} & Check the rotation pattern of your tires.Tires are either directional or non-directional.  \\  \cline{2-3}
\bottomrule
\end{tabularx}
\end{center}
\vspace{-10pt}
\caption{Pseudo labels of \textbf{Node Relation Learning} produced by the \g for the video segment shown in \red{Fig.~\ref{appendix_fig:qua_1032}}.}
\label{tab:nrl_qua_1032}
\end{table*}

\begin{table*}[t]
\footnotesize
\begin{center}
\begin{tabular}{ll}
\toprule
 \textbf{Node ID} &\textbf{Step Headline} \\ \midrule
 \multirow{1}{*}{7256} & Fill a container with cool water and a few drops of shampoo. \\ \hline
 \multirow{1}{*}{4766}   & Add a few drops of dish soap into the water. \\\hline
  \multirow{1}{*}{2192}    & Pour the mixture into a spray bottle. \\ 
\bottomrule
\end{tabular}
\end{center}
\vspace{-10pt}
\caption{Pseudo labels of \textbf{Video-Node Matching} produced by the \g for the video segment shown in \red{Fig.~\ref{appendix_fig:qua_1748}}.}
\label{tab:qua_1748_vnm}
\end{table*}

\begin{table*}[t]
\footnotesize
\begin{center}
\begin{tabular}{l}
\toprule
 \textbf{Task Name} \\ \midrule
How to Fix Doll Hair \\
 How to Make Hair Spray \\
\bottomrule
\end{tabular}
\end{center}
\vspace{-10pt}
\caption{Pseudo labels of \textbf{Video-Task Matching} (using wikiHow task names) produced by the \g for the video segment shown in \red{Fig.~\ref{appendix_fig:qua_1748}}.}
\label{tab:qua_2515_vtm}
\end{table*}

\begin{table*}[t]
\footnotesize
  \aboverulesep=0ex
  \belowrulesep=0ex 
\begin{center}
\begin{tabularx}{\linewidth}{l|lX}
\toprule
\textbf{Relation Type} & \textbf{Node ID} &\textbf{Step Headline} \\ \midrule
\multirow{8}{*}{1 Hop In-Neighbor} & \multirow{6}{*}{49} & Brush your hair. \\
 & & Brush out your hair. \\
 & & Brush your hair so that it is ready for styling. \\
 & & Brush the hair. \\
 & & Gently brush the hair. \\
 & & Brush out your hair well. \\ \cline{2-3}
 & \multirow{1}{*}{3490}   & Fill a container with cool water. \\ \cline{2-3}
 &  \multirow{1}{*}{416}   & Remove the saucepan from heat, and let the mixture cool before adding 4 to 5 drops of your favorite essential oil. \\ \hline
 \multirow{4}{*}{1 Hop Out-Neighbor} & \multirow{1}{*}{5372} & Place the wig into the water. \\  \cline{2-3}
 & \multirow{1}{*}{4950}  & Brush the doll's hair.\\  \cline{2-3}
 & \multirow{2}{*}{40}  & Close the bottle, and shake it before you use it. \\ 
  & & Close the bottle and shake it. \\  \hline
  \multirow{3}{*}{2 Hops In-Neighbor}   & \multirow{1}{*}{4449} & Remove the doll's wig, if possible. \\ \cline{2-3}
 & \multirow{1}{*}{2935}  & Determine what materials the doll and the doll hair are made out of. \\ \cline{2-3}
   & \multirow{1}{*}{446} & Add the coconut oil and stir with a spoon until it has melted. \\ 
 \hline
 \multirow{2}{*}{2 Hops Out-Neighbor} & \multirow{1}{*}{3011} & Run clean water over the wig. \\ \cline{2-3}
 & \multirow{1}{*}{7603} & Consider protecting the doll's face. \\  \hline
\bottomrule
\end{tabularx}
\end{center}
\vspace{-10pt}
\caption{Pseudo labels of \textbf{Node Relation Learning} produced by the \g for the video segment shown in \red{Fig.~\ref{appendix_fig:qua_1748}}.}
\label{tab:nrl_qua_1748}
\end{table*}

\begin{table*}[t]
\footnotesize
\begin{center}
\begin{tabular}{ll}
\toprule
 \textbf{Node ID} &\textbf{Step Headline} \\ \midrule
 \multirow{1}{*}{4251} & Put on your full belly dance skirt, the over-skirt, and a belly dance belt and you are ready to dance! \\ \hline
 \multirow{1}{*}{10001}   & Find suitable tights. \\ \hline
  \multirow{1}{*}{6603}    & Cut your tights at the chosen length. \\ 
\bottomrule
\end{tabular}
\end{center}
\vspace{-10pt}
\caption{Pseudo labels of \textbf{Video-Node Matching} produced by the \g for the video segment shown in \red{Fig.~\ref{appendix_fig:qua_620069}}.}
\label{tab:qua_620069_vnm}
\end{table*}

\begin{table*}[t]
\footnotesize
\begin{center}
\begin{tabular}{l}
\toprule
 \textbf{Task Name} \\ \midrule
How to Make a Full Belly Dance Skirt \\ 
How to Make Leggings from Tights \\
\bottomrule
\end{tabular}
\end{center}
\vspace{-10pt}
\caption{Pseudo labels of \textbf{Video-Task Matching} (using wikiHow task names) produced by the \g for the video segment shown in \red{Fig.~\ref{appendix_fig:qua_620069}}.}
\label{tab:qua_2515_vtm}
\end{table*}

\begin{table*}[t]
\footnotesize
  \aboverulesep=0ex
  \belowrulesep=0ex 
\begin{center}
\begin{tabularx}{\linewidth}{l|lX}
\toprule
\textbf{Relation Type} & \textbf{Node ID} &\textbf{Step Headline} \\ \midrule
\multirow{5}{*}{1 Hop In-Neighbor} & \multirow{1}{*}{3936} & Make an over-skirt from leftover fabric. \\ \cline{2-3}
 & \multirow{1}{*}{9045}   & Decide upon the length of the leggings you'd prefer. \\ \cline{2-3}
 &  \multirow{1}{*}{2120}   & Bring the other end of the sling over your other shoulder. \\ \cline{2-3}
    &  \multirow{1}{*}{6953}  & Pivot your hips forward.\\  \cline{2-3} 
      & \multirow{1}{*}{7311} & Repeat the forward lunge. \\ \hline
 \multirow{5}{*}{1 Hop Out-Neighbor} & \multirow{1}{*}{9045} & Decide upon the length of the leggings you'd prefer.\\  \cline{2-3}
 & \multirow{1}{*}{6342}  & Turn the cut fabric over twice to create a half-inch (1.27cm) hemline. \\ \cline{2-3}
 & \multirow{1}{*}{2120}  & Bring the other end of the sling over your other shoulder. \\  \cline{2-3}
    &  \multirow{1}{*}{6953}  &  Pivot your hips forward.  \\   \cline{2-3}
 & \multirow{1}{*}{7311}  & Repeat the forward lunge. \\ \hline
  \multirow{4}{*}{2 Hops In-Neighbor}   & \multirow{1}{*}{2863} & Hem the skirt with a machine by rolling the fabric between your thumb and forefinger and guide it through the machine without stretching. \\ \cline{2-3}
 & \multirow{1}{*}{10001}  & Find suitable tights. \\ \cline{2-3}
   & \multirow{1}{*}{7311} & Repeat the forward lunge. \\ \cline{2-3}
 \hline
 \multirow{3}{*}{2 Hops Out-Neighbor} & \multirow{1}{*}{6603} & Cut your tights at the chosen length. \\ \cline{2-3}
 & \multirow{1}{*}{9276} & Stitch the hems in place by hand. \\ \cline{2-3}
  & \multirow{1}{*}{4370} & Continue wrapping the rest of the bandage around the tail. \\  \cline{2-3}
\bottomrule
\end{tabularx}
\end{center}
\vspace{-10pt}
\caption{Pseudo labels of \textbf{Node Relation Learning} produced by the \g for the video segment shown in \red{Fig.~\ref{appendix_fig:qua_620069}}.}
\label{tab:nrl_qua_620069}
\end{table*}

\renewcommand{\thesection}{B}
\section{Implementation Details}
\label{appendix_sec:implementation}

\subsection{Node \& Edge Construction of the \textbf{\geos}}
\label{appendix_subsec:node_edge_construct}
The wikiHow database that we used has a total of $10,588$ step headlines from $T=1,053$ task articles~\cite{wikihow}.
This is the same version of the wikiHow database
$\mathbb{B}$ 
that DS~\cite{distant} used.
These wikiHow tasks have at least $100$ video samples in the HowTo100M dataset~\cite{distant}.
The text branch of the author-released 
S3D model pre-trained using the MIL-NCE objective~\cite{s3d} (in the paper, we call it the MIL-NCE model for short) 
outputs the feature of each step headline. 
We used Agglomerative Clustering 
from the scikit-learn library 
for step deduplication (\textbf{\texttt{Step 1}} from \red{Sec. 3.2} in the main paper).
We used a relatively conservative clustering criterion since the goal here is deduplication and avoiding putting two semantically-different steps into one cluster is desired.
Specifically, we used the minimum of the distances between all observations of the two sets as the linkage criterion, with cosine similarity as the distance function and a threshold of $0.09$. The resulting number of step nodes is $10,038$. Among them, $314$ step nodes have more than one step headline as its members.
We list the top $3$ largest step nodes and their randomly sampled members in \red{Table~\ref{tab:step_nodes}} of the Supplementary Material. We find cross-task characteristics of steps, i.e., one step (which can be described slightly differently) can belong to multiple tasks.

We set the video segments to be $9.6$ seconds long in consideration of the temporal lengths of steps and videos in HowTo100M. There is no temporal overlapping or spacing between segments of one video. This leads to $3.7$M video segment samples for the HowTo100M subset~\cite{bertasius2021space} that we have mainly used for model training and the \g building, and $51$M video segment samples for the full HowTo100M dataset.
The MIL-NCE model was pre-trained on the full HowTo100M dataset on $3.2$ seconds long video segments~\cite{s3d}. Therefore, the feature of each $9.6$ seconds long segment was mean pooled from features of the three $3.2$ seconds long segments.
Dot product between MIL-NCE produced feature of a step headline and MIL-NCE produced feature of frames of a video segment, yielded a similarity score, which was considered to be the matching confidence score between the step headline and the video segment.

In order to obtain direct step transitions in HowTo100M (\textbf{\texttt{Step 2}} from \red{Sec. 3.2} in the main paper), we looped through the videos in the HowTo100M subset, and for each  video, we started from the first segment to the second last to collect the candidates of direct step transitions. Specifically, for every two temporally adjacent segments in the video, e.g., for segment $i$ and segment $j$, pair-wise combinations of the matched step headlines of segment $i$ and the matched step headlines of segment $j$ form the candidates, but only if the preceding step headline and the succeeding step headline are not the same.
For each segment, we considered the step headlines with a similarity score higher than $10$ as the matched step headlines of the segment. 
A step headline transition, e.g.,  (step headline $m \rightarrow$ step headline $n$), can appear in multiple videos; we call each occurrence of such transition as one step transition instance.
The score of one step transition instance is the product of the matching score of the preceding step headline and the matching score of the succeeding step headline. Final score of the step transition (step headline $m \rightarrow$ step headline $n$) is the summed score aggregated from all instances of the step transition (step headline $m \rightarrow$ step headline $n$). In this way, step transitions that happen more frequently in the video corpus can have  higher step transition scores (i.e., more confident).

Given the collection of step transition candidates from HowTo100M, we removed these less confident candidates if the step transition score is lower than $1000$ and then performed log min-max normalization to constrain the scores into the range of $[0, 1]$. 

In the above, we have described how we obtained the step transitions from HowTo100M. We describe how we obtained the step transitions from wikiHow in the following.
In an wikiHow article $t$, a pair ($s_{i}^{(t)}$, $s_{i+1}^{(t)}$) is defined to be a direct step transition. We assigned a score of $1$ for all direct step transitions in wikiHow, since these step transitions were all annotated by humans.

Using the mapping from the step headline to step node, step transitions from both wikiHow and HowTo100M were added as edges to connect the step nodes, and thus formed the structure of the \geos. For a directed node pair ($n_1$, $n_2$), if there are multiple step transitions and hence multiple scores, we kept the maximal score as the confidence score of the edge ($n_1$, $n_2$)  (i.e, node $n_1\rightarrow$ node $n_2$).

All thresholding criteria involved in the above graph construction process were empirically chosen through qualitative manual examination.

\subsection{Pseudo Label Generation}
For Video-Node Matching (VNM), the pseudo labels of one video segment are the node IDs of the top $3$ step nodes with the highest matching confidence scores. Since a step node may have multiple step headlines as its members, the matching score between a step node and a segment, is set to be the maximal value of the matching scores that are generated by all pairs of the node's step headline and the segment. 
Because we match each video segment to the ``top 3'' nodes in the graph, there is the risk of forcing a match when no relevant node exists in the graph. However, when the graph is sufficiently large, most segments will find a node of relative relevance. One could further improve over this by thresholding on the matching score and reassigning to a background node when no nodes surpass the threshold.

For Video-Task Matching (VTM), the pseudo labels of one video segment are the wikiHow task names of the step headlines, which are the members of the matched step nodes obtained using VNM. If the HowTo100M task names were used, we first need to obtain the occurrence matrix $O\in \mathbb{R}^{S \times T'}$ where $S$ denotes the number of step headlines in wikiHow ($S=10,588$) and $T'$ is the number of HowTo100M tasks that are covered by the videos ($T'=1,059$ for the HowTo100M subset, and $T'=25$K for the fullset). We populated $O$ by looping through segments of videos. For each segment, given the video's task name annotation, we incremented this task's occurrence of the matched step headlines by $1$ (the step headlines are members of the matched step nodes from VNM). Given the step headlines of the matched step nodes of the video segment, the video segment's VTM pseudo labels using the HowTo100M task annotations would then be the top $3$ HowTo100M task names with the highest occurrences of the step headlines.

Task Context Learning (TCL) is built upon VTM. Given the matched tasks from VTM, using the transposed version of the step-task occurrence matrix, we obtained the step headlines that a task needs. We then mapped these step headlines to step nodes, and the node IDs are the final TCL pseudo labels of the video segment. If the HowTo100M task names were used, the information on the step headlines that a task needs can be noisy because many step headlines may have a non-zero occurrence value (because video-step matching is not perfect). Therefore, we only considered the top $3$ step nodes (mapped from step headlines) with the highest non-zero task occurrence values as the pseudo labels of TCL, if the task names were from HowTo100M.

W.r.t how to obtain Node Relationship Learning (NRL) pseudo labels of one video segment, for each matched step node from VNM, we queried the  \g to obtain its $k$-hop in-neighbors and out-neighbors. One node may have multiple in-neighbors and/or out-neighbors. Suppose node $j$ is one of the $k$-hop in-neighbors of node $i$, it means that there is a directed path from node $j$ to
node $i$ of length exactly $k$ ($k$ edges along the directed path). Similarly, suppose node $j$ is one of the $k$-hop out-neighbors of node $i$, it means that there is a directed path from node $i$ to
node $j$ of length exactly $k$.

The confidence score of the $1$-hop in-neighbors and out-neighbors are the edge confidence scores of the edges that connect the matched step node and its $1$-hop neighbors. If $k>1$, the confidence score of a $k$-hop  \textit{in}-neighbor $i$ is the edge confidence score of the directed edge ($i$, $j$), i.e., node $i \rightarrow$ node $j$, multiplied by the confidence score of node $j$ being the video segment's $(k-1)$-hop \textit{in}-neighbor.
A similar strategy applies to $k$-hop \textit{out}-neighbors: if $k>1$, the confidence score of a $k$-hop out-neighbor $i$ is the edge confidence score of the directed edge ($j$, $i$), i.e., node $j \rightarrow$ node $i$, multiplied by the confidence score of node $j$ being the video segment's $(k-1)$-hop \textit{out}-neighbor.
When the neighbors are available, we considered the top $5$ most confident neighbors for the first hop ($k=1$) in- or out-neighbors, and top $3$ for the second hop ($k=2$) in- or out-neighbors; these node IDs are the pseudo labels.

\subsection{Pretraining \textbf{\ours}} 
The input to $f(\cdot)$
is the video segment's visual feature produced by MIL-NCE~\cite{s3d} with a dimensionality of $512$. In practice, the architecture we chose for $f(\cdot)$ is a shallow MLP with only one hidden layer, which is a bottleneck layer with a dimensionality of $128$. The output layer of the MLP $f(\cdot)$ has a dimensionality of $512$, which means we set the dimensionality of the refined video segment feature to be the same as the original video segment feature, in order to verify the refinement ability of $f(\cdot)$ for procedure understanding. The ReLU non-linear activation was used in between the linear layers of the MLP $f(\cdot)$.

Output representation of $f(\cdot)$ is the input to multiple answer heads $a(\cdot)$ to perform the pre-training objectives. For example, \ours with NRL (\textit{2 hops}) (\red{Table 1} in the main paper) indicates that during pre-training, there were $4$ answer heads because of 2 hops and 2 directions (in and out). \ours with VNM + VTM (\textit{wikiHow} + \textit{HT100M}) + TCL (\textit{wikiHow}) + NRL (\textit{1 hop}) indicates $1$ (VNM)$+2$(VTM)$+1$(TCL)$+2$(NRL) $=6$ answer heads. Parameters of these answer heads were not shared, and their exact architectures used are described in the next paragraph.

All pre-training objectives were modelled as a \textit{multi-label classification} problem. For example, for NRL, 
to predict the $1$-hop out-neighbors, 
the node IDs are the class indices, and for VTM, the task IDs are the class indices. 
The answer heads are MLPs with ReLU being the non-linear activation.
For the pre-training objectives that use the node IDs as the class indices (VNM, TCL and NRL), the answer head MLP has two hidden layers with a dimensionality of $2509$ (\#classes//4) and $5019$ (\#classes//2) respectively, and the output layer has a dimensionality of $10038$ (\#classes) (`\#' means `the number of' and `//' denotes `divide and round down to the nearest integer'). For the pre-training objectives that use the task IDs as the class indices (VTM), the answer head MLP has one hidden layer with a dimensionality of \#classes//2. 
We used \textit{Binary Cross Entropy} as the loss function. We did not tune the loss co-efficient of each pre-training objective 
to maintain simplicity.

We set $f(\cdot)$ and answer heads $a(\cdot)$ to be simple MLPs and did not tune the model architectural settings in order to evaluate the effectiveness and ease of use of
the pseudo labels generated by the \g and the proposed pre-training objectives.
Using our proposed  method,
a simple architecture can
learn a refined video representation that leads to a stronger performance on the downstream procedure understanding tasks.

We emphasize that the  combination of the proposed pre-training objectives leads to better model generalization. Though NRL is the most effective pre-training objective, when $K$=$1$, the combination outperformed NRL in $11$ out of $12$ settings; when $K$=$2$, the combination outperformed NRL in $9$ out of $12$ settings.

We implemented \ours using PyTorch~\cite{paszke2019pytorch}. We used the Adam optimizer~\cite{kingma2014adam}, a learning rate of $0.0001$, a weight decay of $0$, and a batch size of $256$. We saved the model checkpoint at every $10$ epochs, and chose the downstream performance on the Step Forecasting task on the COIN dataset as the indicator for early stopping with a patience of $500$ epochs. The model trained at the epoch that gave the best result on the Step Forecasting task on the COIN dataset, was used to perform all of the downstream evaluation experiments (i.e., the same model checkpoint of $f(\cdot)$ was used for all $12$ evaluation settings).
When the HowTo100M full data was used for training, due to the much larger amount of training data, we saved the model checkpoint by training steps, i.e., we saved the model checkpoint for every $500$ batches and by the end of each training epoch, and trained $f(\cdot)$ up to $100$ epochs.
The pre-training of \ours that uses all four pre-training objectives (\red{Table 1} in the main paper) took roughly $100$ hours on $8$ NVIDIA A100 GPUs. When the full HowTo100M data was used, it took $50$ hours on $16$ NVIDIA A100 GPUs.
As a reference, according to DS~\cite{distant}, their pre-training took $55$ hours using $128$ GPUs, and MIL-NCE~\cite{s3d} reported that their pre-training required $3$ days with $64$ $8$-core TPUs.
Our framework is more efficient to train.

\subsection{Downstream Evaluation}
The downstream task model $t(\cdot)$ is either a MLP or a Transformer encoder layer~\cite{vaswani2017attention}, and the architectural setting is the same across downstream tasks and datasets.
The output of the trained frozen $f(\cdot)$ is the input to the downstream task model $t(\cdot)$. The 
\g and the answer heads used for the pre-training objectives were discarded during the downstream evaluation (i.e., testing time of $f(\cdot)$).

Since the input is a sequence of video segment features for all of the downstream tasks, we used the Learned Absolute Positional Encoding to learn a vectorized representation for each segment position. The segment's video representation and the position representation are summed to form the position-augmented segment feature. 

When the downstream task model $t(\cdot)$ is a MLP, the position-augmented segment features of the input sequence were summed to form a feature vector for the input sequence.
This MLP is the downstream task classifier. 

When the downstream task model $t(\cdot)$ is a Transformer, the input sequence to the Transformer encoder layer is the position-augmented segment features plus a learned CLS (which stands for `classification') token. 
We used only one layer of the Transformer encoder layer.
Transformer encoder performs relation reasoning over tokens in the sequence and updates the token representations. The updated feature of the  CLS token is the input to the MLP downstream task classifier. 

The number of attention heads for the Transformer encoder layer is $8$, the dimensionality of the feed-forward network in the Transformer encoder is $1024$, and the non-linear activation function is ReLU.
The MLP downstream task classifier has
a single hidden layer whose dimensionality is $128$ for Task Recognition, and $768$ for Step Recognition and Step Forecasting.
We trained the downstream task model for up to $1000$ epochs using early stopping 
with a patience of $50$ epochs. The optimizer was Adam, batch size was $16$, learning rate was $0.0001$, and weight decay was $0.001$.

\subsection{Implementation of Baselines}
The training and evaluation of baselines (i.e., MIL-NCE$^*$, DS, DS$^*$ and VSM in \red{Table 1} of the main paper) follow the same implementation protocols (e.g., downstream model architecture settings, hyper-parameters and training schedules, etc.) as described above for a fair comparison with our \ours variants. 

Specifically, the results of MIL-NCE$^*$ described in the main paper were obtained by training and evaluating the downstream task models that took the visual features of video segment from the author-released 
S3D model, which was pre-trained using the MIL-NCE objective~\cite{s3d}, as the input segment features. The results of MIL-NCE$^*$ in \red{Table 1} of the main paper can be interpreted as the results of removing $f(\cdot)$ of our framework. 
Moreover, we used~\cite{s3d}'s video features for all experiments (both pre-training and downstream experiments). 

Model architectures used in our experiments are not what were in~\cite{distant} (\cite{distant} used heavier architectures, e.g., TimeSformer~\cite{bertasius2021space} for pre-training and heavier Transformer downstream models).
In order to obtain the results of DS and DS$^*$, we trained the same MLP-based $f(\cdot)$ model architecture as what \ours used, but instead we trained it using the pre-training objective proposed by DS~\cite{distant} that matches video subtitles to step headlines using the pre-trained language model MPNet~\cite{song2020mpnet}.
In particular, the Step Classification pre-training objective~\cite{distant} was implemented to recognize the top $3$ step headlines from wikiHow that have the highest matching confidence scores (since our VNM objective matches the top $3$ nodes).
Therefore, both \ours and DS used the Binary Cross Entropy loss by modeling the pre-training objectives as multi-label classification problems. 
We also experimented with other loss functions demonstrated in DS~\cite{distant}, i.e., Distribution Matching (KL divergence loss) and Embedding Regression (NCE loss), but we found Step Classification with the Binary Cross Entropy loss is the most robust  one among these different loss forms because it obtained the best results in most cases.

Downstream datasets are relatively small; and thus, the risk of overfitting to small datasets exists.
Therefore, we did not select a pre-trained model (checkpoint) for each specific downstream task/dataset while we have $12$ downstream cases.
In addition, the design choices 
such as model architectures and hyper-parameters 
used in 
both 
pre-training and downstream tasks were set in an early stage when optimizing the results of the DS baseline, and they were kept constant for all subsequent model variants including \ourseos.

\begin{table*}[t]
\footnotesize
\begin{center}
\begin{tabularx}{\linewidth}{lXl}
\toprule
 \textbf{Task Name} & \textbf{Step Headline} &  \textbf{Node ID} \\ \midrule
How to Apply MonoKote   & Unroll a sheet of MonoKote on a clean work surface, and lay the model part on top of it.  & 3185 \\
  & Remove the clear backing from the MonoKote by attaching a small piece of cellophane tape on each side of the sheet and pulling gently. & \multirow{2}{*}{2604} \\
  & Place the MonoKote, adhesive side down, over the part. & 4136 \\
  & Set the heat sealing tool to 275 F (135 C) and allow it to come to temperature before proceeding. &  \multirow{2}{*}{1876} \\ 
  & Pull the MonoKote to the edge of the piece to be covered, and run the heat sealing iron along the edge to activate the adhesive and seal MonoKote to the part. & \multirow{2}{*}{2067} \\ 
  & Unroll a second sheet of MonoKote, and repeat the process on the other side of the aircraft part. & \multirow{2}{*}{2469} \\ 
  & Trim the edges carefully with the utility knife, leaving a 1/8 inch to 1/4 inch (3 to 6 mm) overlap. & \multirow{2}{*}{3139} \\ 
  & Finish the edge by carefully folding the top edge under to the bottom side and using the heating iron to seal it in place. &  \multirow{2}{*}{1982} \\  \midrule
How to Make a Pen Pocket for Your Journal  & Decide how tall and wide you want the pen pocket to be.  & 919 \\
  & Add ½ inch (1.27 centimeters) to the length and width. & \multirow{1}{*}{669} \\
  & Cut two pieces of fabric according to your measurements. & 1150 \\
  & Cut a third piece of for the front pocket. &  782 \\ 
  & Cut your elastic 1 inch (2.54 centimeters) longer than your body pieces. & 591\\ 
  & Fold and iron the top hem of the front pocket. & 443 \\ 
  & Top stitch the hem down. & 1087 \\ 
  & Pin the pocket to the front of one of the larger body pieces. & 233 \\
  & Create the individual pockets, if desired. & 748 \\
  & Place the elastic on top of the other body piece. & 823 \\
  & Pin the pocket piece on top, wrong-side-up. & 692 \\
  & Sew around the fabric, leaving a gap for turning. & 558 \\
  & Turn the fabric inside out through the gap.  & 429 \\
  & Press the entire pocket flat with an iron. & 583 \\
  & Top stitch around the entire piece. &  1470 \\  \midrule
How to Make a Mei Tai Baby Carrier  & Obtain the needed supplies listed below.  & 2296 \\
  & Cut an 18–22 inch (45.7–55.9 cm) square of sturdy cloth. & 3857 \\
  & If you have chosen multiple layers, baste or quilt them together before proceeding. & 2648 \\
  & Hem two opposite sides of your square. &  1965 \\ 
  & Measure your torso.  & 2373\\ 
  & Add 15–20 inches (38.1–50.8 cm) to this length to get the length of your two shoulder straps/ties. & \multirow{2}{*}{2395} \\ 
  & Cut material for two ties (length calculated in previous step) at least 8 inches (20.3 cm) wide. & \multirow{2}{*}{2483} \\ 
  & Fold the ties in half with narrow ends touching to find the center point of the length of the tie. & \multirow{2}{*}{1027} \\
  & Fold the square in half with the two previously hemmed edges together. & 1873 \\
  & After finding the center point, mark the length of the (raw edged) side of your square along the long side of the tie from the center point out. & \multirow{2}{*}{2144} \\
  & Remove the tie, fold it in half lengthwise (make a very long fold) and sew the sides closed. & 1805 \\
  & Miter or clip the seam allowances to allow the corners to turn neatly. & 3220 \\
  & Repeat measuring, marking and sewing steps for the second tie.  & 1744 \\
  & Place the raw edges of the square into the opening in the ties and pin in place. & 1255 \\
  & Iron the creases and pin carefully or baste. & 1655 \\
  & Top stitch the edges of the ties, taking care to sew multiple seams through ALL layers when sewing the raw edges of your square inside the ties. & \multirow{2}{*}{877} \\
  & The pouch is now ready to wear. &  6588 \\ 
\bottomrule
\end{tabularx}
\end{center}
\vspace{-10pt}
\caption{Pseudo labels of \textbf{Task Context Learning}
produced by the \g for the video segment shown in \red{Fig.~\ref{appendix_fig:qua_3515}}.}
\label{tab:qua_3515_tcl}
\end{table*}

\begin{table*}[t]
\footnotesize
\begin{center}
\begin{tabular}{lll}
\toprule
 \textbf{Task Name} & \textbf{Step Headline} &  \textbf{Node ID} \\ \midrule
How to Ink Stamps with Markers   & Select the right types of markers. & 3661 \\
  & Sand your stamp lightly. & 8141 \\
  & Spritz your stamps with water if you want a water color effect. & 8361 \\
  &  Color with the brush tip of your marker. &  2811 \\ 
  & Use different colors as needed.  & 8130 \\ 
  & Breathe on the stamp to moisten the ink.  & 3336 \\ 
  & Stamp the image onto the page. & 5354 \\ 
  & Experiment a little. & 2985 \\
  & Work quickly so the ink stays wet. & 4496 \\
  & Do not touch your inked image until it's dry. &  8988 \\
  & Avoid permanent markers on stamps. & 8589 \\  \midrule
How to Make a Translation Tessellation & Find an A4 size piece of paper.  & 7026 \\
  & Cut out a small square or parallelogram. & 7447 \\
  & Convert this base tessellation into a more interesting shape. & 8591 \\
  & Trace it on the 3" x 6" (7.5cm x 15cm) paper until it is full. &  5543 \\ 
  & Go over the lines with a black permanent marker. & 9399\\ 
  & Color it in anyway you like. & 8492 \\ 
  & Ink it all over. & 6068 \\ 
  & Let dry. &  236 \\ 
\bottomrule
\end{tabular}
\end{center}
\vspace{-10pt}
\caption{Pseudo labels of \textbf{Task Context Learning}
produced by the \g for the video segment shown in \red{Fig.~\ref{appendix_fig:qua_16818}}.}
\label{tab:qua_16818_tcl}
\end{table*}

\begin{table*}[t]
\footnotesize
\begin{center}
\begin{tabular}{lll}
\toprule
 \textbf{Task Name} & \textbf{Step Headline} &  \textbf{Node ID} \\ \midrule
How to Make Milk Paint  & Let the milk come to room temperature. & 3463 \\
  & Combine the milk and vinegar. & 3523 \\
  & Place the milk in a warm place for one to two days. & 3598 \\
  &  Add water to the pigment. &  2932 \\ 
  & Make a paste with the pigment.  & 4936 \\ 
  & Add water to the lime powder.  & 4730 \\ 
  & Mix the lime powder and water to make a wet paste. & 5370 \\ 
  & Line a colander with cheesecloth and put it in the sink. & 1454 \\
  & Pour the container of curdled milk over the colander. & 2289 \\
  & Transfer the quark into a paint bucket. &  5349 \\
  & Add the lime paste to the quark. & 2459 \\
  & Stir in the pigment paste. & 2374 \\
  & Strain the paint through a cheesecloth. & 8226 \\
  & Prepare the surface you are painting. & 4614 \\
  & Stir the milk paint. & 3364 \\
  & Apply the first layer of paint. & 2001 \\
  & Let the first layer dry. & 4198 \\
  & Stir the paint and apply the second layer. & 2929 \\
  & Apply a topcoat if desired. & 303 \\
  & Store milk paint in the fridge for up to three days. & 3008 \\  \midrule
How to Make Shimmering Finger Paints & Mix all the ingredients together in a medium pan.  & 8766 \\
  & Cook over low heat for 10 to 15 minutes. & 8052 \\
  & Keep stirring the finger paint mixture until it is smooth and thick. & 5660 \\
  & After the finger paint has thickened, take the pan off the stove. &  5957 \\ 
  & Place the paints in containers. & 5504 \\ 
  & Add a few drops of food coloring to each container. & 8702 \\ 
  & Try mixing colors or paints to get different colors. & 9377 \\
  & Stir the coloring into the paint to determine the shade of color. & 9026 \\ 
  & If using, add the desired amount of glitter. & 7885 \\
  & Stir well to get the glitter really mixed in. & 6937 \\
  & Cover tightly to store. &  5149 \\ 
\bottomrule
\end{tabular}
\end{center}
\vspace{-10pt}
\caption{Pseudo labels of \textbf{Task Context Learning}
produced by the \g for the video segment shown in \red{Fig.~\ref{appendix_fig:qua_62797}}.}
\label{tab:qua_62797_tcl}
\end{table*}

\begin{table*}[t]
\footnotesize
\begin{center}
\begin{tabular}{lll}
\toprule
 \textbf{Task Name} & \textbf{Step Headline} &  \textbf{Node ID} \\ \midrule
How to Cluster Stitch   & Tie the yarn onto your hook.  & 67 \\
  & Create a foundation chain. & 403 \\
  & Chain two. & 418 \\
  & Yarn over the hook. &  67 \\ 
  & Insert the hook into the next stitch. & 37 \\ 
  & Yarn over and draw through. & 493 \\ 
  & Yarn over and draw through twice. & 493 \\ 
  & Yarn over the hook and insert it into the next stitch. & 10016 \\ 
  & Yarn over and draw through. & 493 \\ 
  & Yarn over and draw through twice. & 493 \\ 
  & Yarn over the hook.  & 67 \\ 
  & Pull the yarn through all three loops. & 4836 \\ 
  & Chain two.  & 418 \\ 
  & Yarn over and insert the hook. & 67 \\ 
  & Yarn over and draw through. & 493 \\ 
  & Yarn over and pull through twice. & 17 \\
  & Work the next part into the next stitch.  & 568 \\
  & Yarn over and pull through.  & 17 \\
  & Wrap the yarn around and pull through twice.  & 17 \\
  & Work into the next stitch.  & 568  \\
  & Draw another loop.  &  284 \\
  & Yarn over and pull through twice. & 17 \\
  & Pull through the remaining loops.  & 4664 \\
  & Work one cluster stitch into a single stitch.  & 4974 \\
  & Chain two.	& 418 \\
  & Work a second cluster stitch into the same stitch. 	& 2909 \\
  & Chain two and repeat as needed.   & 7757 \\  \midrule
How to Crochet Popcorn Stitch  & Double crochet five times into one stitch. & 4989 \\
  & Insert the hook into the first stitch of the group.  & 37 \\
  & Close the group.  & 5520 \\
  & Separate your stitches. & 8514 \\
  & Create a foundation chain.  & 403 \\
  & Single crochet into each stitch.  & 310 \\
  & Single crochet twice.  & 243 \\
  & Make five double crochets. & 304\\
  & Slip stitch into the loops.  & 8616 \\
  & Single crochet into the next three stitches. &  9774 \\
  & Repeat across the row.  & 300 \\
  & Single crochet into each stitch.  & 310 \\
  & Repeat as needed. & 1201 \\
  & Create a foundation chain. & 403 \\
  & Double crochet into each stitch. & 243 \\
  & Chain three. & 8662 \\
  & Double crochet into the next five stitches. & 243 \\
  & Make five double crochets. &  304\\
  & Resituate the hook and close the group. & 6878 \\
  & Double crochet to the end of the row. & 2437 \\
  & Repeat as needed. & 1201 \\ 
\bottomrule
\end{tabular}
\end{center}
\vspace{-10pt}
\caption{Pseudo labels of \textbf{Task Context Learning (Part \RNum{1})} produced by the \g for the video segment shown in \red{Fig.~\ref{appendix_fig:qua_1680}}.}
\label{tab:qua_1680_tcl_part1}
\end{table*}

\begin{table*}[t]
\footnotesize
\begin{center}
\begin{tabular}{lll}
\toprule
 \textbf{Task Name} & \textbf{Step Headline} &  \textbf{Node ID} \\ \midrule
 How to Surface Crochet   & Insert the hook into the first stitch.  & 37 \\
  & Tie the yarn onto the hook.  & 67  \\
  & Pull up a loop.  & 588 \\
  & Insert the hook into the next spot.  &  8289 \\
  & Yarn over.  & 144 \\
  & Pull up the loop. & 588 \\
  & Pull the second loop through the first. & 8430 \\
  & Repeat as needed. & 1201 \\
  & Bind off the yarn.  & 137 \\ \midrule
How to Half Double Crochet  & Yarn over. &  144 \\
  & Insert the hook into the stitch. & 37 \\
  & Yarn over. &  144 \\
  & Draw up another loop. &  284 \\
  & Yarn over. & 144 \\
  & Draw through all three loops on the hook.  &  2027 \\
  & Create a foundation chain.  & 403 \\
  & Skip the first two chain stitches. & 2016 \\
  & Work a half double crochet. & 625  \\
  & Work another half double crochet. & 625 \\
  & Repeat across the chain. & 4574 \\
  & Create a turning chain. & 213\\
  & Skip a stitch. & 2529 \\
  & Work a half double crochet into the next stitch. & 243 \\
  & Repeat across the row. & 300 \\ \midrule
How to Treble Crochet & Make a slip knot.  & 241 \\ 
 & Yarn over. & 144 \\
 & Draw the yarn through the slipknot. & 9994 \\
 & Crochet a chain of your desired length. & 4798 \\
 & Crochet a turning chain. & 180 \\
 & Turn your work. & 79 \\
 & Yarn over twice. & 622 \\
 & Insert your hook. & 450 \\
 & Yarn over and draw through. & 493 \\
 & Yarn over and draw through two loops. & 624 \\
 & Yarn over and draw through two loops. & 624 \\
 & Yarn over and draw through the last two. & 4860 \\
 & Yarn over twice. & 622 \\
 & Insert your hook. & 450 \\
 & Yarn over and draw through. & 493 \\
 & Yarn over and draw through two loops. & 624 \\
 & Yarn over and draw through two loops. & 624 \\
 & Yarn over and draw through two loops. & 624 \\
 & Repeat steps 1-6 of this section. & 3526 \\
 & Turn your work. & 79 \\
 & Crochet a turning chain.  & 180 \\
 & Yarn over twice and insert hook. & 67 \\
 & Yarn over one and pull through two. & 4922 \\
 & Yarn over and pull through. & 17 \\
 & Repeat steps 3-5 until you reach the end of your chain. & 3464 \\
\bottomrule
\end{tabular}
\end{center}
\vspace{-10pt}
\caption{Pseudo labels of \textbf{Task Context Learning (Part \RNum{2})} produced by the \g for the video segment shown in \red{Fig.~\ref{appendix_fig:qua_1680}}.}
\label{tab:qua_1680_tcl_part2}
\end{table*}

\begin{table*}[t]
\footnotesize
\begin{center}
\begin{tabularx}{\linewidth}{lXl}
\toprule
 \textbf{Task Name} & \textbf{Step Headline} &  \textbf{Node ID} \\ \midrule
How to Fit an Electric Shower  & Pick a location for your electric shower that is near the main cold water supply and close to a spot where you can install an independent circuit.  & \multirow{2}{*}{8120} \\
  & Consult an electrician for advice on the size and type of independent circuit to install for your shower to ensure that you have adequate electricity. & \multirow{2}{*}{1810} \\
  & Consult a plumber to ensure that your building's plumbing system will be able to accommodate an electric shower. & \multirow{2}{*}{4289} \\
  & Install the independent circuit near the location of the electric shower along with any necessary consumer units or earth cables. & \multirow{2}{*}{4465} \\
  & Wire the independent circuit to an isolating switch, which should be located above the shower. & \multirow{2}{*}{4893} \\
  & Attach the electric cable from the isolating switch to the back of the electric shower power unit and wire accordingly. & \multirow{2}{*}{8259} \\
  & Secure a pipe from the cold water main supply to the spot where the shower unit will be mounted. & \multirow{2}{*}{8721} \\
  & Attach a non-return valve or stop tap to the pipe to isolate the shower's water supply from the rest of the building. & \multirow{2}{*}{4735} \\
  & Attach the pipe to the shower unit using a compression fitting. & 8332 \\
  & Mount the shower unit and the shower head to the wall. & 3765 \\
  & Turn on the water supply and the independent circuit. & 4498 \\
  & Check to see that the electric shower is heating the water quickly and efficiently. & \multirow{2}{*}{2931} \\ \midrule
How to Install a Grab Bar & Assemble the necessary tools. & 4193 \\
 & Examine the grab bar kit. & 1614 \\
 & Determine mounting location. & 4317 \\
 & Mark the location of the stud. & 6528 \\
 & Pre-drill pilot holes. & 6772 \\
 & Install the wall anchors. & 3616 \\
 & Seal the seams with silicone caulk. & 1846 \\
 & Test at the end by pulling on it.  & 3147 \\ \midrule
How to Install Electric Radiant Heat Mat Under a Tile Floor & Prepare your floor for tile by installing tile backer board on the floor, securing it to the existing sub-floor with thin-set mortar and cement board screws or nails. & \multirow{3}{*}{1464} \\
 & Make a scale drawing of the bathroom floor, including toilet, tub and vanity locations, and bring it to the tile store or home center so you can buy the proper size mat or combination of mats. & \multirow{3}{*}{888} \\
 & Check the wiring with a continuity tester, after purchasing the mat, to make sure it wasn’t damaged during manufacturing or shipping. & \multirow{2}{*}{2409} \\
 & Install an electrical outlet box 5 feet (1.5 m). & 377 \\
 & Note: following your preliminary layout, you should mark the path of the thick “power lead” between the mat and wall cavity and chisel a shallow trench into the floor. & \multirow{3}{*}{1547} \\
 & Draw the layout lines for the tile on the floor. & 1828 \\
 & Install the mat, securing it lightly to the cement board with double-face tape. &  \multirow{2}{*}{1183} \\
 & Check the mat wiring again with the continuity tester. &  1264 \\
 & Install conduit connectors to both ends of two pieces of 1/2-in. & 1725 \\
 & Spread the mortar over a 5- to 10-sq.-ft. area of floor. & 1057 \\
 & Lay the tile, then tap it firmly into place. & 2823 \\
 & Connect the power lead and thermostat wire to the thermostat, following manufacturer's instructions. & \multirow{2}{*}{837} \\
\bottomrule
\end{tabularx}
\end{center}
\vspace{-10pt}
\caption{Pseudo labels of \textbf{Task Context Learning}
produced by the \g for the video segment shown in \red{Fig.~\ref{appendix_fig:qua_2345}}.}
\label{tab:qua_2345_tcl}
\end{table*}

\begin{table*}[t]
\footnotesize
\begin{center}
\begin{tabularx}{\linewidth}{lXl}
\toprule
 \textbf{Task Name} & \textbf{Step Headline} &  \textbf{Node ID} \\ \midrule
How to Foundation Piece a Quilt Block   & Select your pattern and photocopy or print enough of them to make your quilt.  & \multirow{2}{*}{1075} \\
 & Select your fabrics. & 200 \\
 & Launder all of your fabrics. & 726 \\
 & Iron the fabrics smooth if necessary. & 1215 \\
 & Cut rectangles or squares in sizes which will cover the shapes in your pattern blocks. & \multirow{2}{*}{915} \\
 & Notice that the pattern pieces are numbered in the order in which you should sew the pieces. & \multirow{2}{*}{2310} \\
 & Place the cloth for piece \#1 on the BACK side of the paper with the back/wrong side of the cloth towards the paper. & \multirow{2}{*}{895} \\
 & Hold the paper up to a light to verify that the fabric is oriented so that it covers all of the area of piece one with at least a quarter inch of overlap in all directions. & \multirow{3}{*}{564} \\
 & Place the cloth for piece \#2 (white) with its right/front side facing the right/front side of piece \#1 (red) and its seam edge aligned with the seam line and overlapping by a minimum of a quarter inch. &  \multirow{4}{*}{108} \\
 & Pin the two fabrics in place on the paper. & 191 \\
 & Flip the paper to the front/printed side. & 1112 \\
 & Machine stitch the seam line from the printed side. & 496 \\
 & Trim the seam allowances to 1/4 inch (0.6 cm). & 74 \\
 & Unpin the fabrics and flip piece \#2 over the seam and pin it in place over its allotted area on the block. & \multirow{2}{*}{411} \\
 & Hold up the paper block pattern to the light to check that piece \#2 will cover its allotted area. & \multirow{2}{*}{348} \\
 & Place the cloth for piece \#3 with its right/front side facing the right/front side of piece \#2... and its seam edge aligned with the seam line and overlapping by a minimum of a quarter inch (6 mm). & \multirow{4}{*}{108} \\
 & Pin the two fabrics in place on the paper. & 191 \\
 & Flip the paper to the front/printed side and use back light to check the placement. & \multirow{2}{*}{1599} \\
 & Machine stitch the seam line from the printed side. & 496 \\
 & Trim the seam allowances to 1/4 inch (6 mm). & 74 \\
 & Unpin the fabrics and flip piece \#3 over the seam and pin it in place over its allotted area on the block. & \multirow{2}{*}{411} \\
 & Repeat the process of placing, pinning, checking then sewing and trimming for each successively numbered piece. & \multirow{2}{*}{765} \\
 & Machine baste around the perimeter of your block when complete. & \multirow{2}{*}{432} \\
 & Before trimming - note the ragged edges. & 1017 \\
 & Tear away the paper ``backing''. & 388 \\
 & Voilà! & 485 \\ \midrule
How to Replace the Front Brake Pads on a 1998 to 2002 Honda Accord  &  Make sure the vehicle is on a level surface, the transmission is in park, and the emergency brake is engaged. &  \multirow{2}{*}{8099} \\
 & Starting on either side, locate the brake caliper (which is directly behind the tire). & \multirow{2}{*}{6849} \\
 & Lift the bottom of the caliper so that it pivots off of the top bolt. & 6132 \\
 & Remove both brake pads. & 9478 \\
 & Use the brake cleaner to spray down the entire brake assembly. & 6835 \\
 & On the inside of the brake caliper is the piston which extends out towards the back pad. & \multirow{2}{*}{6536} \\
 & Apply brake grease to the new pads. & 9816 \\
 & Install the new brake pads in place. & 8066 \\
 & Use any leftover grease on the bottom bolt if it needs any. & 5567 \\
 & Put the tire back on and tighten the lug nuts as best as you can. & 68 \\
 & Repeat steps 3-11 on the other side of the vehicle. & 5197 \\
 & Clear any tools from under the vehicle, and pump the brakes until they begin to firm up. & \multirow{2}{*}{8864} \\ 
\bottomrule
\end{tabularx}
\end{center}
\vspace{-10pt}
\caption{Pseudo labels of \textbf{Task Context Learning  (Part \RNum{1})} produced by the \g for the video segment shown in \red{Fig.~\ref{appendix_fig:qua_1032}}.}
\label{tab:qua_1032_tcl_part1}
\end{table*}

\begin{table*}[t]
\footnotesize
\begin{center}
\begin{tabularx}{\linewidth}{lXl}
\toprule
 \textbf{Task Name} & \textbf{Step Headline} &  \textbf{Node ID} \\ \midrule
How to Change a CV Axle and Front Wheel Bearing on a 2001 4X4 Dodge Dakota  & Troubleshoot your front wheel drive system to determine if maintenance is needed.  & \multirow{2}{*}{7869}   \\
 & Look underneath the truck between the wheels and the front differential to see if any visible damage is evident. & \multirow{3}{*}{5109} \\
 & Drive the truck on a smooth surface with the windows down, listening for clicking, grinding, or other unusual noises from the front of your truck. & \multirow{3}{*}{8245} \\
 & Jack up the front end of your truck and shake the wheels to see if any play is evident. & \multirow{2}{*}{9951} \\
 & Rotate the tires while they are off the ground, listening and feeling for anything unusual. & \multirow{2}{*}{8293} \\
 & Rock the wheels back and forth and listen for sounds from the CV joints. & \multirow{2}{*}{3486} \\
 & Have a mechanic do the inspection if you have any doubts, since the parts needed for this project cost several hundred dollars. & \multirow{3}{*}{3976} \\
 & Make a list of the parts you need to complete the project. & \multirow{2}{*}{2874} \\
 & Write down the information on your truck before going shopping. & \multirow{2}{*}{6557} \\
 & Choose the quality of the parts you want to use. & 7493 \\
 & Jack up the truck and set it on jack stands, making sure the wheels are blocked to prevent it from rolling. & \multirow{3}{*}{7520} \\
 & Remove the wheel from the side you are working one. & \multirow{2}{*}{8910} \\
 & Remove the cotter pin from the axle shaft at the center of the hub. & \multirow{2}{*}{3599} \\
 & Remove the brake caliper. & 139 \\
 & Remove the brake rotor. & 9584 \\
 & Remove the axle nut. & 9066 \\
 & Remove the hub bearing assembly. & 3789 \\
 & Remove the upper ball joint from the knuckle or unbolt it from the upper control (A) arm. & \multirow{2}{*}{8639} \\
 & Pull the CV axle off of the differential side shaft. & 9496 \\
 & Remove the oil seal at the side shaft if you are replacing it. & \multirow{2}{*}{9112} \\
 & Install the new oil seal at the side shaft. & 8206 \\
 & Slide the new CV axle over the side shaft on the differential. & \multirow{2}{*}{7638} \\
 & Guide the outer end of the axle through the center hole in the knuckle. & \multirow{2}{*}{6924} \\
 & Install the new hub bearing assembly. & 7721 \\
 & Reinstall (or install a new) the upper ball joint. & 6509 \\
 & Replace the brake rotor. & 4759 \\
 & Reinstall the brake caliper. & 4534 \\
 & Install the antilock sensor cable, anchoring it to the brake fluid line and frame in the same locations the previous cable was anchored, if you have replaced the wheel bearing hub assembly and your vehicle is equipped with four wheel antilock brakes. & \multirow{5}{*}{4680} \\
 & Install the CV axle nut and tighten it securely. & 9593 \\
 & Put your tire back on the truck and tighten the lug nuts. & \multirow{2}{*}{68} \\
 & Test drive the vehicle, making sure the steering feels tight, and there are no unusual sounds from the new parts which could indicate improper installation or other damaged parts. & \multirow{4}{*}{5659} \\
 & Return any parts which had a core charge to the retailer you purchased them from, for a refund of the core charge. & \multirow{3}{*}{7860} \\
\bottomrule
\end{tabularx}
\end{center}
\vspace{-10pt}
\caption{Pseudo labels of \textbf{Task Context Learning  (Part \RNum{2})} produced by the \g for the video segment shown in \red{Fig.~\ref{appendix_fig:qua_1032}}.}
\label{tab:qua_1032_tcl_part2}
\end{table*}

\begin{table*}[t]
\footnotesize
\begin{center}
\begin{tabular}{lll}
\toprule
 \textbf{Task Name} & \textbf{Step Headline} &  \textbf{Node ID} \\ \midrule
How to Rotate Tires  & Get some jack stands.  & 6789  \\
 & Find a level work surface. & 9857 \\
 & Remove the hubcaps and loosen the lug nuts. & 9667 \\
 & Raise the car in the air. & 8608 \\
 & Check the rotation pattern of your tires.Tires are either directional or non-directional. & 8221 \\
 & Remove the lug nuts from the first tire you've raised and remove it. & 4386 \\
 & Rotate the tires in the correct pattern. & 6012 \\
 & Lower the car. & 28 \\
 & Tighten lug nuts using the star pattern. & 8878 \\
 & Place hubcaps back on the wheels by replacing the lug nuts. & 9528 \\
\bottomrule
\end{tabular}
\end{center}
\vspace{-10pt}
\caption{Pseudo labels of \textbf{Task Context Learning  (Part \RNum{3})} produced by the \g for the video segment shown in \red{Fig.~\ref{appendix_fig:qua_1032}}.}
\label{tab:qua_1032_tcl_part3}
\end{table*}

\begin{table*}[t]
\footnotesize
\begin{center}
\begin{tabular}{lll}
\toprule
 \textbf{Task Name} & \textbf{Step Headline} &  \textbf{Node ID} \\ \midrule
How to Fix Doll Hair &  Determine the problem. & \multirow{1}{*}{4443} \\
 &  Start with brushing your doll's hair. & 4572  \\
  &  Consider trimming the doll hair. & 8981  \\
  & Consider curling the hair. & 9590 \\
  & Consider washing the doll hair. & 4647 \\
  & Determine what materials the doll and the doll hair are made out of. & 2935 \\
  & Fill a container with cool water. & 3490 \\
  & Add a few drops of dish soap into the water. & 4766 \\
  & Brush the doll's hair. & 4950 \\
  & Consider protecting the doll's face. & 7603 \\
  & Dip the doll's hair into the water. & 3560 \\
  & Lather the doll’s hair. & 4682 \\
  & Rinse the doll hair with clean, cool water. & 8298 \\
  & Consider using conditioner to detangle the doll's hair. & 3654 \\
  & Consider styling the hair. & 1091 \\
  & Set the hair on a towel to dry. & 4713 \\
  & Brush the hair. & 49 \\
  & Remove the doll's wig, if possible. & 4449 \\
  & Gently brush the hair. & 49 \\
  & Fill a container with cool water and a few drops of shampoo. & 7256 \\
  & Place the wig into the water. & 5372 \\
  & Run clean water over the wig. & 3011 \\
  & Consider a vinegar soak to make the hair shiny. & 6070 \\
  & Lay the wig on a towel. & 205 \\
  & Place another towel over the wig. & 205 \\
  & Place the wig onto a fresh towel. & 205 \\
  & Glue the wig back onto the doll head. & 3884 \\
  & Consider styling the hair. & 1091 \\
  & Consider curling your doll's hair. & 4690 \\
  & Obtain something to curl your doll hair with. & 4488 \\
  & Wrap some hair around your curler. & 3074 \\
  & Secure the curler. & 1403 \\
  & Repeat for the rest of the hair. & 8562 \\
  & Wait until the hair dries. & 3287 \\
  & Take out the curlers. & 3945 \\
\bottomrule
\end{tabular}
\end{center}
\vspace{-10pt}
\caption{Pseudo labels of \textbf{Task Context Learning (Part \RNum{1})}
produced by the \g for the video segment shown in \red{Fig.~\ref{appendix_fig:qua_1748}}.}
\label{tab:qua_1748_tcl_part1}
\end{table*}

\begin{table*}[t]
\footnotesize
\begin{center}
\begin{tabularx}{\linewidth}{lXl}
\toprule
 \textbf{Task Name} & \textbf{Step Headline} &  \textbf{Node ID} \\ \midrule
How to Make Hair Spray &  Bring 1 cup (240 milliliters) of water to a simmer in a saucepan. & \multirow{1}{*}{644} \\
 & Stir in 1 tablespoon of sea salt. & 777 \\
 & Add the coconut oil and stir with a spoon until it has melted. & 446 \\
 & Remove the saucepan from heat, and let the mixture cool before adding 4 to 5 drops of your favorite essential oil. & \multirow{2}{*}{416} \\
 & Pour the mixture into a spray bottle. & 2192 \\
 & Close the bottle, and shake it before you use it. & 40 \\
\bottomrule
\end{tabularx}
\end{center}
\vspace{-10pt}
\caption{Pseudo labels of \textbf{Task Context Learning (Part \RNum{2})}
produced by the \g for the video segment shown in \red{Fig.~\ref{appendix_fig:qua_1748}}.}
\label{tab:qua_1748_tcl_part2}
\end{table*}

\begin{table*}[t]
\footnotesize
\begin{center}
\begin{tabularx}{\linewidth}{lXl}
\toprule
 \textbf{Task Name} & \textbf{Step Headline} &  \textbf{Node ID} \\ \midrule
How to Make a Full Belly Dance Skirt  & Select your fabric. & 200 \\
 & Cut the panels. & 6480 \\
 & Fold each panel in half. & 6304 \\
 & Sew your panels all together in a row, selvage to selvage starting at the hip cut, ignoring any difference at the bottom, as this is the side you will cut and hem. & \multirow{2}{*}{4673} \\
 & Put your skirt on a clamping slack hanger and hang it high. & 7088 \\
 & Put on your giant, uncut skirt. & 8316 \\
 & Hem the skirt with a machine by rolling the fabric between your thumb and forefinger and guide it through the machine without stretching. & \multirow{2}{*}{2863} \\
 & Make an over-skirt from leftover fabric. & 3936 \\
 & Put on your full belly dance skirt, the over-skirt, and a belly dance belt and you are ready to dance! & \multirow{2}{*}{4251} \\ \midrule
How to Make Leggings from Tights & Find suitable tights. & 10001 \\
 & Decide upon the length of the leggings you'd prefer. & 9045 \\
 & Cut your tights at the chosen length. & 6603 \\
 & Turn the cut fabric over twice to create a half-inch (1.27cm) hemline. & 6342 \\
 & Stitch the hems in place by hand. & 9276 \\
 & If wished, you can add embroidery, lace, beads, sequins, etc. & 6563 \\
 & Wear your new leggings. & 6513 \\
\bottomrule
\end{tabularx}
\end{center}
\vspace{-10pt}
\caption{Pseudo labels of \textbf{Task Context Learning}
produced by the \g for the video segment shown in \red{Fig.~\ref{appendix_fig:qua_620069}}.}
\label{tab:qua_620069_tcl}
\end{table*}

\end{document}